\pdfoutput=1
\documentclass[10pt,letterpaper]{article}

\usepackage[plainheader]{rad_arxiv}
\usepackage{amsmath,amsfonts,bm}

\def\eqref#1{equation~\ref{#1}}

\def\1{\bm{1}}

\def\rvepsilon{{\bm{\epsilon}}}

\def\rvf{{\mathbf{f}}}

\def\rvh{{\mathbf{h}}}
\def\rvu{{\mathbf{i}}}

\def\rvu{{\mathbf{u}}}

\def\rvw{{\mathbf{w}}}
\def\rvx{{\mathbf{x}}}
\def\rvy{{\mathbf{y}}}
\def\rvz{{\mathbf{z}}}

\def\vzero{{\bm{0}}}

\def\vmu{{\bm{\mu}}}

\def\va{{\bm{a}}}
\def\vb{{\bm{b}}}
\def\vc{{\bm{c}}}

\def\ve{{\bm{e}}}

\def\vg{{\bm{g}}}
\def\vh{{\bm{h}}}

\def\vm{{\bm{m}}}

\def\vv{{\bm{v}}}

\def\vy{{\bm{y}}}

\def\mF{{\bm{F}}}

\def\mI{{\bm{I}}}

\def\mL{{\bm{L}}}

\def\mSigma{{\bm{\Sigma}}}

\DeclareMathAlphabet{\mathsfit}{\encodingdefault}{\sfdefault}{m}{sl}
\SetMathAlphabet{\mathsfit}{bold}{\encodingdefault}{\sfdefault}{bx}{n}

\def\gH{{\mathcal{H}}}

\def\gN{{\mathcal{N}}}

\def\gU{{\mathcal{U}}}

\newcommand{\E}{\mathbb{E}}
\newcommand{\Ls}{\mathcal{L}}
\newcommand{\R}{\mathbb{R}}

\newcommand{\KL}{D_{\mathrm{KL}}}

\usepackage{amsmath}
\usepackage{booktabs}
\usepackage{colortbl}
\usepackage{algorithm}
\usepackage[noend]{algpseudocode}
\usepackage{url}
\usepackage{hyperref}
\usepackage{etoc}
\hypersetup{
  colorlinks=true,
  allcolors=radRust,
  pdftitle={Representation-Augmented Diffusions},
  pdfauthor={Farrin Marouf Sofian, Kushagra Pandey, Stephan Mandt},
}

\newcommand{\tabcite}[1]{[\citenum{#1}]}

\usepackage{pifont}
\newcommand{\cmark}{\textcolor{radRust}{\ding{51}}}
\newcommand{\xmark}{\textcolor{radGray}{\ding{55}}}

\usepackage{tikz}
\usetikzlibrary{arrows.meta,backgrounds,calc,fit,positioning}
\tikzset{
  gm/.style={>={Latex[length=3.5pt,width=3pt]}, line width=0.6pt, font=\small, node
  distance=0.38cm},
  gmvar/.style={circle, draw, fill=white, minimum size=0.82cm, inner sep=0pt},
  gmx/.style={gmvar, draw=radIndigo},
  gmx0/.style={gmvar, draw=radIndigo, fill=radIndigo!22},
  gmu/.style={gmvar, draw=radRust, fill=radRust!15},
  gmedge/.style={->, radIndigo},
  gmradedge/.style={->, radRust},
  gmdots/.style={font=\small, text=radGray, inner sep=1pt},
  gmpanel/.style={rounded corners=3pt, inner sep=5pt},
}

\title{Representation-Augmented Diffusions}
\author{Farrin Marouf Sofian\authormark{1,$\ast$},
        Kushagra Pandey\authormark{1,$\ast$},
        Stephan Mandt\authormark{1}}
\affiliation{\authormark{1}University of California, Irvine
             \qquad \authormark{$\ast$}Equal contribution}
\correspondence{Farrin Marouf Sofian at \email{fmaroufs@uci.edu}}

\newcommand{\figOverview}{%
\begin{figure}[!h]
    \centering
    \includegraphics[width=\linewidth]{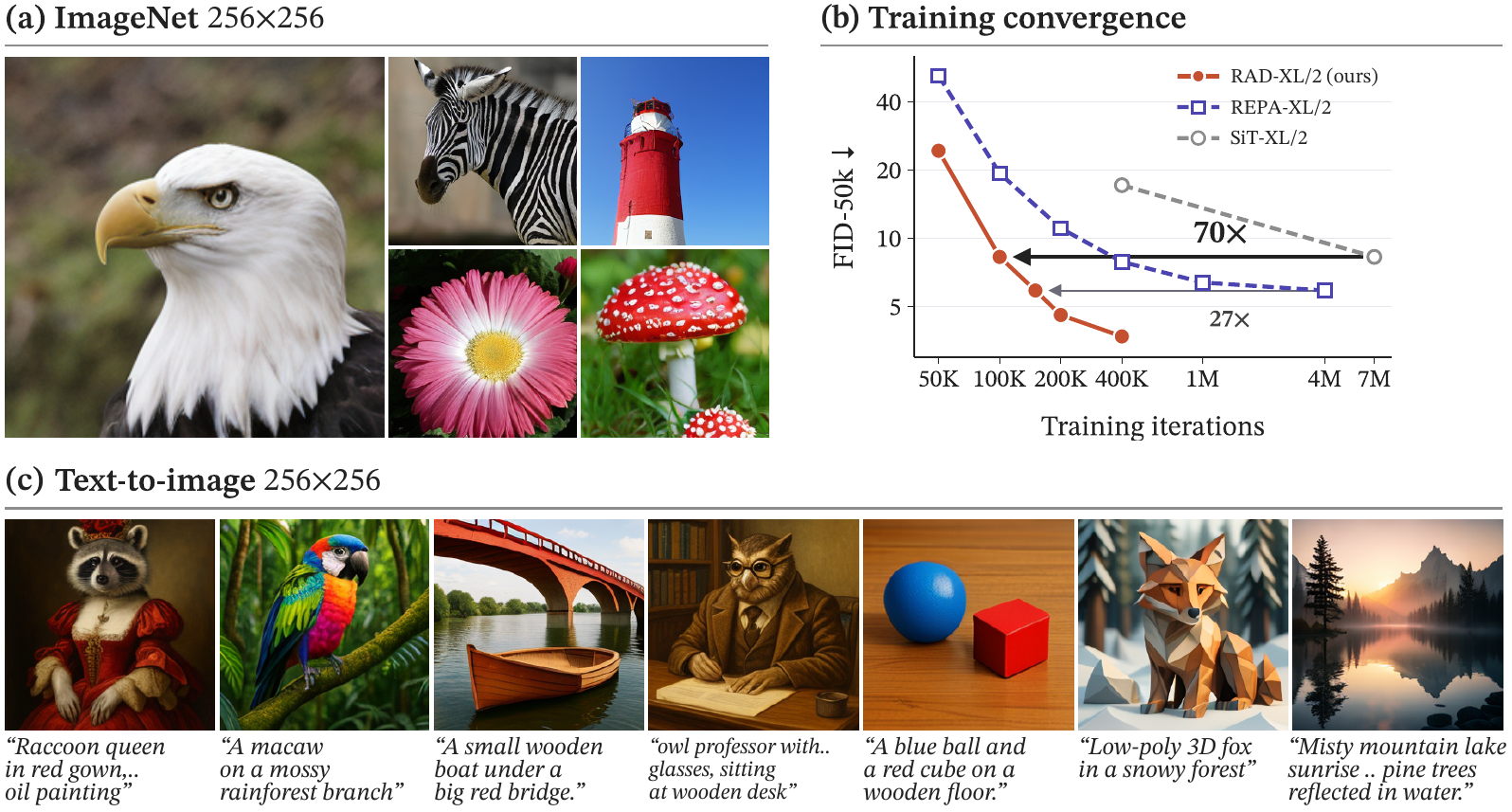}
    \caption{\textbf{Representation-Augmented Diffusions (RAD) generate high-quality samples
    while converging faster during training.} (a) ImageNet samples. (b) Training
    convergence on ImageNet: RAD matches the FID of SiT \citep{ma2024sit} with
    $70\times$ and of REPA \citep{yu2025repa} with $27\times$ fewer iterations.
    (c) Text-to-image samples after only 400k iterations: RAD follows prompts
    faithfully across object composition, attributes and artistic styles.}
    \label{fig:overview}
\end{figure}
}

\newcommand{\figGraphicalModel}{%
\begin{figure}[!t]
    \centering
    \begin{subfigure}[b]{0.49\linewidth}
        \centering
        \begin{tikzpicture}[gm, node distance=0.5cm]
            \node[gmx] (xT) {$\rvx_T$};
            \node[gmdots, right=of xT] (d1) {$\cdots$};
            \node[gmx, right=of d1] (xt) {$\rvx_t$};
            \node[gmx, right=of xt] (xt1) {$\rvx_{t-1}$};
            \node[gmdots, right=of xt1] (d2) {$\cdots$};
            \node[gmx0, right=of d2] (x0) {$\rvx_0$};
            \node[gmvar, draw=none, fill=none, below=0.4cm of $(xt)!0.5!(xt1)$] (ph) {};
            \draw[gmedge] (xT) -- (d1);
            \draw[gmedge] (d1) -- (xt);
            \draw[gmedge] (xt) -- (xt1);
            \draw[gmedge] (xt1) -- (d2);
            \draw[gmedge] (d2) -- (x0);
            \begin{scope}[on background layer]
                \node[gmpanel, fill=radIndigoTint, fit=(xT) (x0) (ph)] {};
            \end{scope}
        \end{tikzpicture}
        \caption{Standard diffusion}
    \end{subfigure}
    \hfill
    \begin{subfigure}[b]{0.49\linewidth}
        \centering
        \begin{tikzpicture}[gm, node distance=0.5cm]
            \node[gmx] (xT) {$\rvx_T$};
            \node[gmdots, right=of xT] (d1) {$\cdots$};
            \node[gmx, right=of d1] (xt) {$\rvx_t$};
            \node[gmx, right=of xt] (xt1) {$\rvx_{t-1}$};
            \node[gmdots, right=of xt1] (d2) {$\cdots$};
            \node[gmx0, right=of d2] (x0) {$\rvx_0$};
            \node[gmu, below=0.4cm of $(xt)!0.5!(xt1)$] (ut) {$\rvu_t$};
            \draw[gmedge] (xT) -- (d1);
            \draw[gmedge] (d1) -- (xt);
            \draw[gmedge] (xt) -- (xt1);
            \draw[gmedge] (xt1) -- (d2);
            \draw[gmedge] (d2) -- (x0);
            \draw[gmradedge] (xt) -- (ut);
            \draw[gmradedge] (ut) -- (xt1);
            \begin{scope}[on background layer]
                \node[gmpanel, fill=radTint, fit=(xT) (x0) (ut)] {};
            \end{scope}
        \end{tikzpicture}
        \caption{RAD (ours)}
    \end{subfigure}
    \caption{\textbf{RAD inserts a latent representation into every denoising
    step.} Graphical models of the generative process of (a) Standard Diffusion
    and (b) RAD.}
    \label{fig:graphical_model}
\end{figure}
}

\newcommand{\radchg}[1]{{\setlength{\fboxsep}{1.2pt}\colorbox{radTint}{\color{radRust}#1}}}

\newcommand{\figAlgorithms}{%
\begin{figure}[t]
\footnotesize
\begin{minipage}[t]{0.49\linewidth}
\captionof{algorithm}{Training}
\label{alg:training}
\vspace{-4pt}\hrule\vspace{2pt}
\begin{algorithmic}
\Repeat
    \State $(\rvx_0, \vc) \sim$ data, $t \sim \gU(0, 1)$, $\rvepsilon \sim \gN(\vzero, \mI)$
    \State $\rvx_t \gets \alpha_t \rvx_0 + \sigma_t \rvepsilon$
    \State $\rvh_t \gets \vv_\theta^{(\ell)}(\rvx_t, t, \vc)$ \Comment{first $\ell$ blocks}
    \State \radchg{$\rvu_t \sim q(\rvu_t \mid \rvx_0) = \gN(\alpha_t f_{\mathrm{pre}}(\rvx_0),\, \beta^2 \sigma_t^2 \mI)$}
    \State $\vv \gets \vv_\theta^{(>\ell)}(\rvh_t \radchg{$+\, \mathrm{Inject}(\rvu_t)$}, t, \vc)$
    \State $\Ls \gets \|\vv - \vv_{\mathrm{ref}}\|_2^2 + \radchg{$\KL(q(\rvu_t \mid \rvx_0) \| p_\phi(\rvu_t \mid \rvx_t))$}$
    \State Gradient step on $(\theta, \phi)$
\Until{converged}
\end{algorithmic}
\vspace{2pt}\hrule
\end{minipage}\hfill
\begin{minipage}[t]{0.49\linewidth}
\captionof{algorithm}{Sampling (Euler, $N$ steps)}
\label{alg:sampling}
\vspace{-4pt}\hrule\vspace{2pt}
\begin{algorithmic}
\State $\rvx \sim \gN(\vzero, \mI)$, \; $\Delta \gets 1/N$
\For{$i = N, \dots, 1$}
    \State $t \gets i \Delta$
    \State $\rvh_t \gets \vv_\theta^{(\ell)}(\rvx, t, \vc)$ \Comment{first $\ell$ blocks}
    \State \radchg{$\rvu_t \sim p_\phi(\rvu_t \mid \rvx)$}
    \State $\vv \gets \vv_\theta^{(>\ell)}(\rvh_t \radchg{$+\, \mathrm{Inject}(\rvu_t)$}, t, \vc)$
    \State $\rvx \gets \rvx - \Delta\, \vv$
\EndFor
\State \Return $\rvx$
\end{algorithmic}
\vspace{2pt}\hrule
\end{minipage}
\caption*{\textbf{Comparison of training and sampling between RAD and standard flow matching/diffusion models.} Highlighted parts denote the additions in RAD. Here, $q(\rvu_t \mid \rvx_0)$ is a fixed Gaussian distribution. The $\mathrm{Inject}$ operation is defined in Eq.~\ref{eq:scale_shift} (see the architecture in Fig.~\ref{fig:architecture}, App.~\ref{sec:design}).}
\end{figure}
}

\newcommand{\figEfficiency}{%
\begin{figure}[t]
\centering
\includegraphics[width=\linewidth]{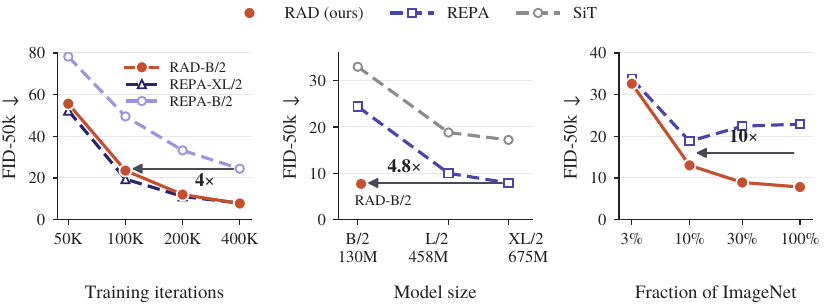}\\[-12pt]
\begin{subfigure}[t]{0.33333\linewidth}
\centering
\caption{B/2 training}
\end{subfigure}\hfill
\begin{subfigure}[t]{0.33333\linewidth}
\centering
\caption{Model size at 400K}
\end{subfigure}\hfill
\begin{subfigure}[t]{0.33333\linewidth}
\centering
\caption{B/2 with less data}
\end{subfigure}
\caption{\textbf{RAD converges faster and is more parameter- and data-efficient
    than REPA and SiT.} (a) RAD-B/2 converges $4\times$ faster than REPA-B/2 and
    is comparable to REPA-XL/2 during training. (b) RAD-B/2 is $4.8\times$ more
    parameter efficient than REPA. (c) RAD-B/2 is $10\times$ more data efficient
    than REPA-B/2.}
\label{fig:efficiency}
\end{figure}
}

\newcommand{\figSamplingEfficiency}{%
\begin{figure}[t]
\centering
\begin{subfigure}[t]{0.33333\linewidth}
\centering
\includegraphics[width=\linewidth]{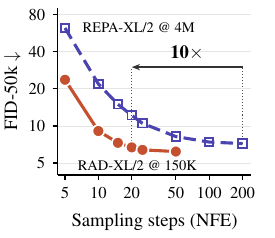}
\caption{RAD vs.\ REPA, Euler ODE}
\label{fig:sampling_nfe}
\end{subfigure}\hfill
\begin{subfigure}[t]{0.33333\linewidth}
\centering
\includegraphics[width=\linewidth]{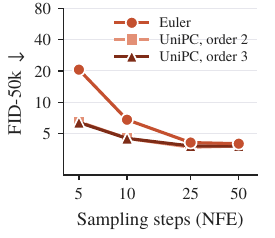}
\caption{ODE samplers}
\label{fig:samplers}
\end{subfigure}\hfill
\begin{subfigure}[t]{0.33333\linewidth}
\centering
\includegraphics[width=\linewidth]{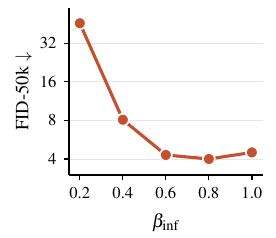}
\caption{Inference policy scale}
\label{fig:beta_inf_fid}
\end{subfigure}
\caption{\textbf{Inference with RAD.} (a) FID-50k vs.\ sampling steps for
    RAD-XL/2 (150K) and REPA-XL/2 (4M) with the Euler ODE sampler. (b) Euler
    vs.\ UniPC for RAD-XL/2 (400K). (c) FID-50k of RAD-XL/2 (400K) as
    $\beta_{\mathrm{inf}}$ varies; training uses $\beta = 1$, Euler ODE sampler
    with 50 steps. No guidance.}
\label{fig:sampling_efficiency_xl}
\end{figure}
}

\newcommand{\figInterpolation}{%
\begin{figure}[t]
\centering
\includegraphics[width=\linewidth]{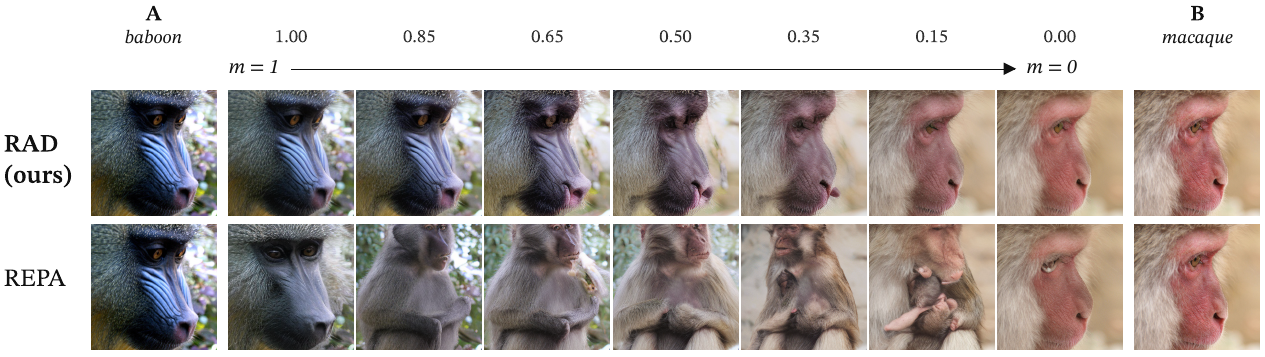}
\caption{\textbf{RAD interpolates smoothly in representation space.}
Top: RAD-XL/2 (400K iters) preserves attributes like pose and viewpoint.
Bottom: REPA-XL/2 (4M iters) yields artifacts and often changes semantics.}
\label{fig:interpolation}
\end{figure}
}

\newcommand{\figRegTextToImageLinearProbe}{%
\begin{figure}[t]
\begin{minipage}[t]{2.85in}
\centering
\captionof{table}{\textbf{RAD improves REG training convergence.} FID-50k, Euler
SDE, w/o CFG}
\label{tab:reg_convergence}
\fontsize{7.7}{9.2}\selectfont
\setlength{\tabcolsep}{10.7pt}
\renewcommand{\arraystretch}{1.3}
\begin{tabular}{llrc}
\toprule
Model & \#Params & Iter. & FID$\downarrow$ \\
\midrule
\multicolumn{4}{l}{\textit{SiT-B/2}~\tabcite{ma2024sit}} \\
REG~\tabcite{wu2025reg} & 132M & 100K & 36.1 \\
\rowcolor{radTint} REG + RAD (ours) & 143M & 100K & \textbf{22.4} \\
REG~\tabcite{wu2025reg} & 132M & 400K & 15.2 \\
\rowcolor{radTint} REG + RAD (ours) & 143M & 400K & \textbf{8.9} \\
\midrule
\multicolumn{4}{l}{\textit{SiT-L/2}~\tabcite{ma2024sit}} \\
REG~\tabcite{wu2025reg} & 460M & 100K & 11.4 \\
\rowcolor{radTint} REG + RAD (ours) & 474M & 100K & \textbf{7.8} \\
REG~\tabcite{wu2025reg} & 460M & 400K & 4.6 \\
\rowcolor{radTint} REG + RAD (ours) & 474M & 400K & \textbf{3.9} \\
\bottomrule
\end{tabular}

\vspace{19pt}
\captionof{table}{\textbf{RAD improves text-to-image FID over REPA at both
    scales.} FID-50k and GenEval overall score on BLIP3o pre-training captions
    at $256\times256$ with a frozen Qwen3-0.6B text encoder. Best in bold. No
    guidance.}
\label{tab:t2i}
\fontsize{7.7}{9.2}\selectfont
\setlength{\tabcolsep}{7.25pt}
\renewcommand{\arraystretch}{1.3}
\begin{tabular}{l cc cc}
\toprule
& \multicolumn{2}{c}{B/2} & \multicolumn{2}{c}{XL/2} \\
\cmidrule(lr){2-3}\cmidrule(lr){4-5}
Method & FID$\downarrow$ & GenEval$\uparrow$ & FID$\downarrow$ & GenEval$\uparrow$ \\
\midrule
REPA~\tabcite{yu2025repa} & 10.77 & 0.138 & 4.31 & 0.258 \\
\rowcolor{radTint} RAD (ours) & \textbf{8.42} & \textbf{0.166} & \textbf{3.94} & \textbf{0.266} \\
\bottomrule
\end{tabular}
\end{minipage}\hfill
\begin{minipage}[t]{2.85in}
\centering
\vspace{0pt}
\begin{subfigure}{\linewidth}
\centering
\includegraphics[width=\linewidth]{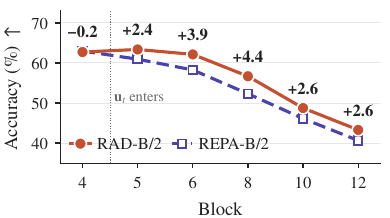}
\caption{\textbf{RAD features remain linearly separable by class deeper into the network.}
    Per-block accuracy of RAD-B/2 vs.\ REPA-B/2 (both aligned at block 4).}
\end{subfigure}\\[2pt]
\begin{subfigure}{\linewidth}
\centering
\includegraphics[width=\linewidth]{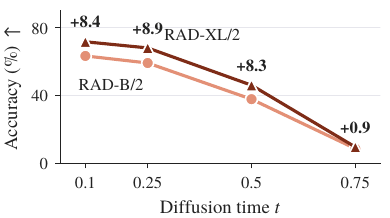}
\caption{\textbf{RAD representations improve with scale.}
    Best-block accuracy vs.\ $t$ for RAD-B/2 and RAD-XL/2.}
\end{subfigure}
\captionof{figure}{\textbf{Linear probing with RAD.}}
\label{fig:linear_probe}
\end{minipage}
\end{figure}
}

\newcommand{\figArchitecture}{%
\begin{figure}[t]
\centering
\includegraphics[width=\linewidth]{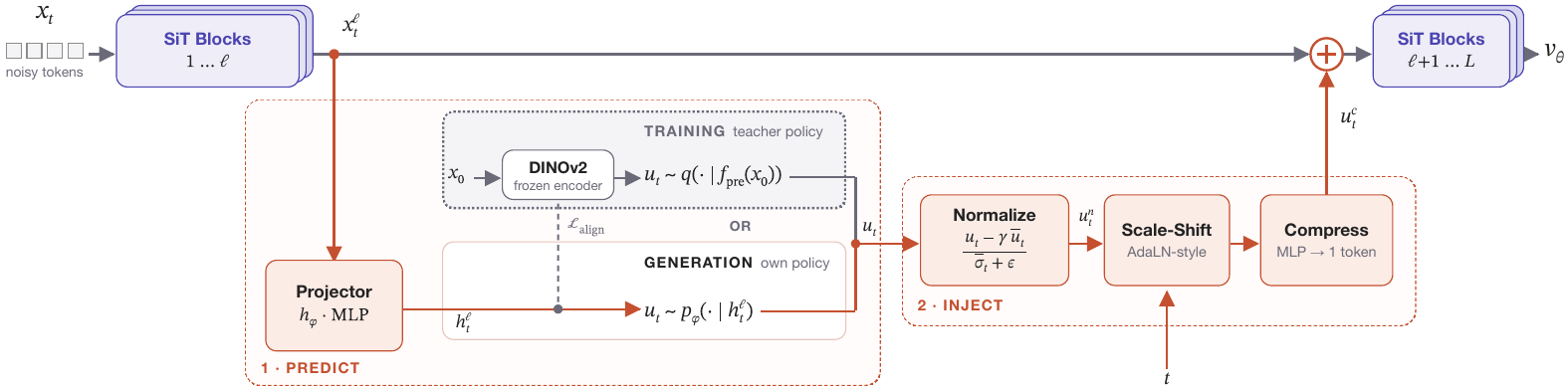}
\caption{\textbf{RAD architecture.} (1) Predict: the projector $h_\phi$ maps
the hidden state $\rvx_t^{\ell}$ after the first $\ell$ SiT blocks to the
representation policy. During training, $\rvu_t$ is sampled from the teacher
policy $q(\cdot \mid f_{\mathrm{pre}}(\rvx_0))$, and during generation from the
model's own policy $p_\phi(\cdot \mid h_t^{\ell})$. (2) Inject: $\rvu_t$ is
normalized, compressed to a single token and modulated
    (Eq.~\ref{eq:scale_shift}),
then added to the hidden state before the remaining SiT blocks predict
$\vv_\theta$.}
\label{fig:architecture}
\end{figure}
}

\newcommand{\figSamplingNfeSde}{%
\begin{center}
\begin{minipage}{\radfloatwidth}
\centering
\includegraphics[scale=1.0909]{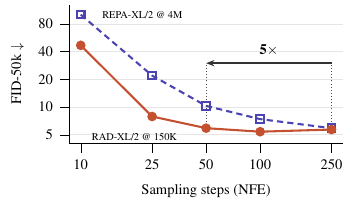}
\captionof{figure}{\textbf{RAD vs.\ REPA, Euler--Maruyama SDE.} FID-50k vs.\ sampling
    steps for RAD-XL/2 (150K) and REPA-XL/2 (4M) with the Euler--Maruyama SDE
    sampler. No guidance. Values are listed in
    Table~\ref{tab:sampling_efficiency_sde}.}
\label{fig:sampling_nfe_sde}
\end{minipage}
\end{center}
}

\newcommand{\figBetaInfTrain}{%
\begin{center}
\begin{minipage}{\radfloatwidth}
\centering
\begin{minipage}[t]{0.525\linewidth}
\centering
\includegraphics[scale=0.96]{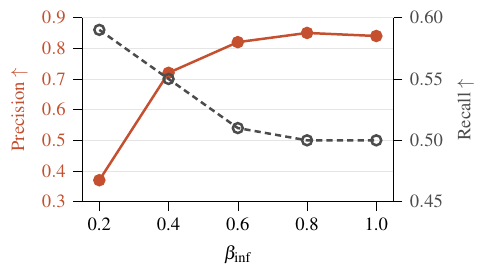}
\captionof{figure}{\textbf{Effect of the inference policy scale $\beta_{\mathrm{inf}}$ on
precision and recall.} Precision and recall on ImageNet $256\times256$ of
RAD-XL/2 trained for 400K iterations as $\beta_{\mathrm{inf}}$ varies (FID in
Fig.~\ref{fig:beta_inf_fid}). Training uses $\beta = 1$. Euler ODE sampler with
50 steps, no guidance. Values are listed in Table~\ref{tab:beta_inf}.}
\label{fig:beta_inf}
\end{minipage}\hfill
\begin{minipage}[t]{0.45\linewidth}
\centering
\includegraphics[scale=1.0909]{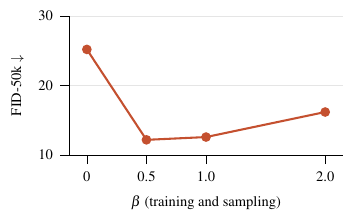}
\captionof{figure}{\textbf{Effect of the training policy scale $\beta$.} FID-50k on
ImageNet $256\times256$ of RAD-B/2 trained for 400K iterations with the
Gaussian policy (config \textsc{b} in Table~\ref{tab:system_analysis}), policy
scale $\beta$, and sampled with $\beta_{\mathrm{inf}} = \beta$. Euler--Maruyama SDE
sampler with 250 steps, no guidance. Values are listed in
Table~\ref{tab:beta_train}.}
\label{fig:beta_train}
\end{minipage}
\end{minipage}
\end{center}
}

\newcommand{\figInterpCross}{%
\begin{figure}[!htb]
\centering
\begin{subfigure}[t]{\linewidth}
\centering
\includegraphics[width=\linewidth]{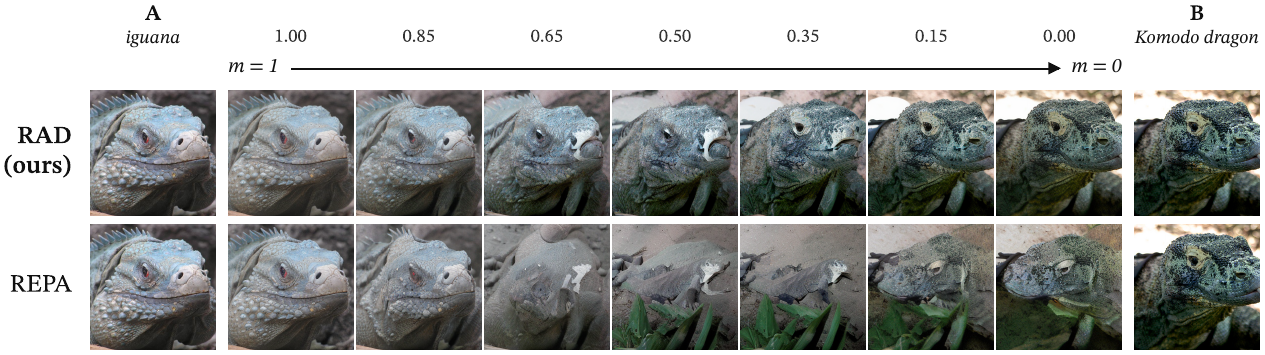}
\caption{Iguana $\rightarrow$ Komodo dragon}
\end{subfigure}

\vspace{0.5em}
\begin{subfigure}[t]{\linewidth}
\centering
\includegraphics[width=\linewidth]{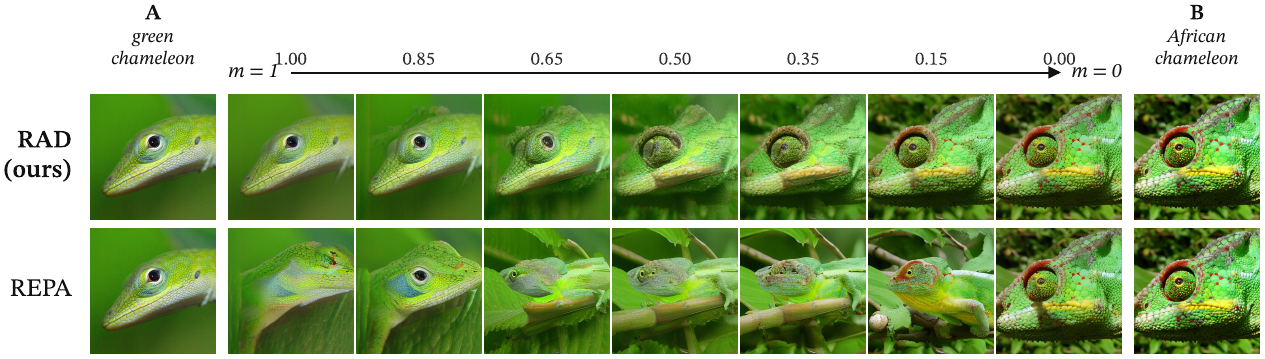}
\caption{Green chameleon $\rightarrow$ African chameleon}
\end{subfigure}

\vspace{0.5em}
\begin{subfigure}[t]{\linewidth}
\centering
\includegraphics[width=\linewidth]{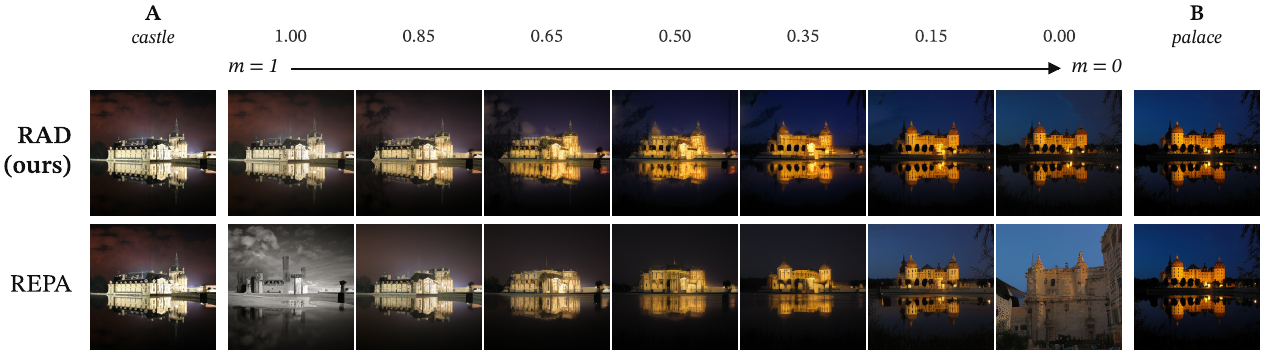}
\caption{Castle $\rightarrow$ palace}
\end{subfigure}
\caption{\textbf{Additional interpolations across classes.} Samples interpolated
from image A ($m=1$) to image B ($m=0$), with the real images shown in the outer
columns. Top rows: RAD-XL/2 trained for 400K iterations, interpolating the
representation, $\rvx_1$, and the class embedding (Eq.~\ref{eq:interp}). Bottom
rows: REPA-XL/2 trained for 4M iterations, i.e., $10\times$ longer than RAD,
interpolating $\rvx_1$ and the class embedding. No guidance.}
\label{fig:interp_cross}
\end{figure}
}

\newcommand{\figInterpSame}{%
\begin{figure}[!htb]
\centering
\includegraphics[width=\linewidth]{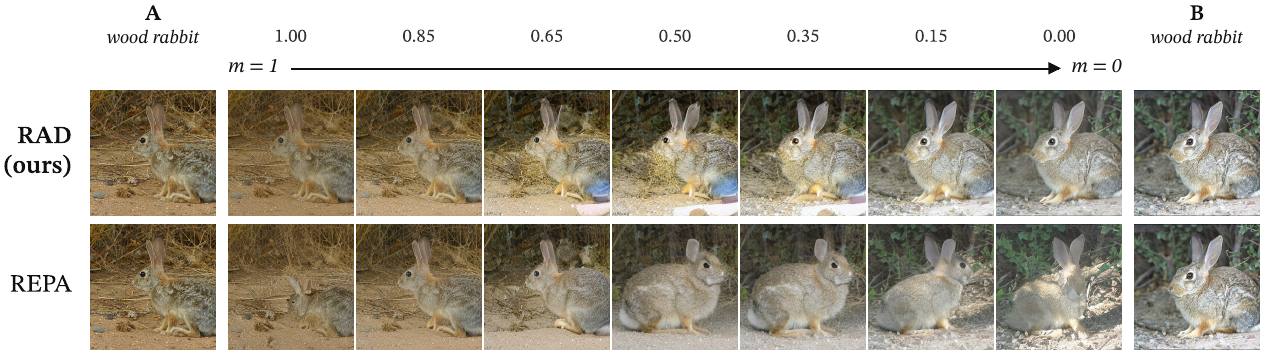}
\caption{\textbf{Additional interpolation within the same class.} Samples
interpolated between two wood rabbits, from image A ($m=1$) to image B ($m=0$),
with the real images shown in the outer columns. Top: RAD-XL/2 trained for 400K
iterations, interpolating the representation and $\rvx_1$. Bottom: REPA-XL/2
trained for 4M iterations, i.e., $10\times$ longer than RAD, interpolating
$\rvx_1$ only. The class embedding is fixed in both. No guidance.}
\label{fig:interp_same}
\end{figure}
}

\newcommand{\figSamplesMacaw}{%
\begin{figure}[!htb]
\centering
\includegraphics[width=\linewidth]{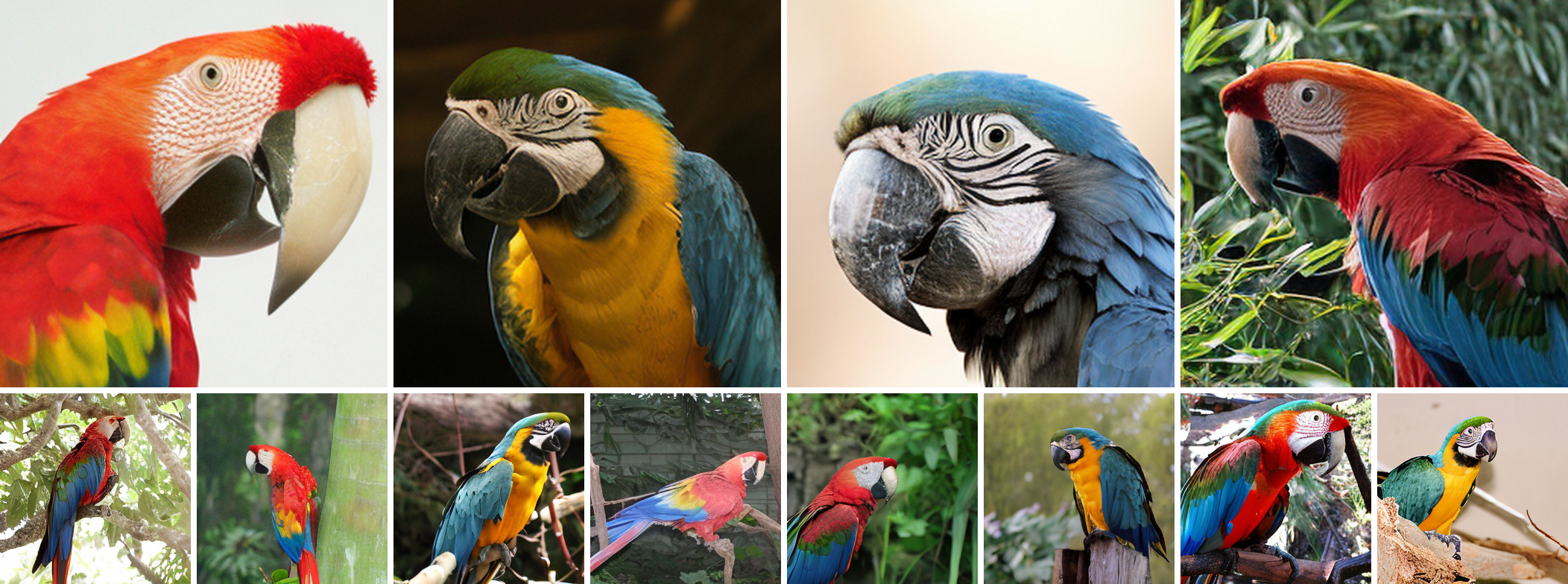}
\caption{Class-conditional samples from ImageNet (Macaw, class 88).}
\label{fig:samples_macaw}
\end{figure}
}

\newcommand{\figSamplesLorikeet}{%
\begin{figure}[!htb]
\centering
\includegraphics[width=\linewidth]{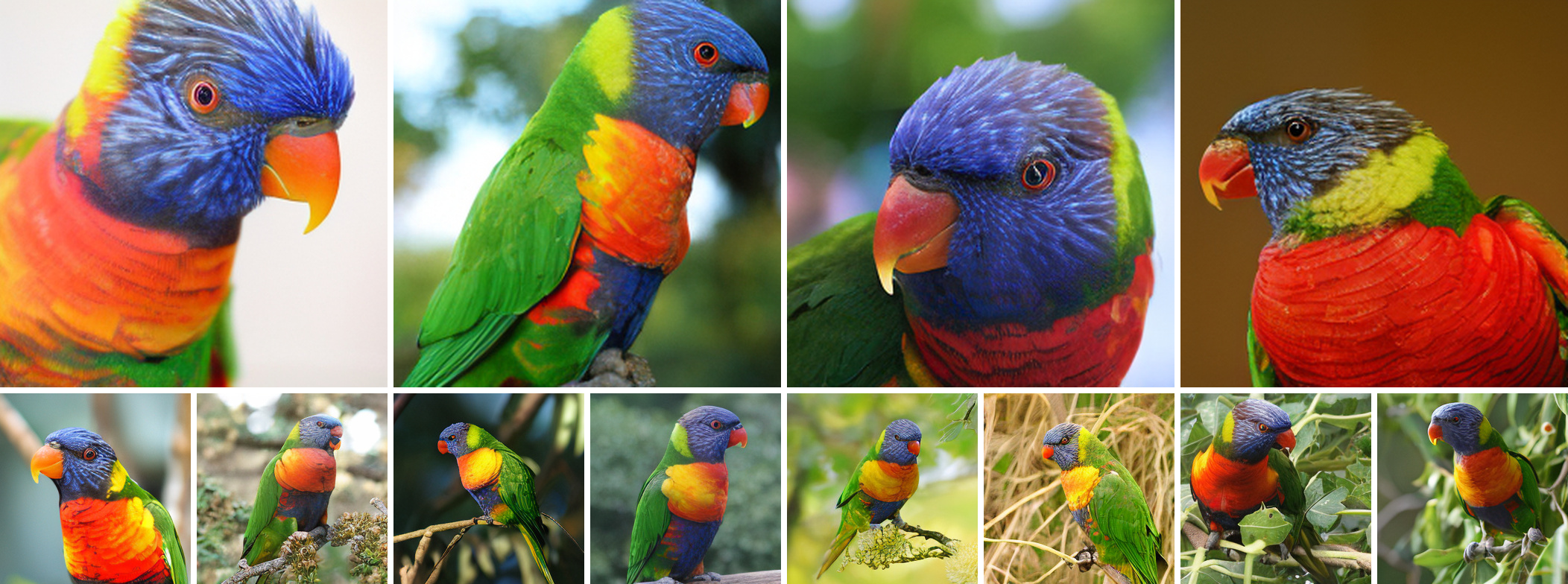}
\caption{Class-conditional samples from ImageNet (Lorikeet, class 90).}
\label{fig:samples_lorikeet}
\end{figure}
}

\newcommand{\figSamplesLesserPanda}{%
\begin{figure}[!htb]
\centering
\includegraphics[width=\linewidth]{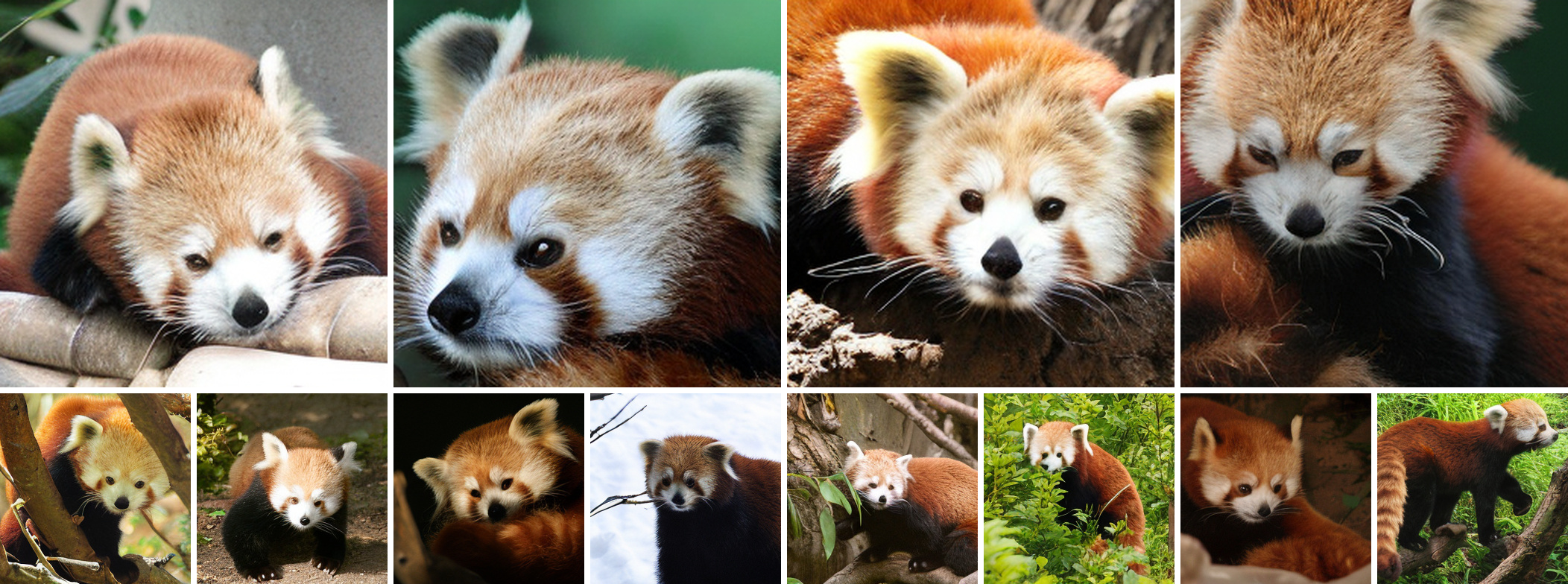}
\caption{Class-conditional samples from ImageNet (Red panda, class 387).}
\label{fig:samples_lesser_panda}
\end{figure}
}

\newcommand{\figSamplesGiantPanda}{%
\begin{figure}[!htb]
\centering
\includegraphics[width=\linewidth]{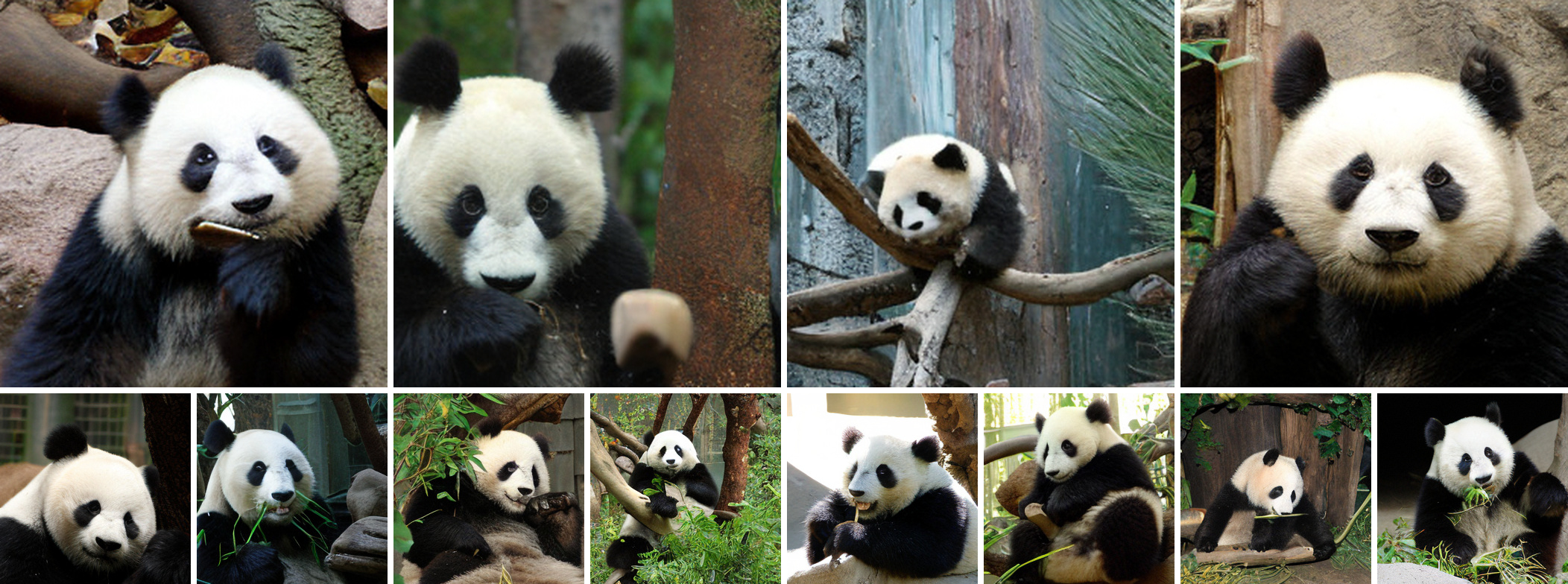}
\caption{Class-conditional samples from ImageNet (Giant panda, class 388).}
\label{fig:samples_giant_panda}
\end{figure}
}

\newcommand{\figSamplesRedFox}{%
\begin{figure}[!htb]
\centering
\includegraphics[width=\linewidth]{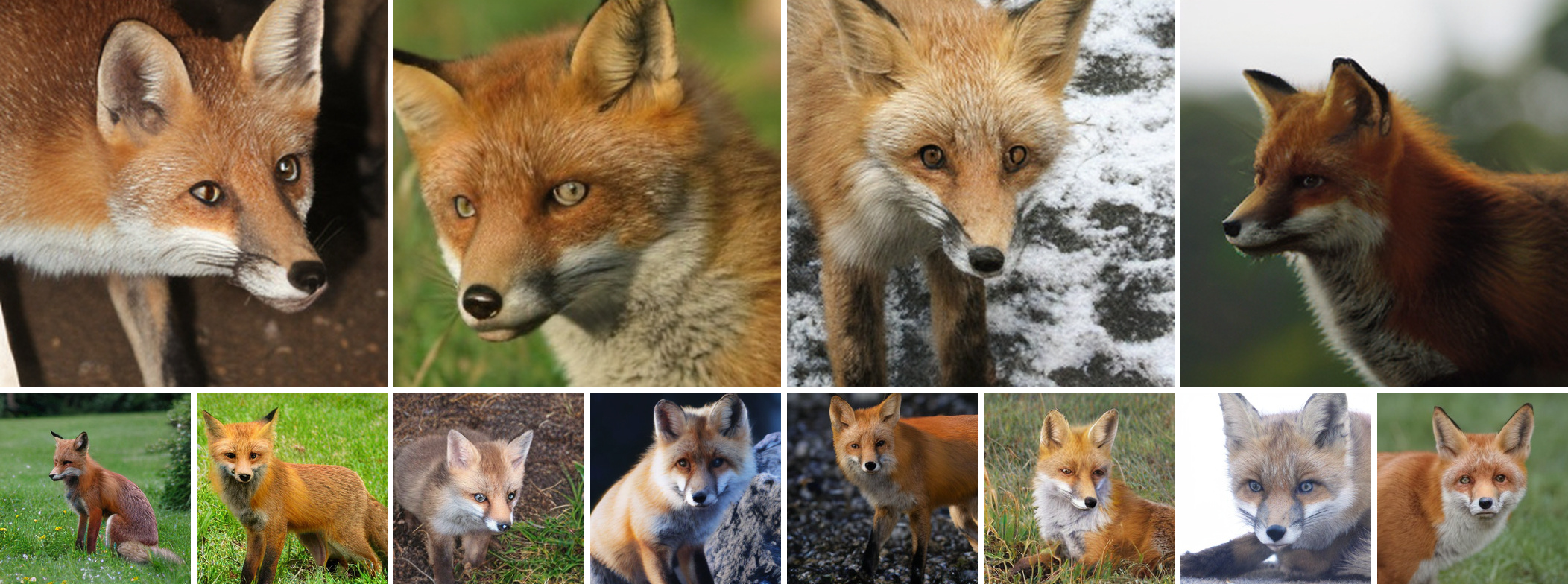}
\caption{Class-conditional samples from ImageNet (Red fox, class 277).}
\label{fig:samples_red_fox}
\end{figure}
}

\newcommand{\figSamplesAnemoneFish}{%
\begin{figure}[!htb]
\centering
\includegraphics[width=\linewidth]{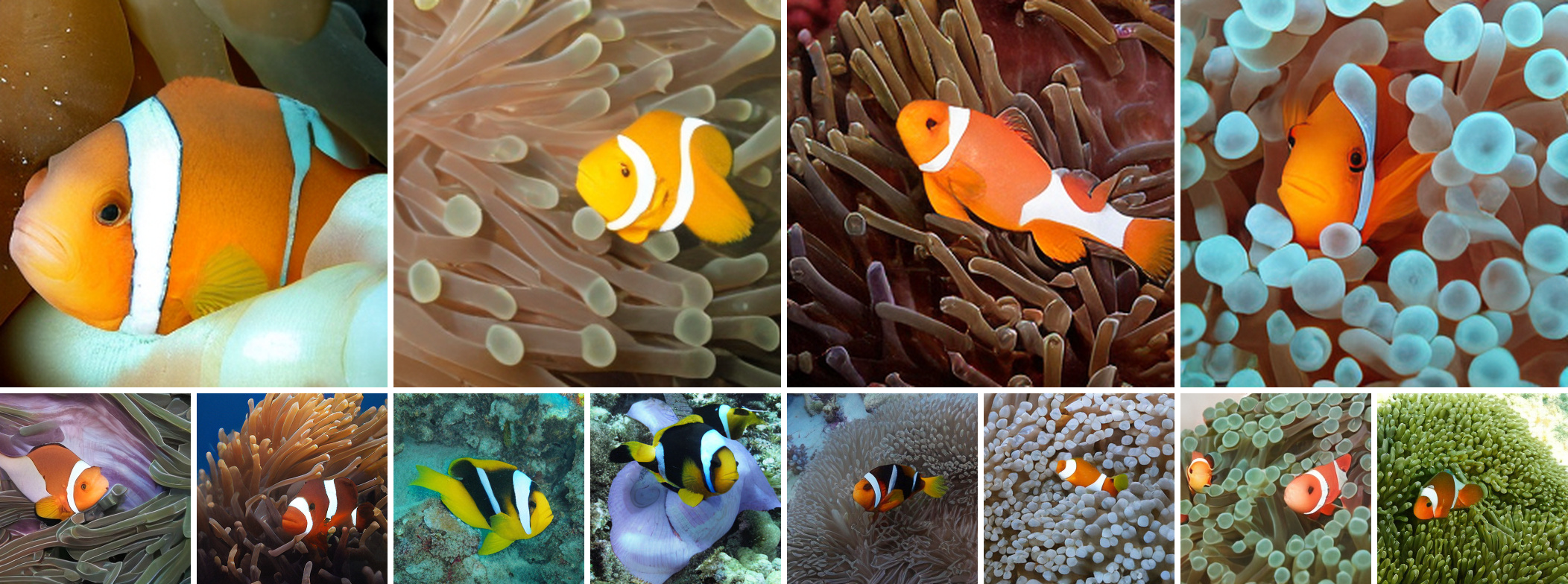}
\caption{Class-conditional samples from ImageNet (Anemone fish, class 393).}
\label{fig:samples_anemone_fish}
\end{figure}
}

\newcommand{\figSamplesAgaric}{%
\begin{figure}[!htb]
\centering
\includegraphics[width=\linewidth]{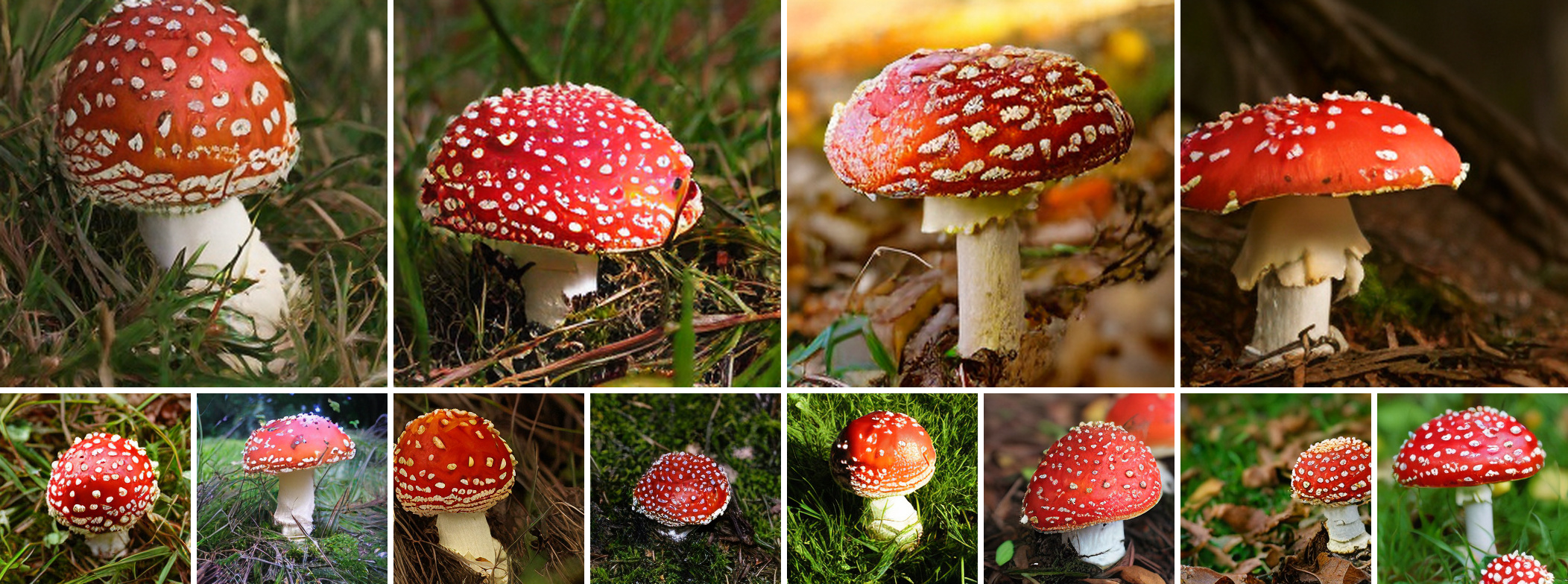}
\caption{Class-conditional samples from ImageNet (Agaric, class 992).}
\label{fig:samples_agaric}
\end{figure}
}

\newcommand{\figSamplesCoralFungus}{%
\begin{figure}[!htb]
\centering
\includegraphics[width=\linewidth]{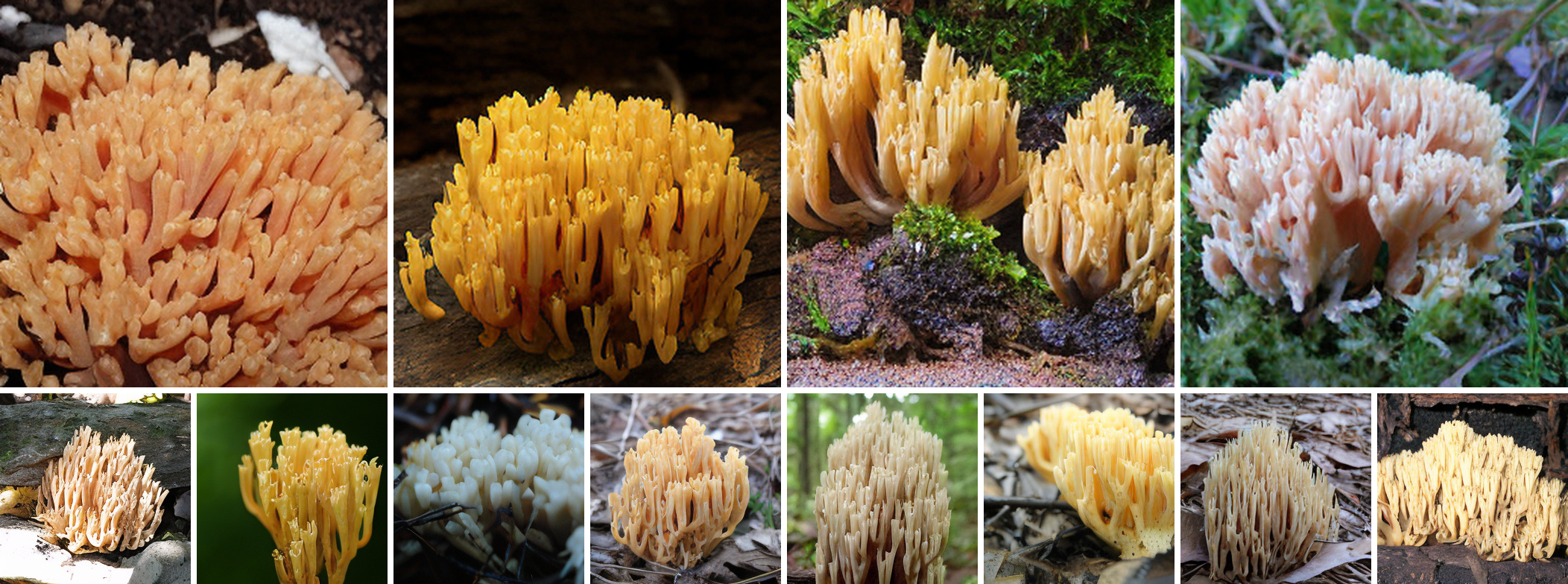}
\caption{Class-conditional samples from ImageNet (Coral fungus, class 991).}
\label{fig:samples_coral_fungus}
\end{figure}
}

\newcommand{\figSamplesStrawberry}{%
\begin{figure}[!htb]
\centering
\includegraphics[width=\linewidth]{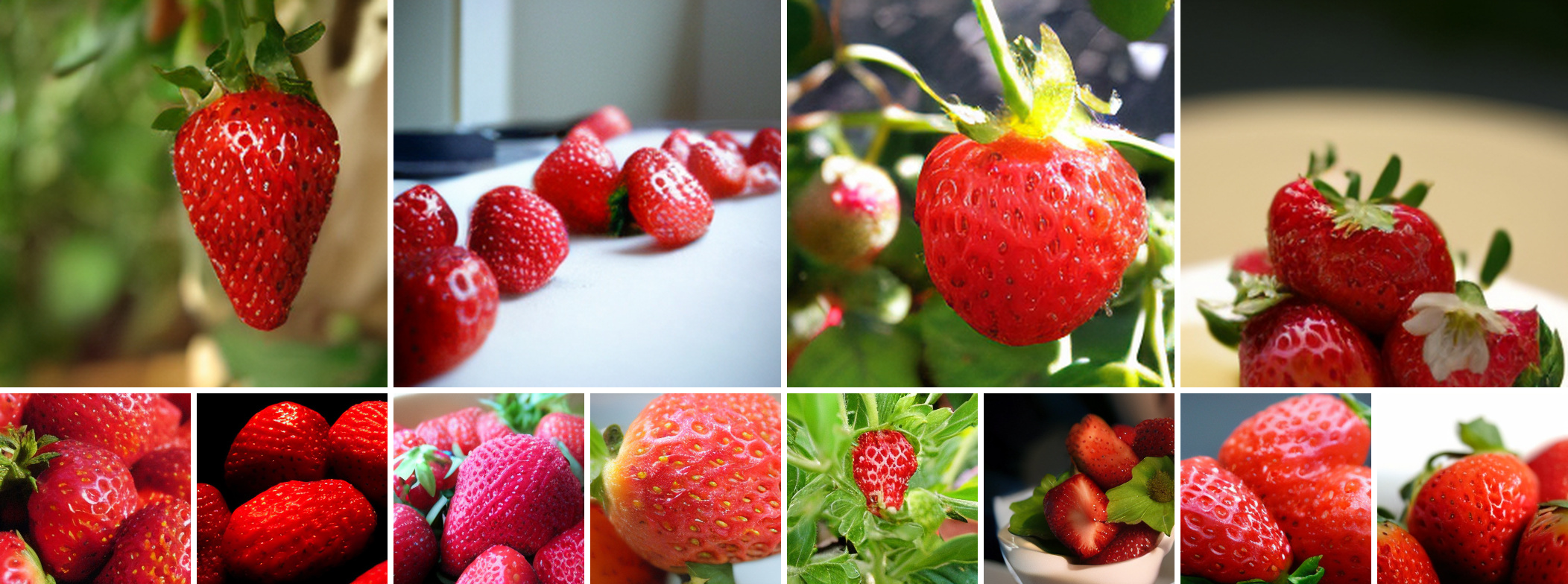}
\caption{Class-conditional samples from ImageNet (Strawberry, class 949).}
\label{fig:samples_strawberry}
\end{figure}
}

\newcommand{\figSamplesTeapot}{%
\begin{figure}[!htb]
\centering
\includegraphics[width=\linewidth]{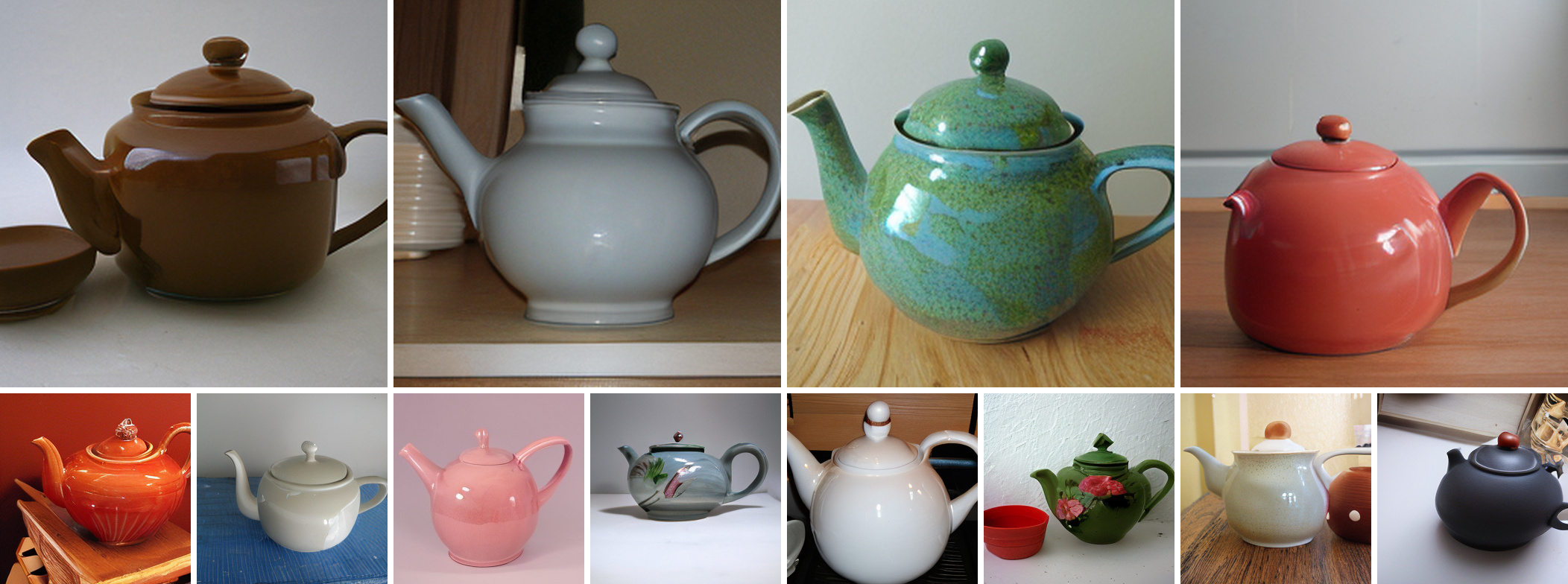}
\caption{Class-conditional samples from ImageNet (Teapot, class 849).}
\label{fig:samples_teapot}
\end{figure}
}

\newcommand{\figSamplesSchooner}{%
\begin{figure}[!htb]
\centering
\includegraphics[width=\linewidth]{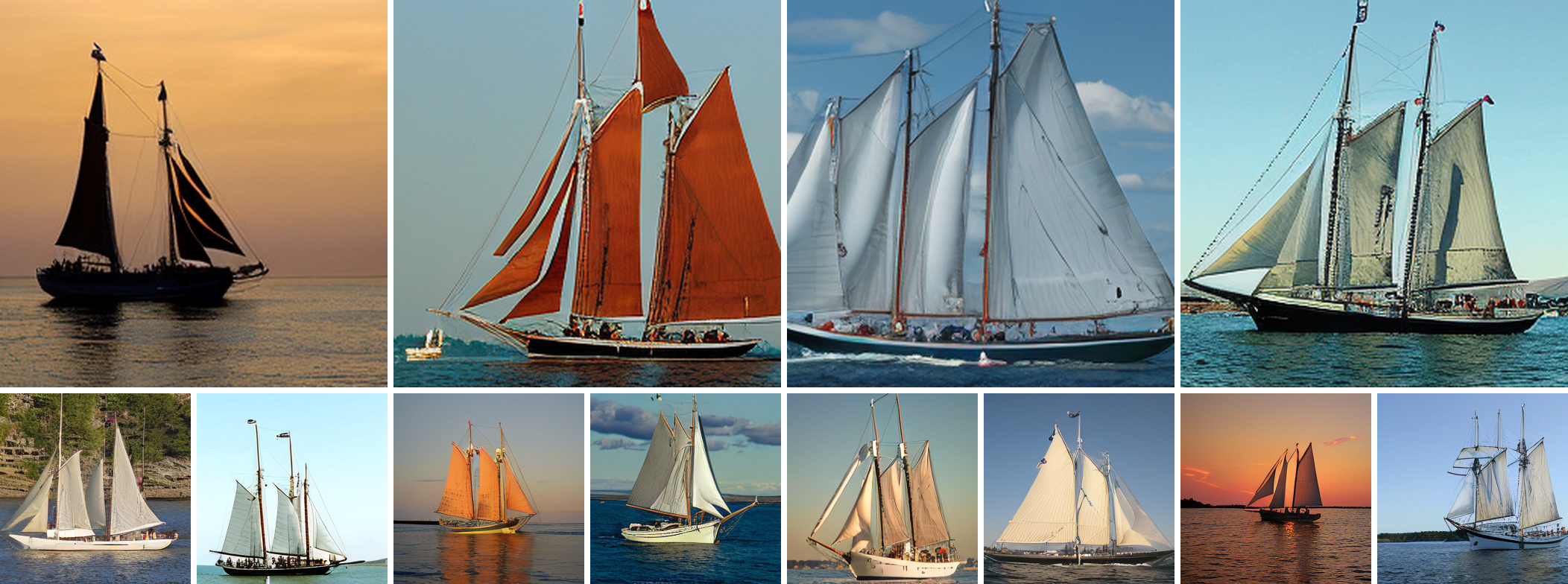}
\caption{Class-conditional samples from ImageNet (Schooner, class 780).}
\label{fig:samples_schooner}
\end{figure}
}

\newcommand{\figSamplesChurch}{%
\begin{figure}[!htb]
\centering
\includegraphics[width=\linewidth]{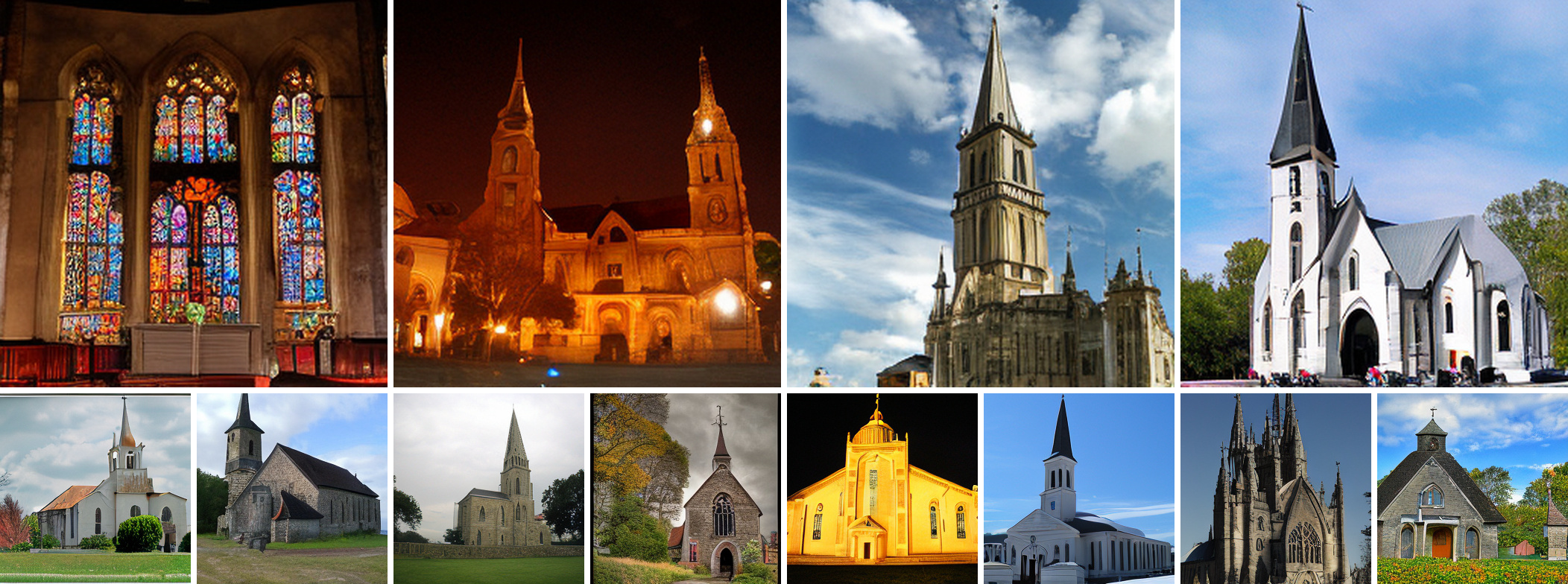}
\caption{Class-conditional samples from ImageNet (Church, class 497).}
\label{fig:samples_church}
\end{figure}
}

\newcommand{\figSamplesArcticFox}{%
\begin{figure}[!htb]
\centering
\includegraphics[width=\linewidth]{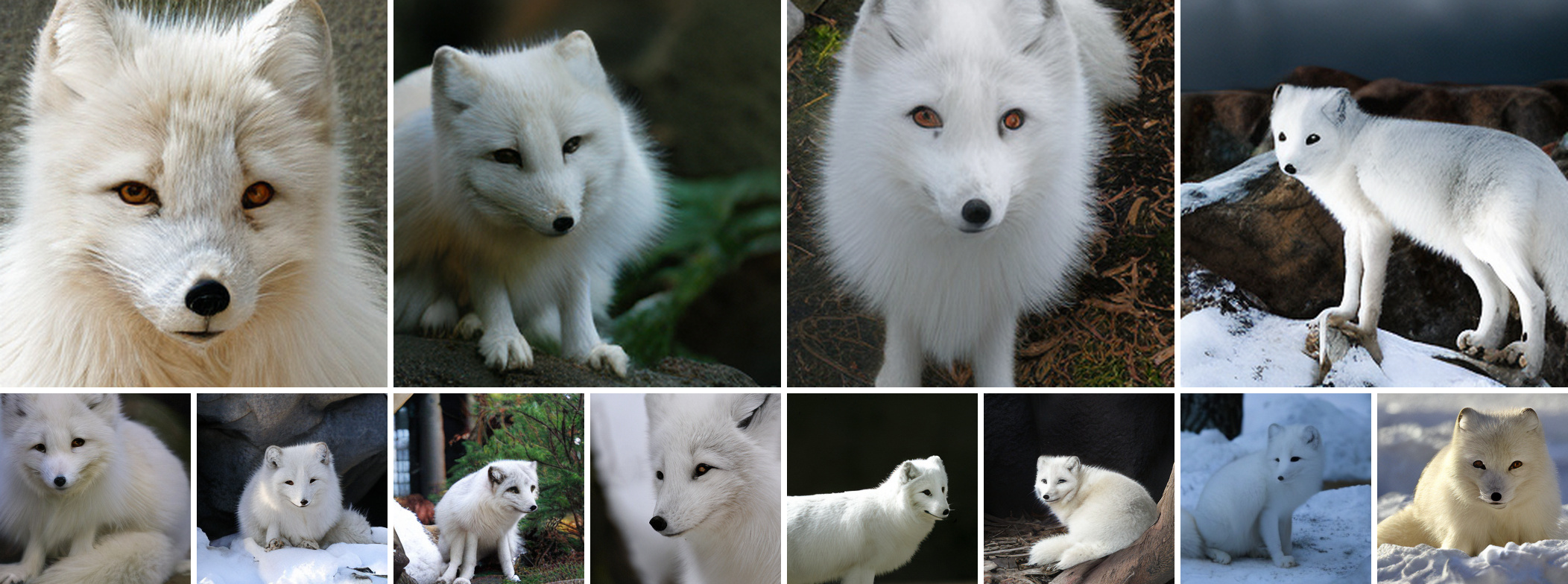}
\caption{Class-conditional samples from ImageNet (Arctic fox, class 279).}
\label{fig:samples_arctic_fox}
\end{figure}
}

\newcommand{\figSamplesValley}{%
\begin{figure}[!htb]
\centering
\includegraphics[width=\linewidth]{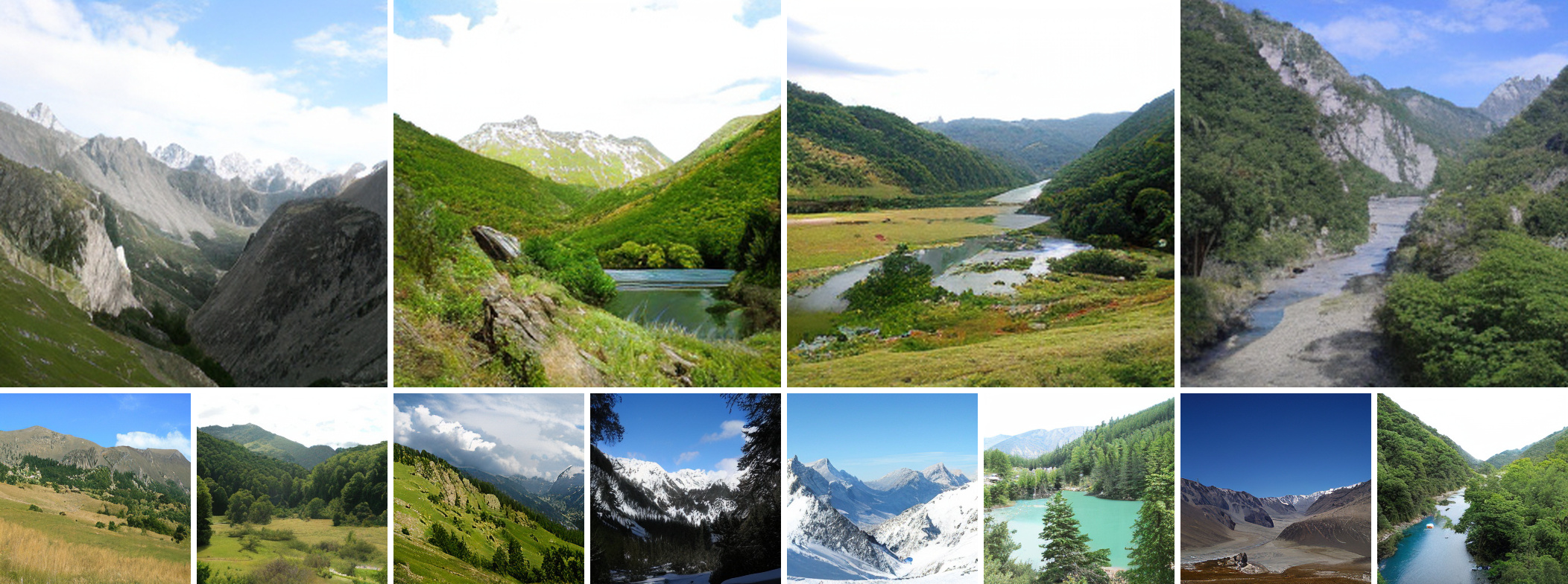}
\caption{Class-conditional samples from ImageNet (Valley, class 979).}
\label{fig:samples_valley}
\end{figure}
}

\newcommand{\figSamplesCheeseburger}{%
\begin{figure}[!htb]
\centering
\includegraphics[width=\linewidth]{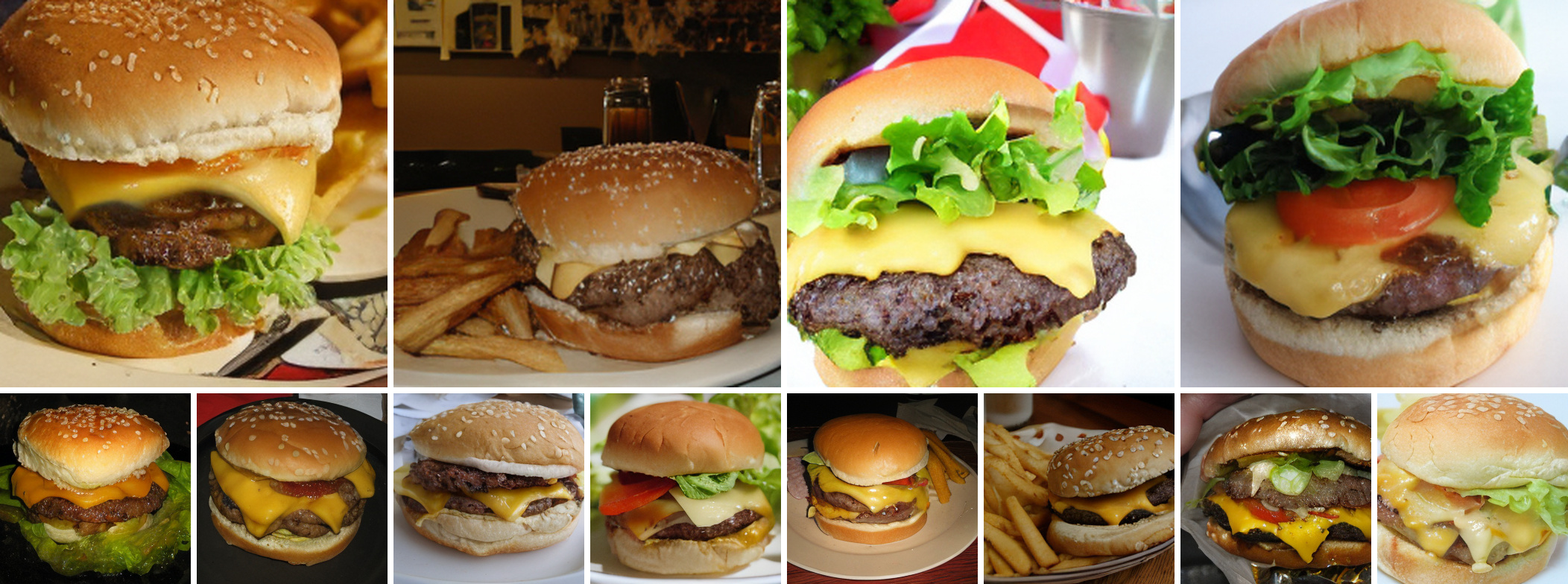}
\caption{Class-conditional samples from ImageNet (Cheeseburger, class 933).}
\label{fig:samples_cheeseburger}
\end{figure}
}

\newcommand{\figSamplesTextToImageFox}{%
\begin{figure}[!htb]
\centering
\includegraphics[width=\linewidth]{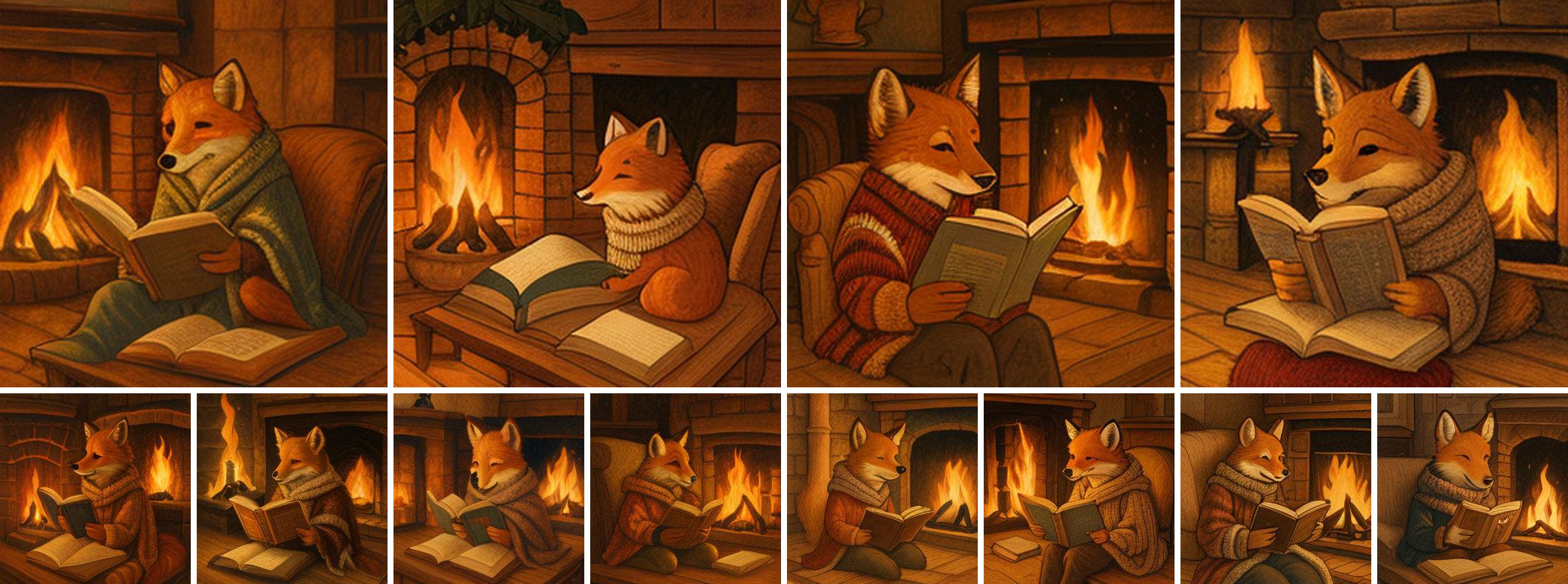}
\caption{``A fox wearing a knitted scarf, reading a book by a cozy fireplace,
warm storybook illustration.''}
\label{fig:samples_t2i_fox}
\end{figure}
}

\newcommand{\figSamplesTextToImageCat}{%
\begin{figure}[!htb]
\centering
\includegraphics[width=\linewidth]{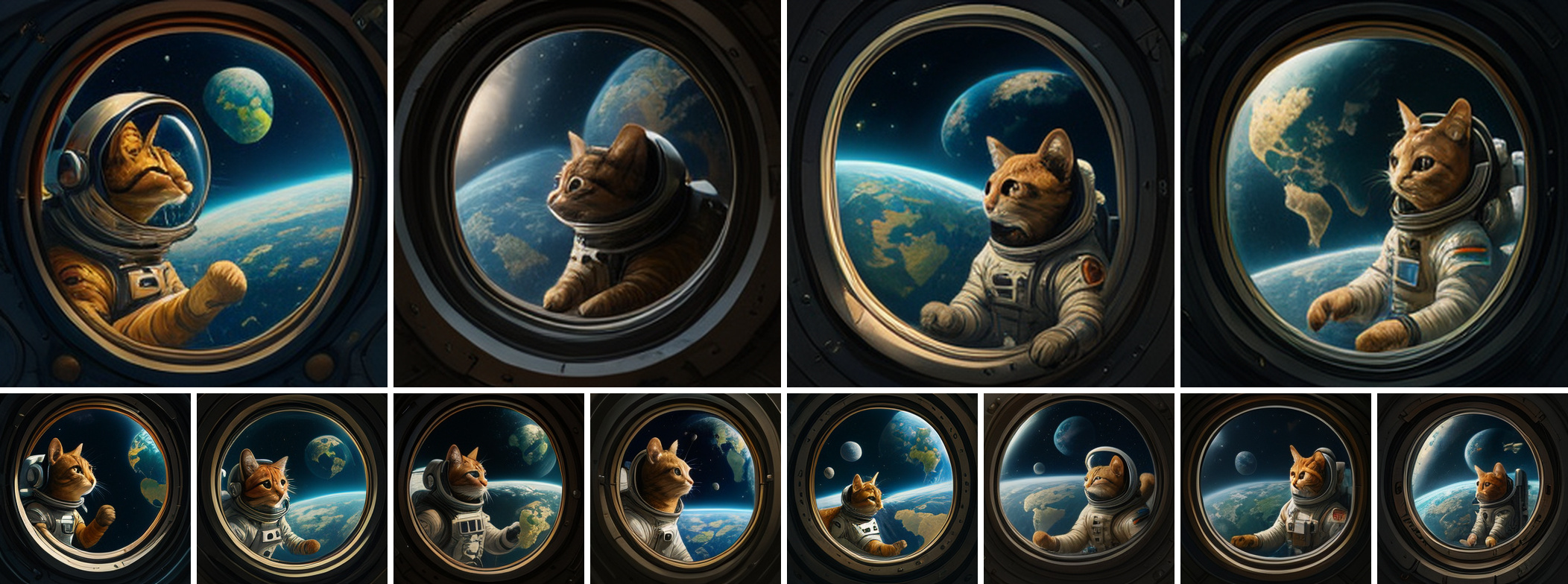}
\caption{``A cat astronaut floating inside a spaceship, looking at the Earth
through a round window.''}
\label{fig:samples_t2i_cat}
\end{figure}
}

\newcommand{\figSamplesTextToImageTeacup}{%
\begin{figure}[!htb]
\centering
\includegraphics[width=\linewidth]{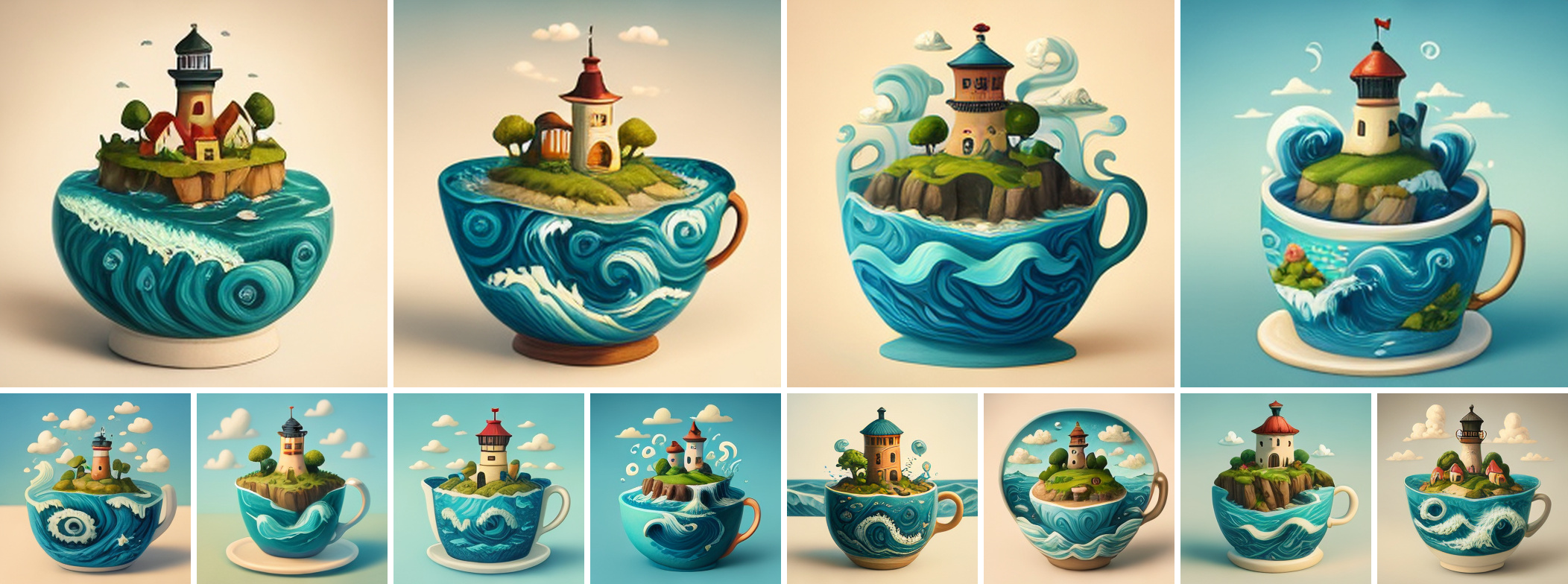}
\caption{``A tiny island with a lighthouse inside a teacup, ocean waves
swirling, whimsical 3D render.''}
\label{fig:samples_t2i_teacup}
\end{figure}
}

\newcommand{\figSamplesTextToImageRetriever}{%
\begin{figure}[!htb]
\centering
\includegraphics[width=\linewidth]{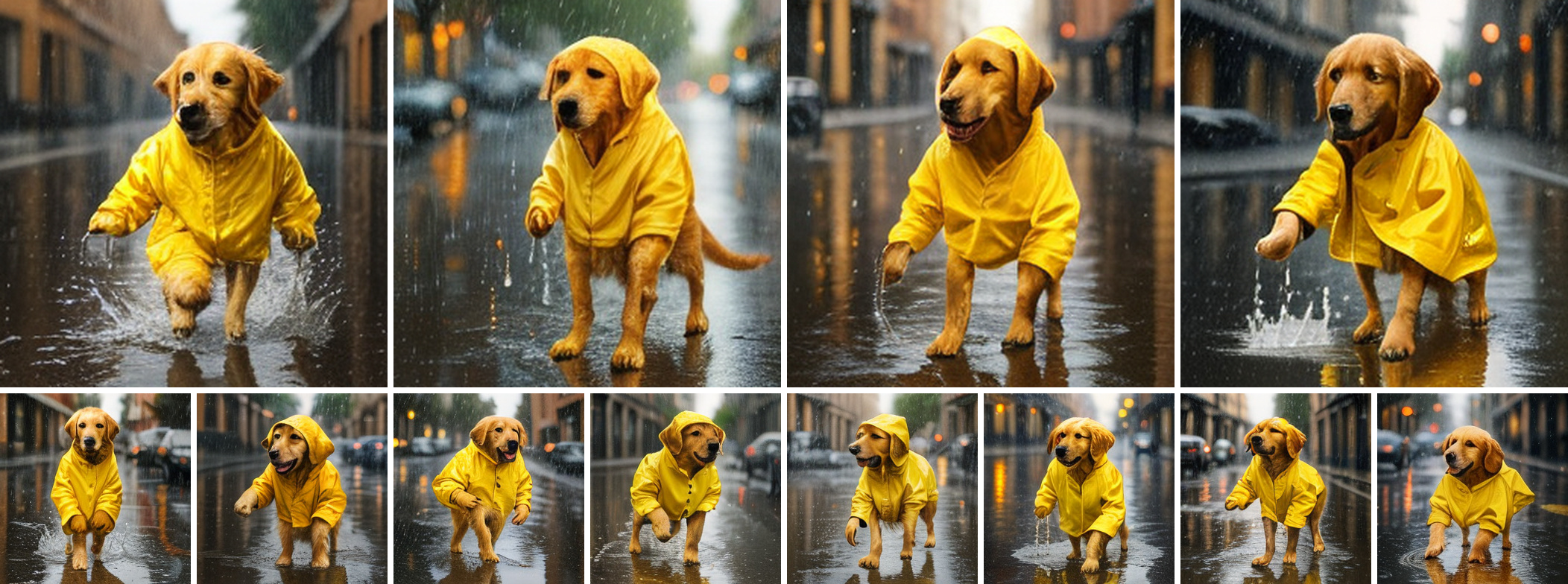}
\caption{``A golden retriever wearing a yellow raincoat, splashing in puddles on
a rainy street.''}
\label{fig:samples_t2i_retriever}
\end{figure}
}

\newcommand{\figSamplesTextToImageTerrarium}{%
\begin{figure}[!htb]
\centering
\includegraphics[width=\linewidth]{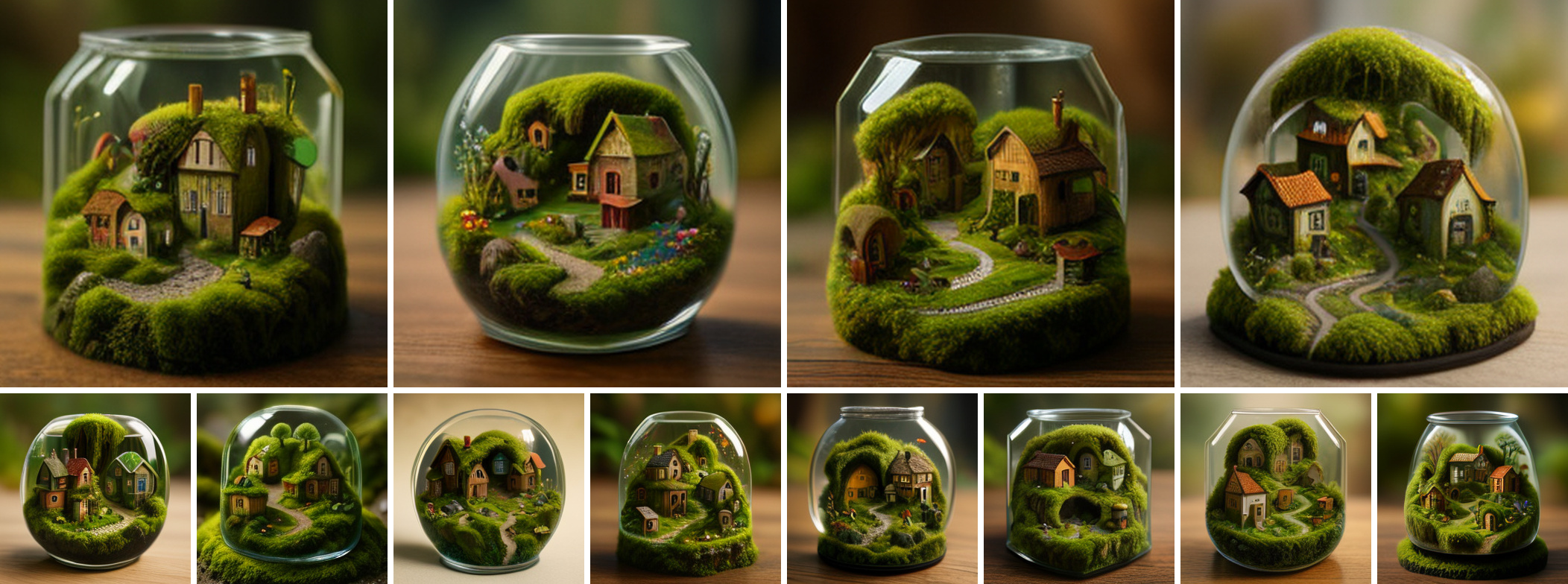}
\caption{``A miniature village inside a glass terrarium with moss, tiny houses
and a winding path, macro photo.''}
\label{fig:samples_t2i_terrarium}
\end{figure}
}

\newcommand{\figSamplesTextToImagePhoenix}{%
\begin{figure}[!htb]
\centering
\includegraphics[width=\linewidth]{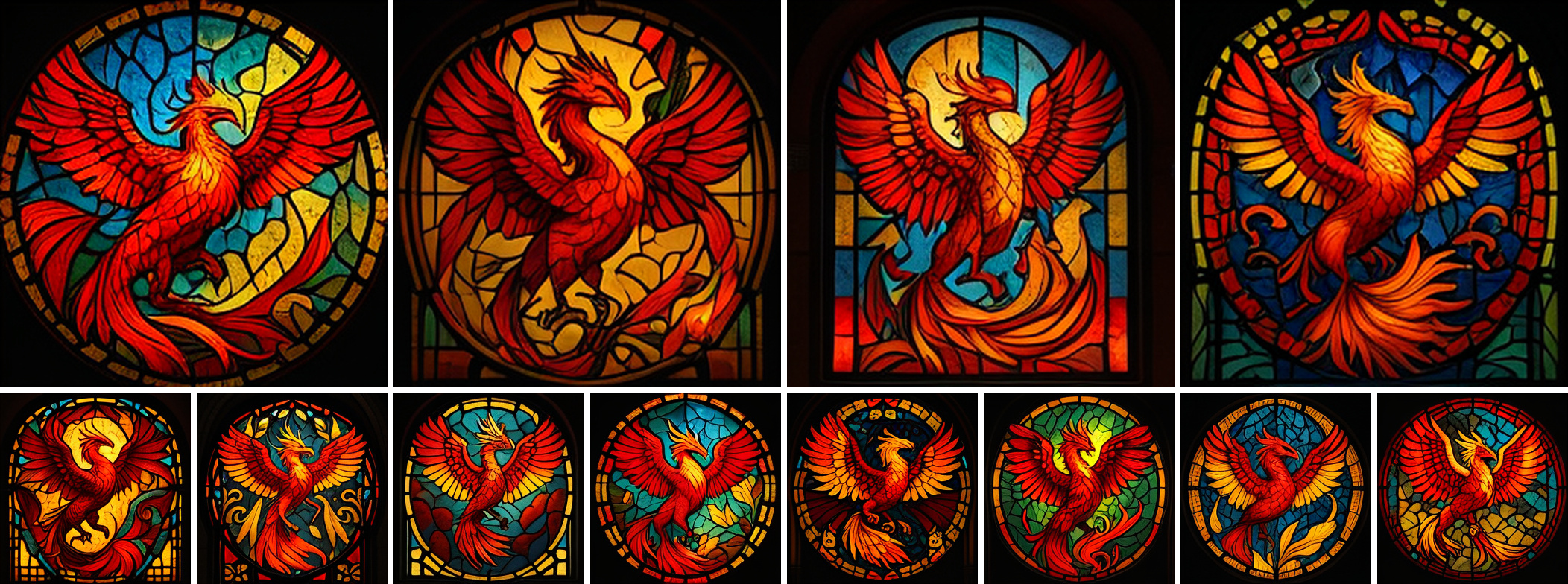}
\caption{``A stained glass window depicting a phoenix rising from flames, vivid
red and gold.''}
\label{fig:samples_t2i_phoenix}
\end{figure}
}

\newcommand{\figSamplesTextToImageCastle}{%
\begin{figure}[!htb]
\centering
\includegraphics[width=\linewidth]{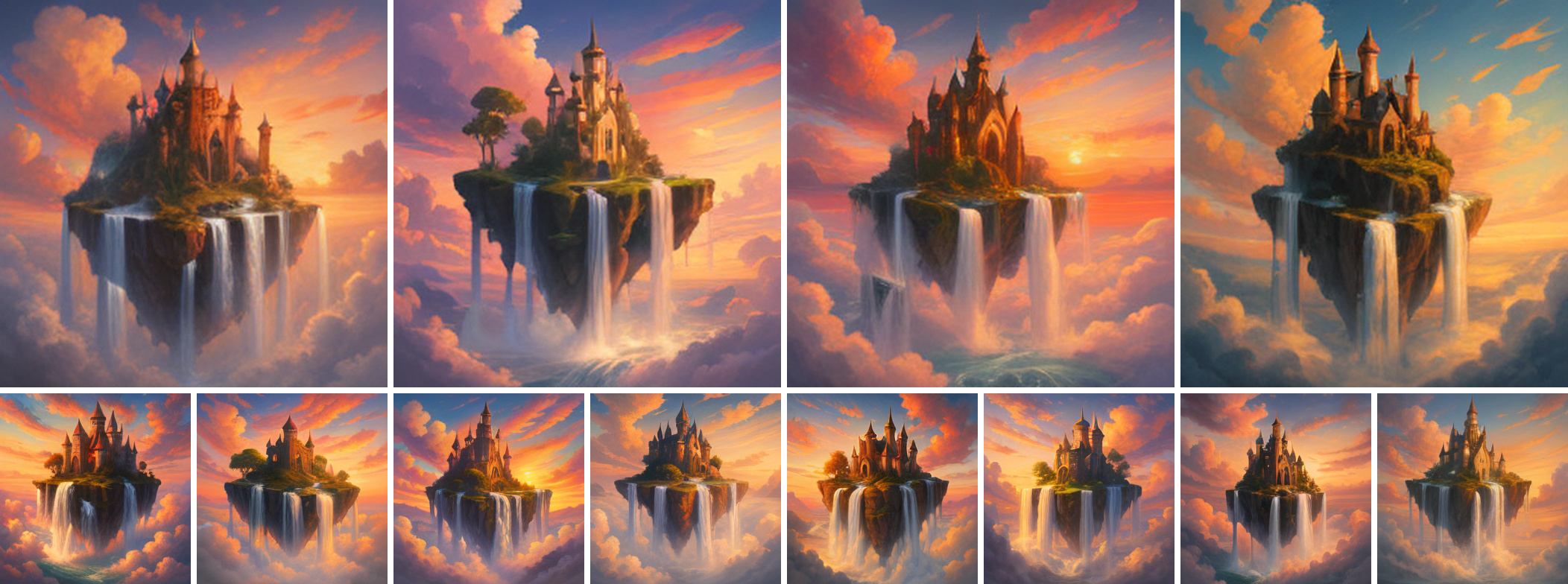}
\caption{``A floating castle above the clouds at sunset, waterfalls pouring off
its edges, fantasy art.''}
\label{fig:samples_t2i_castle}
\end{figure}
}

\newcommand{\figSamplesTextToImageBedroom}{%
\begin{figure}[!htb]
\centering
\includegraphics[width=\linewidth]{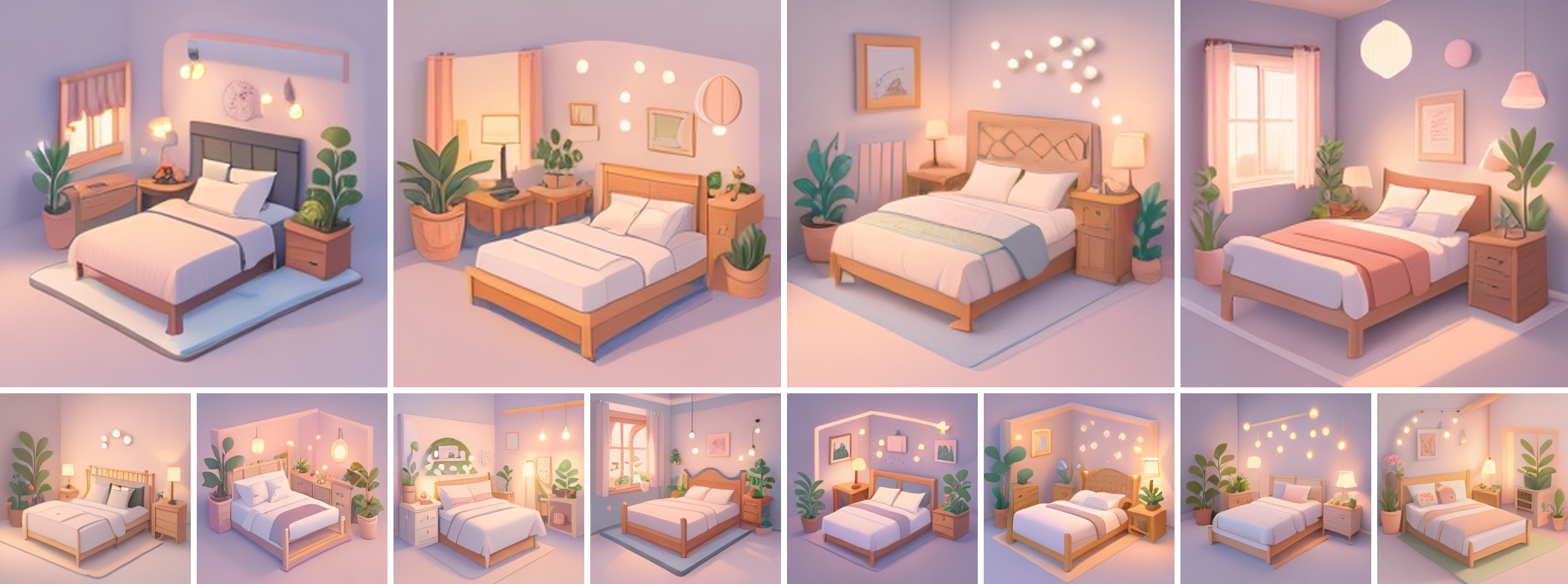}
\caption{``An isometric 3D render of a cozy bedroom with a bed, plants and fairy
lights, pastel colors.''}
\label{fig:samples_t2i_bedroom}
\end{figure}
}

\newcommand{\figSamplesTextToImageCactus}{%
\begin{figure}[!htb]
\centering
\includegraphics[width=\linewidth]{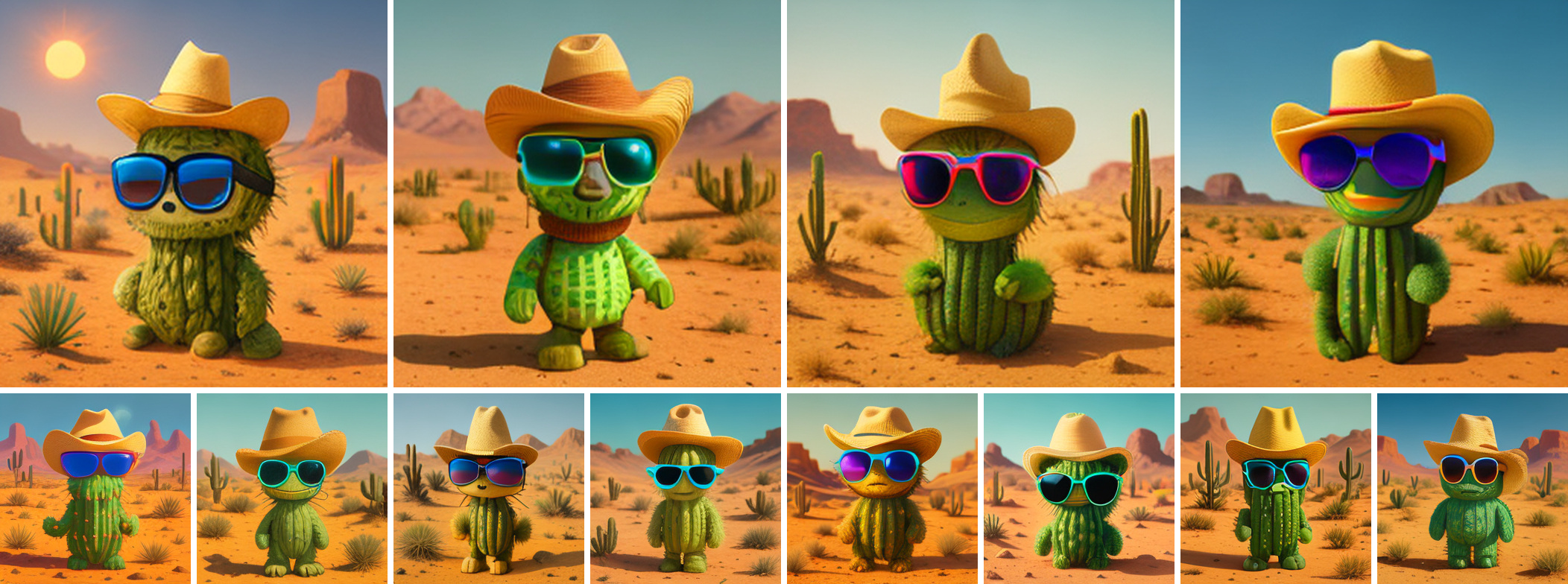}
\caption{``A small cactus wearing a straw hat and neon sunglasses in the Sahara
desert.''}
\label{fig:samples_t2i_cactus}
\end{figure}
}

\newcommand{\tabGenQualityComparison}{%
\begin{table}[t]
\centering
\begin{minipage}[t]{3.226in}
\centering
\captionof{table}{\textbf{RAD outperforms baselines in sample quality across model scales.} FID-50k on ImageNet w/o CFG.}
\label{tab:gen_quality_baselines}
\scriptsize
\setlength{\tabcolsep}{1.5pt}
\renewcommand{\arraystretch}{1.22}
\resizebox{\linewidth}{!}{%
\begin{tabular}{lc cc cc cc}
\toprule
& & \multicolumn{2}{c}{B/2} & \multicolumn{2}{c}{L/2} & \multicolumn{2}{c}{XL/2} \\
\cmidrule(lr){3-4}\cmidrule(lr){5-6}\cmidrule(lr){7-8}
Method & Epochs & \#Params & FID$\downarrow$ & \#Params & FID$\downarrow$ & \#Params
    & FID$\downarrow$ \\
\midrule
\multicolumn{8}{l}{\textit{Diffusion transformers}} \\
DiT~\tabcite{peebles2023dit} & 80 & 130M & 43.5 & 458M & 23.3 & 675M & 19.5 \\
DiT~\tabcite{peebles2023dit} & 1400 & -- & -- & -- & -- & 675M & 9.6 \\
SiT~\tabcite{ma2024sit} & 80 & 130M & 33.0 & 458M & 18.8 & 675M & 17.2 \\
SiT~\tabcite{ma2024sit} & 1400 & -- & -- & -- & -- & 675M & 8.3 \\
\midrule
\multicolumn{8}{l}{\textit{Representation alignment}} \\
DiT + REPA~\tabcite{yu2025repa} & 80 & -- & -- & 458M & 15.6 & 675M & 12.3 \\
REPA~\tabcite{yu2025repa} & 80 & 130M & 24.4 & 458M & 10.0 & 675M & 7.9 \\
REPA~\tabcite{yu2025repa} & 200 & -- & -- & -- & -- & 675M & 6.4 \\
iREPA~\tabcite{singh2026irepa} & 80 & 130M & 21.4 & 458M & 9.4 & 675M & 7.5 \\
ReDi~\tabcite{kouzelis2026boosting} & 80 & -- & -- & 458M & 9.4 & 675M & 7.5 \\
sREPA~\tabcite{xu2026srepa} & 80 & 130M & 20.5 & 458M & 8.7 & 675M & 7.2 \\
HASTE~\tabcite{wang2025haste} & 80 & 130M & 19.6 & 458M & 8.9 & 675M & 7.3 \\
REED~\tabcite{wang2025reed} & 80 & 130M & 19.5 & 458M & 8.9 & 675M & 7.3 \\
TREAD~\tabcite{krause2025tread} & 80 & 130M & 16.0 & 458M & 5.9 & 675M & 4.9 \\
REG~\tabcite{wu2025reg} & 80 & 132M & 15.2 & 460M & 4.6 & 677M & 3.4 \\
REPA-E$^\dagger$~\tabcite{leng2025repae} & 80 & -- & -- & -- & -- & 675M & 4.1 \\
\midrule
\multicolumn{8}{l}{\textit{Representation augmentation}} \\
\rowcolor{radTint} RAD (ours) & 80 & 142M & 7.7 & 472M & 3.9 & 690M & 3.7 \\
\bottomrule
\end{tabular}}
\end{minipage}\hfill
\begin{minipage}[t]{2.524in}
\centering
\captionof{table}{\textbf{RAD can be readily combined with CFG.} FID-50k on ImageNet with CFG.}
\label{tab:gen_quality_cfg}
\scriptsize
\setlength{\tabcolsep}{1.5pt}
\renewcommand{\arraystretch}{1.22}
\resizebox{\linewidth}{!}{%
\begin{tabular}{lcccccc}
\toprule
Method & Ep. & \#Params & FID$\downarrow$ & IS$\uparrow$ & Prec.$\uparrow$ & Rec.$\uparrow$ \\
\midrule
\multicolumn{7}{l}{\textit{Pixel diffusion}} \\
ADM~\tabcite{dhariwal2021adm} & 400 & 554M & 3.94 & 215.8 & 0.83 & 0.53 \\
PixelFlow~\tabcite{chen2025pixelflow} & 320 & 677M & 1.98 & 282.1 & 0.81 & 0.60 \\
SiD2~\tabcite{hoogeboom2025sid2} & 1280 & -- & 1.38 & -- & -- & -- \\
\midrule
\multicolumn{7}{l}{\textit{Latent diffusion}} \\
DiT~\tabcite{peebles2023dit} & 1400 & 675M & 2.27 & 278.2 & 0.83 & 0.57 \\
MaskDiT~\tabcite{zheng2024maskdit} & 1600 & 675M & 2.28 & 276.6 & 0.80 & 0.61 \\
SiT~\tabcite{ma2024sit} & 1400 & 675M & 2.06 & 270.3 & 0.82 & 0.59 \\
\midrule
\multicolumn{7}{l}{\textit{Representation alignment}} \\
REPA~\tabcite{yu2025repa} & 80 & 675M & 2.09 & 279.7 & 0.81 & 0.59 \\
DDT~\tabcite{wang2025ddt} & 80 & 675M & 1.52 & 263.7 & 0.78 & 0.63 \\
REPA-E$^\dagger$~\tabcite{leng2025repae} & 80 & 675M & 1.67 & 266.3 & 0.80 & 0.63 \\
REG~\tabcite{wu2025reg} & 80 & 677M & 1.86 & 321.4 & 0.76 & 0.63 \\
iREPA~\tabcite{singh2026irepa} & 80 & 675M & 2.10 & 280.3 & 0.82 & 0.59 \\
sREPA~\tabcite{xu2026srepa} & 80 & 675M & 2.25 & 257.5 & 0.83 & 0.58 \\
\midrule
\multicolumn{7}{l}{\textit{Representation augmentation}} \\
\rowcolor{radTint} RAD (ours) & 80 & 690M & 1.96 & 252.6 & 0.77 & 0.62 \\
\rowcolor{radTint} RAD (ours) & 200 & 690M & 1.79 & 269.7 & 0.79 & 0.62 \\
\bottomrule
\end{tabular}}
\end{minipage}
\par\vspace{3pt} {\scriptsize\raggedright $^\dagger$REPA-E also fine-tunes the
VAE end-to-end.\par}
\end{table}
}

\newcommand{\tabAblations}{%
\begin{table}[t]
\centering
\begin{minipage}[t]{2.61in}
\centering
\captionof{table}{\textbf{Elucidating the role of different design components in
RAD.} $\Delta$FLOPs is the increase in inference FLOPs added by each stage,
relative to REPA (23.0 GFLOPs). \cmark{}: tuned $\beta_\text{inf}$,
\xmark{}: $\beta_{\mathrm{inf}} = 1$.}
\label{tab:system_analysis}
\fontsize{7.3}{8.8}\selectfont
\setlength{\tabcolsep}{3.05pt}
\renewcommand{\arraystretch}{1.0}
\begin{tabular}{@{}l@{\hspace{4pt}}lcccc@{}}
\toprule
\multicolumn{2}{@{}l}{Training configuration} & $\beta_{\mathrm{inf}}$ & FID$\downarrow$ & $\Delta$ & $\Delta$FLOPs \\
\midrule
\textsc{a} & REPA~\tabcite{yu2025repa} & -- & 24.4 & -- & -- \\
\textsc{b} & + $\rvu_t \sim \gN(.) \rightarrow$ denoiser & \xmark & 12.6 & $-11.8$ & $+8.4\%$ \\
\textsc{c} & + Tuned $\beta_{\mathrm{inf}}$ & \cmark & 10.3 & $-2.3$ & $+0.0\%$ \\
\textsc{d} & + Flow policy & \cmark & 8.7 & $-1.6$ & $+3.4\%$ \\
\textsc{e} & + Control Norm + Mod & \cmark & 8.2 & $-0.6$ & $+0.0\%$ \\
\rowcolor{radTint}
\textsc{f} & + Early inject + p=1.0 & \cmark & \textbf{7.7} & $-0.5$ & $+0.0\%$ \\
\bottomrule
\end{tabular}

\vspace{8pt}
\captionof{table}{\textbf{Stochastic policies outperform deterministic ones in RAD.}
Policy ablation for RAD-B/2 (400K iters). Here $\rvz_0 =
f_{\mathrm{pre}}(\rvx_0)$ and $\epsilon \sim \gN(\vzero, \mI)$. \mbox{250-step} Euler SDE
sampler. Best $\beta_{\mathrm{inf}}$ per row.}
\label{tab:det_vs_stoch}
\fontsize{7.3}{8.8}\selectfont
\setlength{\tabcolsep}{5.1pt}
\renewcommand{\arraystretch}{1.0}
\begin{tabular}{@{}lccc@{}}
\toprule
Policy & Training & Sampling & FID$\downarrow$ \\
 & $q(\rvu_t \mid \rvz_0)$ & $p_\phi(\rvu_t \mid \rvx_t)$ & \\
\midrule
Deterministic & $\alpha_t \rvz_0$ & $\alpha_t \vh_\phi$ & 25.2 \\
\rowcolor{radTint} Gaussian & $\alpha_t \rvz_0 + \beta\sigma_t \epsilon$ & $\alpha_t \vh_\phi + \beta\sigma_t \epsilon$ & 10.3 \\
\rowcolor{radTint} Flow & $\alpha_t \rvz_0 + \beta\sigma_t \epsilon$
    & $\alpha_t \vh_\phi + \beta\sigma_t \vg_\zeta(\epsilon)$ & \textbf{8.7} \\
\bottomrule
\end{tabular}

\end{minipage}\hfill
\begin{minipage}[t]{3.13in}
\centering
\captionof{table}{\textbf{Ablation: Impact of the choice of external encoder,
injection depth and alignment loss weighting.} Sample quality on ImageNet
$256\times256$ of SiT-B/2 models (400K iterations)}
\label{tab:design_choices}
\fontsize{7.3}{8.8}\selectfont
\setlength{\tabcolsep}{2.27pt}
\begin{tabular}{llcc ccccc}
\toprule
Method & Encoder & Depth & $p$ & FID$\downarrow$ & sFID$\downarrow$ & IS$\uparrow$
    & Prec.$\uparrow$ & Rec.$\uparrow$ \\
\midrule
SiT-B/2~\tabcite{ma2024sit} & -- & -- & -- & 33.0 & 6.5 & 43.7 & 0.53 & 0.63 \\
+ REPA~\tabcite{yu2025repa} & DINOv2-B & 4 & -- & 24.4 & 6.4 & 59.9 & 0.59 & 0.65 \\
+ REG~\tabcite{wu2025reg} & DINOv2-B & 4 & -- & 15.2 & 6.9 & 94.6 & 0.64 & 0.63 \\
\midrule
\multicolumn{9}{l}{\textit{+ RAD (ours): encoder}} \\
\rowcolor{radTint} & DINOv2-B & 4 & 0 & 8.5 & 6.1 & 129.5 & 0.76 & 0.53 \\
& DINOv2-L & 4 & 0 & 8.4 & 7.0 & 129.8 & 0.76 & 0.54 \\
& SigLIP-L & 4 & 0 & 8.9 & 7.1 & 133.1 & 0.77 & 0.51 \\
\cmidrule(lr){1-9}
\multicolumn{9}{l}{\textit{+ RAD (ours): injection depth}} \\
\rowcolor{radTint} & DINOv2-B & 2 & 0 & \textbf{8.2} & 7.1 & 141.9 & 0.78 & 0.51 \\
& DINOv2-B & 4 & 0 & 8.5 & 6.1 & 129.5 & 0.76 & 0.53 \\
& DINOv2-B & 6 & 0 & 10.5 & 7.4 & 108.8 & 0.71 & 0.55 \\
& DINOv2-B & 8 & 0 & 15.1 & 7.7 & 85.6 & 0.66 & 0.56 \\
\cmidrule(lr){1-9}
\multicolumn{9}{l}{\textit{+ RAD (ours): alignment loss weighting}} \\
& DINOv2-B & 4 & 0 & 8.5 & 6.1 & 129.5 & 0.76 & 0.53 \\
& DINOv2-B & 4 & 0.5 & 8.7 & 6.5 & 125.7 & 0.75 & 0.53 \\
\rowcolor{radTint} & DINOv2-B & 4 & 1.0 & \textbf{8.3} & 6.7 & 129.0 & 0.75 & 0.53 \\
& DINOv2-B & 4 & 1.5 & 8.6 & 6.5 & 129.6 & 0.75 & 0.52 \\
\bottomrule
\end{tabular}
\end{minipage}
\end{table}
}

\newcommand{\tabHyperparams}{%
\begin{table}[t]
\centering
\small
\setlength{\tabcolsep}{6pt}
\caption{\textbf{RAD Hyperparameters on class-conditional ImageNet
    per scale.} Values spanning all columns are shared.}
\label{tab:hyperparams}
\begin{tabular}{l*{3}{>{\centering\arraybackslash}p{2.1cm}}}
\toprule
& RAD-B/2 & RAD-L/2 & RAD-XL/2 \\
\midrule
\rowcolor{radTint} \multicolumn{4}{l}{\textit{Backbone}} \\
SiT blocks & 12 & 24 & 28 \\
Hidden width & 768 & 1024 & 1152 \\
Attention heads & 12 & 16 & 16 \\
Patch size / tokens & \multicolumn{3}{c}{($2\times2$) / 256} \\
\#Params & 142M & 472M & 690M \\
\midrule
\rowcolor{radTint} \multicolumn{4}{l}{\textit{Representation pathway}} \\
Encoder $f_{\mathrm{pre}}$ & \multicolumn{3}{c}{DINOv2-B/14 (768-dim patch tokens)} \\
Alignment and injection depth $\ell$ & 2/4 & 8 & 8 \\
Control projection & -- & $768 \rightarrow 1024$ & $768 \rightarrow 1152$ \\
Projector $h_\phi$ & \multicolumn{3}{c}{MLP, width 2048, two hidden layers, SiLU} \\
Control Normalization ($\gamma$ / $\delta$) & \multicolumn{3}{c}{0.6 / $10^{-6}$} \\
Control tokens $M$ & \multicolumn{3}{c}{1} \\
Flow policy & \multicolumn{3}{c}{4 affine couplings, width 512, context 256} \\
Alignment weight & \multicolumn{3}{c}{$0.5\,\lambda_t$} \\
Adaptive weighting $p$ & 1.0 & 1.0 & 1.0 \\
Training policy scale $\beta$ & \multicolumn{3}{c}{1.0} \\
\midrule
\rowcolor{radTint} \multicolumn{4}{l}{\textit{Optimization}} \\
Optimizer & \multicolumn{3}{c}{AdamW, lr $10^{-4}$, $(0.9, 0.999)$, no weight decay} \\
Batch size / iterations & \multicolumn{3}{c}{256 / 400K} \\
EMA decay & \multicolumn{3}{c}{0.9999} \\
Label dropout & \multicolumn{3}{c}{0.1} \\
Training time (4 GPUs) & 11 h & 16 h & 22 h \\
\midrule
\rowcolor{radTint} \multicolumn{4}{l}{\textit{Inference}} \\
$\beta_{\mathrm{inf}}$ & 0.7 & 0.8 & 0.8 \\
Extra FLOPs over REPA & 11.8\% & 3.5\% & 2.5\% \\
\bottomrule
\end{tabular}
\end{table}
}

\newcommand{\tabGenQuality}{%
\begin{table}[t]
\centering
\small
\setlength{\tabcolsep}{5pt}
\caption{\textbf{Full ImageNet $256\times256$ results of SiT, REPA, and RAD at
    every training budget} (numbers behind
    Table~\ref{tab:gen_quality_baselines}, Fig.~\ref{fig:overview}b, and
    Fig.~\ref{fig:efficiency}a,b). Baseline numbers are from the respective
    papers. All metrics on 50k samples, Euler--Maruyama SDE sampler with 250
    steps, no guidance. RAD uses $\beta_{\mathrm{inf}} = 0.7$ for B/2 and $0.8$
    for L/2 and XL/2 at every budget.}
\label{tab:gen_quality}
\begin{tabular}{llrccccc}
\toprule
Model & \#Params & Iter. & FID$\downarrow$ & sFID$\downarrow$ & IS$\uparrow$ & Prec.$\uparrow$
    & Rec.$\uparrow$ \\
\midrule
SiT-B/2 & 130M & 400K & 33.0 & 6.46 & 43.7 & 0.53 & 0.63 \\
+ REPA & 130M & 50K & 78.2 & 11.71 & 17.1 & 0.33 & 0.48 \\
+ REPA & 130M & 100K & 49.5 & 7.00 & 27.5 & 0.46 & 0.59 \\
+ REPA & 130M & 200K & 33.2 & 6.68 & 43.7 & 0.54 & 0.63 \\
+ REPA & 130M & 400K & 24.4 & 6.40 & 59.9 & 0.59 & 0.65 \\
\rowcolor{radTint} + RAD (ours) & 142M & 50K & 55.6 & 21.94 & 25.4 & 0.44 & 0.46 \\
\rowcolor{radTint} + RAD (ours) & 142M & 100K & 23.5 & 9.05 & 64.8 & 0.64 & 0.51 \\
\rowcolor{radTint} + RAD (ours) & 142M & 200K & 12.0 & 7.49 & 110.5 & 0.74 & 0.50 \\
\rowcolor{radTint} + RAD (ours) & 142M & 400K & 7.7 & 6.72 & 148.0 & 0.78 & 0.49 \\
\midrule
SiT-L/2 & 458M & 400K & 18.8 & 5.29 & 72.0 & 0.64 & 0.64 \\
+ REPA & 458M & 50K & 55.4 & 24.0 & 23.0 & 0.43 & 0.53 \\
+ REPA & 458M & 100K & 24.1 & 6.25 & 55.7 & 0.62 & 0.60 \\
+ REPA & 458M & 200K & 14.0 & 5.18 & 86.5 & 0.67 & 0.64 \\
+ REPA & 458M & 400K & 10.0 & 5.20 & 109.2 & 0.69 & 0.65 \\
\rowcolor{radTint} + RAD (ours) & 472M & 50K & 28.2 & 16.88 & 48.4 & 0.64 & 0.48 \\
\rowcolor{radTint} + RAD (ours) & 472M & 100K & 9.3 & 8.24 & 120.1 & 0.78 & 0.48 \\
\rowcolor{radTint} + RAD (ours) & 472M & 200K & 5.2 & 7.10 & 169.7 & 0.82 & 0.49 \\
\rowcolor{radTint} + RAD (ours) & 472M & 400K & 3.9 & 5.59 & 203.7 & 0.84 & 0.49 \\
\midrule
SiT-XL/2 & 675M & 7M & 8.3 & 6.32 & 131.7 & 0.68 & 0.67 \\
+ REPA & 675M & 50K & 52.3 & 31.24 & 24.3 & 0.45 & 0.53 \\
+ REPA & 675M & 100K & 19.4 & 6.06 & 67.4 & 0.64 & 0.61 \\
+ REPA & 675M & 200K & 11.1 & 5.05 & 100.4 & 0.69 & 0.64 \\
+ REPA & 675M & 400K & 7.9 & 5.06 & 122.6 & 0.70 & 0.65 \\
+ REPA & 675M & 4M & 5.9 & 5.73 & 157.8 & 0.70 & 0.69 \\
\rowcolor{radTint} + RAD (ours) & 690M & 50K & 24.4 & 19.65 & 56.7 & 0.65 & 0.45 \\
\rowcolor{radTint} + RAD (ours) & 690M & 100K & 7.8 & 6.36 & 134.3 & 0.82 & 0.47 \\
\rowcolor{radTint} + RAD (ours) & 690M & 200K & 4.6 & 6.38 & 185.3 & 0.84 & 0.49 \\
\rowcolor{radTint} + RAD (ours) & 690M & 400K & 3.7 & 5.66 & 217.8 & 0.85 & 0.50 \\
\rowcolor{radTint} + RAD (ours) & 690M & 1M & 3.4 & 5.12 & 239.6 & 0.84 & 0.51 \\
\bottomrule
\end{tabular}
\end{table}
}

\newcommand{\tabGenQualityReg}{%
\begin{table}[t]
\centering
\small
\setlength{\tabcolsep}{5pt}
\caption{\textbf{Full ImageNet $256\times256$ results of REG and REG + RAD at
    every training budget} (numbers behind Table~\ref{tab:reg_convergence}). REG
    numbers are from its paper. All metrics on 50k samples, best
    $\beta_{\mathrm{inf}}$ per row, Euler--Maruyama SDE sampler with 250 steps,
    no guidance.}
\label{tab:gen_quality_reg}
\begin{tabular}{llrccccc}
\toprule
Model & \#Params & Iter. & FID$\downarrow$ & sFID$\downarrow$ & IS$\uparrow$ & Prec.$\uparrow$
    & Rec.$\uparrow$ \\
\midrule
SiT-B/2 & 130M & 400K & 33.0 & 6.46 & 43.7 & 0.53 & 0.63 \\
+ REG & 132M & 50K & 64.7 & 9.47 & 23.2 & 0.40 & 0.51 \\
+ REG & 132M & 100K & 36.1 & 7.74 & 45.5 & 0.53 & 0.61 \\
+ REG & 132M & 200K & 22.1 & 7.19 & 72.2 & 0.60 & 0.63 \\
+ REG & 132M & 400K & 15.2 & 6.69 & 94.6 & 0.64 & 0.63 \\
\rowcolor{radTint} + REG + RAD (ours) & 143M & 50K & 50.5 & 20.54 & 28.1 & 0.45 & 0.51 \\
\rowcolor{radTint} + REG + RAD (ours) & 143M & 100K & 22.4 & 7.11 & 67.3 & 0.64 & 0.54 \\
\rowcolor{radTint} + REG + RAD (ours) & 143M & 200K & 12.9 & 6.60 & 102.8 & 0.70 & 0.54 \\
\rowcolor{radTint} + REG + RAD (ours) & 143M & 400K & 8.9 & 5.76 & 124.6 & 0.74 & 0.53 \\
\midrule
SiT-L/2 & 458M & 400K & 18.8 & 5.29 & 72.0 & 0.64 & 0.64 \\
+ REG & 460M & 50K & 30.1 & 8.92 & 52.6 & 0.58 & 0.57 \\
+ REG & 460M & 100K & 11.4 & 5.36 & 108.8 & 0.70 & 0.60 \\
+ REG & 460M & 200K & 6.6 & 5.16 & 145.4 & 0.73 & 0.63 \\
+ REG & 460M & 400K & 4.6 & 5.21 & 167.6 & 0.75 & 0.63 \\
\rowcolor{radTint} + REG + RAD (ours) & 474M & 50K & 23.7 & 11.03 & 61.9 & 0.62 & 0.53 \\
\rowcolor{radTint} + REG + RAD (ours) & 474M & 100K & 7.8 & 6.13 & 136.1 & 0.77 & 0.52 \\
\rowcolor{radTint} + REG + RAD (ours) & 474M & 200K & 4.6 & 5.81 & 183.6 & 0.80 & 0.52 \\
\rowcolor{radTint} + REG + RAD (ours) & 474M & 400K & 3.9 & 4.78 & 211.0 & 0.84 & 0.52 \\
\bottomrule
\end{tabular}
\end{table}
}

\newcommand{\tabGuidedHyperparams}{%
\begin{table}[t]
\centering
\small
\setlength{\tabcolsep}{6pt}
\caption{\textbf{Guided sampling hyperparameters for RAD-XL/2.} Values spanning
    both columns are shared.}
\label{tab:guided_hyperparams}
\begin{tabular}{l*{2}{>{\centering\arraybackslash}p{3.2cm}}}
\toprule
& 400K iterations & 1M iterations \\
\midrule
Sampler & \multicolumn{2}{c}{Euler--Maruyama SDE} \\
Sampling steps & \multicolumn{2}{c}{50} \\
Inference policy scale $\beta_{\mathrm{inf}}$ & \multicolumn{2}{c}{0.8} \\
Policy noise & \multicolumn{2}{c}{fixed} \\
Noise coupling $\rho$ & 0.12 & 0.10 \\
CFG scale $s$ & 3.25 & 2.5 \\
CFG controls $\rvu_t$ & \multicolumn{2}{c}{shared} \\
Guidance interval & \multicolumn{2}{c}{$t \in [0, 0.75]$} \\
Autoguidance weak model & \multicolumn{2}{c}{same run, 100K checkpoint} \\
Autoguidance scale $w$ & \multicolumn{2}{c}{1.5} \\
Weak model $\beta_{\mathrm{inf}}$ & \multicolumn{2}{c}{0.8} \\
Seed & \multicolumn{2}{c}{0} \\
\bottomrule
\end{tabular}
\end{table}
}

\newcommand{\tabGuidedAblation}{%
\begin{table}[!htb]
\centering
\small
\caption{\textbf{Noise coupling helps both CFG and autoguidance.} RAD-XL/2
    (400K iterations), Euler--Maruyama SDE sampler with 50 steps,
    $\beta_{\mathrm{inf}} = 0.8$. CFG and autoguidance (AG) are applied to
    independent ($\rho = 0$) and coupled ($\rho = 0.1$) policy noise;
    CFG uses $s = 3.0$ and AG $w = 2.0$.}
\label{tab:guided_ablation}
\setlength{\tabcolsep}{3pt}
\begin{tabular}{lcccc@{\hspace{8pt}}cccc}
\toprule
& \multicolumn{4}{c}{Independent ($\rho = 0$)} &
    \multicolumn{4}{c}{Noise-coupled ($\rho = 0.1$)} \\
\cmidrule(lr){2-5}\cmidrule(lr){6-9}
Guidance & FID$\downarrow$ & IS$\uparrow$ & Prec.$\uparrow$ & Rec.$\uparrow$ &
    FID$\downarrow$ & IS$\uparrow$ & Prec.$\uparrow$ & Rec.$\uparrow$ \\
\midrule
None & 3.97 & 206.5 & 0.84 & 0.51 & 4.31 & 160.5 & 0.77 & 0.59 \\
CFG~\tabcite{ho2022cfg} & 3.51 & 296.8 & 0.87 & 0.49 & 2.18 & 241.2 & 0.80 & 0.58 \\
AG~\tabcite{karras2024autoguidance} & 3.12 & 247.7 & 0.85 & 0.52 & 2.39 & 192.7 & 0.78 & 0.60 \\
\bottomrule
\end{tabular}
\end{table}
}

\newcommand{\tabConvergenceFid}{%
\noindent\begin{minipage}{\linewidth}
\begin{center}
\small
\setlength{\tabcolsep}{5pt}
\captionof{table}{\textbf{Training convergence} (Fig.~\ref{fig:overview}b and
Fig.~\ref{fig:efficiency}a). FID-50k on ImageNet $256\times256$ against training
iterations. Baseline numbers are from the respective papers. A dash marks a
budget that is not reported in the respective paper (baselines) or not
evaluated (RAD). Euler--Maruyama SDE sampler with 250 steps,
no guidance.}
\label{tab:convergence_fid}
\begin{tabular}{lcccccccc}
\toprule
Model & 50K & 100K & 150K & 200K & 400K & 1M & 4M & 7M \\
\midrule
SiT-XL/2~\tabcite{ma2024sit} & -- & -- & -- & -- & 17.2 & -- & -- & 8.3 \\
REPA-B/2~\tabcite{yu2025repa} & 78.2 & 49.5 & -- & 33.2 & 24.4 & -- & -- & -- \\
REPA-XL/2~\tabcite{yu2025repa} & 52.3 & 19.4 & -- & 11.1 & 7.9 & 6.4 & 5.9 & --
    \\
\rowcolor{radTint} RAD-B/2 (ours) & 55.6 & 23.5 & -- & 12.0 & 7.7 & -- & -- & --
    \\
\rowcolor{radTint} RAD-XL/2 (ours) & 24.4 & 7.8 & 5.7 & 4.6 & 3.7 & 3.4 & -- & --
    \\
\bottomrule
\end{tabular}
\end{center}
\end{minipage}
}

\newcommand{\tabDataEfficiency}{%
\noindent\begin{minipage}{\linewidth}
\begin{center}
\small
\setlength{\tabcolsep}{6pt}
\captionof{table}{\textbf{Data efficiency} (Fig.~\ref{fig:efficiency}c). FID-50k
of SiT-B/2 models trained for 400K iterations on a random fraction of ImageNet
$256\times256$, best checkpoint of each run. Euler ODE sampler with 50 steps, no
guidance.}
\label{tab:data_efficiency}
\begin{tabular}{lcccc}
\toprule
Fraction of ImageNet & 3\% & 10\% & 30\% & 100\% \\
\midrule
REPA-B/2~\tabcite{yu2025repa} & 33.8 & 18.8 & 22.4 & 22.9 \\
\rowcolor{radTint} RAD-B/2 (ours) & \textbf{32.6} & \textbf{13.0} & \textbf{8.9}
    & \textbf{7.8} \\
\bottomrule
\end{tabular}
\end{center}
\end{minipage}
}

\newcommand{\tabSamplingEfficiency}{%
\noindent\begin{minipage}{\linewidth}
\begin{center}
\small
\setlength{\tabcolsep}{6pt}
\captionof{table}{\textbf{Sampling efficiency against REPA}
    (Fig.~\ref{fig:sampling_nfe}). FID-50k on ImageNet
    $256\times256$ against the number of Euler ODE steps for RAD-XL/2 trained
    for 150K iterations and the official REPA-XL/2 checkpoint trained for 4M
    iterations without CFG. With the 250-step Euler--Maruyama SDE sampler the
    same models reach 5.7 and 5.9, respectively.}
\label{tab:sampling_efficiency}
\begin{tabular}{lcccccccc}
\toprule
Model & 5 & 10 & 15 & 20 & 25 & 50 & 100 & 200 \\
\midrule
REPA-XL/2 (4M)~\tabcite{yu2025repa} & 62.3 & 21.9 & 15.0 & 12.2 & 10.5 & 8.2 &
    7.4 & 7.2 \\
\rowcolor{radTint} RAD-XL/2 (150K, ours) & \textbf{23.7} & \textbf{9.1} &
    \textbf{7.3} & \textbf{6.7} & \textbf{6.4} & \textbf{6.2} & -- & -- \\
\bottomrule
\end{tabular}
\end{center}
\end{minipage}
}

\newcommand{\tabSamplingEfficiencySde}{%
\noindent\begin{minipage}{\linewidth}
\begin{center}
\small
\setlength{\tabcolsep}{6pt}
\captionof{table}{\textbf{Stochastic sampling efficiency against REPA}
    (Fig.~\ref{fig:sampling_nfe_sde}). FID-50k on ImageNet $256\times256$
    against the number of Euler--Maruyama SDE steps for RAD-XL/2 (150K
    iterations) and the official REPA-XL/2 (4M iterations) w/o CFG.}
\label{tab:sampling_efficiency_sde}
\begin{tabular}{lccccc}
\toprule
Model & 10 & 25 & 50 & 100 & 250 \\
\midrule
REPA-XL/2 (4M)~\tabcite{yu2025repa} & 101.0 & 22.0 & 10.3 & 7.4 & 5.9 \\
\rowcolor{radTint} RAD-XL/2 (150K, ours) & \textbf{46.9} & \textbf{7.9} &
    \textbf{5.9} & \textbf{5.4} & \textbf{5.7} \\
\bottomrule
\end{tabular}
\end{center}
\end{minipage}
}

\newcommand{\tabPolicyNoise}{%
\noindent\begin{minipage}{\linewidth}
\begin{center}
\small
\setlength{\tabcolsep}{6pt}
\captionof{table}{\textbf{ODE samplers and policy noise}
    (Fig.~\ref{fig:samplers}). FID-50k on ImageNet
    $256\times256$ against the number of sampling steps for RAD-XL/2 (400K
    iterations) with the Euler ODE sampler and UniPC~\citep{zhao2023unipc}
    of order 2 and 3 without CFG, drawing fresh policy noise at every step or
    keeping it fixed over the trajectory. The best value per step budget is in
    bold.}
\label{tab:policy_noise}
\begin{tabular}{llcccc}
\toprule
Sampler & Policy noise & 5 & 10 & 25 & 50 \\
\midrule
Euler & Fresh & 20.6 & 6.8 & 4.1 & 4.0 \\
 & Fixed & 20.0 & 6.5 & 4.2 & 3.9 \\
\midrule
UniPC, order 2 & Fresh & 6.5 & 4.5 & \textbf{3.7} & \textbf{3.8} \\
 & Fixed & 6.7 & \textbf{4.4} & 3.8 & \textbf{3.8} \\
\midrule
UniPC, order 3 & Fresh & \textbf{6.4} & 4.5 & 3.8 & \textbf{3.8} \\
 & Fixed & \textbf{6.4} & \textbf{4.4} & 3.9 & 3.9 \\
\bottomrule
\end{tabular}
\end{center}
\end{minipage}
}

\newcommand{\tabBetaInf}{%
\noindent\begin{minipage}{\linewidth}
\begin{center}
\small
\setlength{\tabcolsep}{6pt}
\captionof{table}{\textbf{Inference policy scale} (Figs.~\ref{fig:beta_inf_fid}
    and~\ref{fig:beta_inf}).
    FID-50k, precision, and recall on ImageNet $256\times256$ of RAD-XL/2
    (400K iterations, $\beta = 1$) against $\beta_{\mathrm{inf}}$. Euler ODE
    sampler with 50 steps, no guidance. Best per column in bold.}
\label{tab:beta_inf}
\begin{tabular}{lccc}
\toprule
$\beta_{\mathrm{inf}}$ & FID$\downarrow$ & Prec.$\uparrow$ & Rec.$\uparrow$ \\
\midrule
0.2 & 45.9 & 0.37 & \textbf{0.59} \\
0.4 & 8.1 & 0.72 & 0.55 \\
0.6 & 4.3 & 0.82 & 0.51 \\
0.75 & 4.2 & 0.84 & 0.49 \\
0.8 & \textbf{4.0} & \textbf{0.85} & 0.50 \\
1.0 (train) & 4.5 & 0.84 & 0.50 \\
\bottomrule
\end{tabular}
\end{center}
\end{minipage}
}

\newcommand{\tabBetaTrain}{%
\noindent\begin{minipage}{\linewidth}
\begin{center}
\small
\setlength{\tabcolsep}{6pt}
\captionof{table}{\textbf{Training policy scale} (Fig.~\ref{fig:beta_train}).
    FID-50k on ImageNet $256\times256$ of RAD-B/2 (400K iterations) with the
    Gaussian policy (config \textsc{b} in Table~\ref{tab:system_analysis}),
    policy scale $\beta$, and sampled with $\beta_{\mathrm{inf}} = \beta$.
    Euler--Maruyama SDE sampler with 250 steps, no guidance. The best value is
    in bold.}
\label{tab:beta_train}
\begin{tabular}{lc}
\toprule
$\beta$ & FID-50k$\downarrow$ \\
\midrule
0.0 & 25.2 \\
0.5 & \textbf{12.2} \\
1.0 & 12.6 \\
2.0 & 16.2 \\
\bottomrule
\end{tabular}
\end{center}
\end{minipage}
}

\newcommand{\tabLinearProbeBlocks}{%
\noindent\begin{minipage}{\linewidth}
\begin{center}
\small
\setlength{\tabcolsep}{4pt}
\captionof{table}{\textbf{Linear probing per block}
(Fig.~\ref{fig:linear_probe}a). ImageNet top-1 accuracy (\%) of a linear
classifier on the hidden state at block $k$ of REPA-B/2 and RAD-B/2 trained for
400K iterations, for a noised input at diffusion time $t$. Both models are
aligned at block 4, after which RAD injects $\rvu_t$ (sampled with
$\beta_{\mathrm{inf}} = 0.7$). The better value per block and $t$ is in bold.}
\label{tab:linear_probe_blocks}
\begin{tabular}{c c>{\columncolor{radTint}}c c>{\columncolor{radTint}}c c>{\columncolor{radTint}}c c>{\columncolor{radTint}}c}
\toprule
& \multicolumn{2}{c}{$t=0.1$} & \multicolumn{2}{c}{$t=0.25$} &
    \multicolumn{2}{c}{$t=0.5$}
    & \multicolumn{2}{c}{$t=0.75$} \\
\cmidrule(lr){2-3}\cmidrule(lr){4-5}\cmidrule(lr){6-7}\cmidrule(lr){8-9}
Block & REPA & RAD & REPA & RAD & REPA & RAD & REPA & RAD \\
\midrule
4  & \textbf{63.0} & 62.7 & \textbf{59.1} & 59.0 & 37.2 & \textbf{37.8} & 8.3 &
    \textbf{8.6} \\
5  & 60.9 & \textbf{63.4} & 57.4 & \textbf{59.2} & 36.2 & \textbf{36.4} &
    \textbf{8.3} & 6.6 \\
6  & 58.3 & \textbf{62.1} & 54.9 & \textbf{58.1} & 35.4 & \textbf{36.0} &
    \textbf{8.2} & 6.8 \\
8  & 52.3 & \textbf{56.7} & 50.2 & \textbf{53.9} & 33.8 & \textbf{34.6} &
    \textbf{8.1} & 6.8 \\
10 & 46.1 & \textbf{48.8} & 44.6 & \textbf{47.6} & 31.7 & \textbf{32.4} &
    \textbf{8.2} & 6.9 \\
12 & 40.7 & \textbf{43.3} & 40.2 & \textbf{43.3} & 29.3 & \textbf{30.5} &
    \textbf{7.8} & 6.9 \\
\bottomrule
\end{tabular}
\end{center}
\end{minipage}
}

\newcommand{\tabLinearProbeScale}{%
\noindent\begin{minipage}{\linewidth}
\begin{center}
\small
\setlength{\tabcolsep}{6pt}
\captionof{table}{\textbf{Linear probing across model scale}
(Fig.~\ref{fig:linear_probe}b). Best ImageNet top-1 accuracy (\%) over all
blocks of RAD-B/2 and RAD-XL/2 trained for 400K iterations, at each diffusion
time $t$, with the block that attains it in parentheses. B/2 is probed with
$\beta_{\mathrm{inf}} = 0.7$ and XL/2 with $\beta_{\mathrm{inf}} = 0.8$.}
\label{tab:linear_probe_scale}
\begin{tabular}{lcccc}
\toprule
Model & $t=0.1$ & $t=0.25$ & $t=0.5$ & $t=0.75$ \\
\midrule
RAD-B/2 & 63.4 (5) & 59.2 (5) & 37.8 (4) & 8.6 (4) \\
RAD-XL/2 & \textbf{71.7} (9) & \textbf{68.1} (9) & \textbf{46.1} (8) &
    \textbf{9.5} (8) \\
\bottomrule
\end{tabular}
\end{center}
\end{minipage}
}

\begin{document}

\maketitle

\begin{abstract}
Diffusion models achieve state-of-the-art sample quality but lack an inherent
mechanism to learn explicit semantic representations of the data. We propose
\emph{Representation-Augmented Diffusions} (RAD), a new class of diffusion
models that augments the denoising process with semantic representations as
latent variables. We derive a variational lower bound for this model that
recovers existing representation alignment methods as special cases and
motivates a practical training objective. Our framework reveals a
simple design choice that existing representation alignment methods leave
unused: \emph{feeding the aligned representation back into the denoiser}. This
idea leads to order-of-magnitude reductions in the training iterations and
sampling steps needed to reach a given sample quality, along with improved data
efficiency, over standard diffusion. For instance, on class-conditional ImageNet
256 $\times$ 256 without classifier-free guidance, RAD matches the FID of SiT
(trained for 7M iterations) with $70\times$ fewer training iterations while
preserving discriminative properties, demonstrating that strong representations
can be learned without compromising sample quality. Across model scales, RAD
outperforms or performs comparably to state-of-the-art representation alignment
methods. RAD is also particularly effective with few sampling steps, and
transfers to text-conditioned image generation. Overall, RAD is a strong
alternative for efficient diffusion model training. Our code and models 
are available at \url{https://github.com/mandt-lab/RAD}

\end{abstract}

\figOverview
\figGraphicalModel

\section{Introduction}
\label{sec:intro}

Diffusion models \citep{ho2020denoising,song2021score} and related families
\citep{lipman2023flow,albergo2023stochastic} achieve state-of-the-art sample
quality by iteratively converting noise into data. Beyond sample quality,
however, many downstream tasks need compact representations of the data. World
models, for instance, plan and act in a learned latent state rather than in the
pixel space \citep{hafner2023mastering,zhou2025dinowm,
assran2025vjepa2,maes2026leworldmodelstableendtoendjointembedding}. This raises
a natural question: \emph{Does denoising also yield useful representations of
the data?} The answer is unclear for two reasons. First, diffusion models are
trained only to denoise, and although their features carry semantic information
\citep{xiang2023ddae}, they lag well behind those of self-supervised encoders
\citep{yu2025repa}. Second, unlike GANs and VAEs
\citep{goodfellow2014gan,kingma2014vae}, which generate from a semantic latent
space by construction, diffusion models lack such a compact latent space as
every intermediate latent is a noisier version of the data. Moreover, semantic
features stay implicit within the denoiser.

A recent line of work based on representation alignment addresses the first
problem by proposing to align the representations of an intermediate layer in
the denoiser with those of a frozen pretrained encoder, as in REPA
\citep{yu2025repa}, or toward the model's own deeper features
\citep{jiang2025sra}. These methods show that good representations substantially
accelerate the training of state-of-the-art diffusion transformers
\citep{peebles2023dit,ma2024sit}. However, the representation serves only as a
training-time regularizer and is discarded at generation, which leaves the
second problem unresolved. REG \citep{wu2025reg} denoises a semantic token
jointly with the image latents so that it guides generation, and REED
\citep{wang2025reed} formulates representation guidance variationally but drops
the explicit conditioning in practice (App.~\ref{sec:related}). Conversely,
another line of work adds a latent variable that steers generation by combining
diffusion models with auxiliary generative models such as a VAE
\citep{pandey2022diffusevae,preechakul2022diffae}. These methods address the
second problem but are often learned from a pixel reconstruction objective that
favors low-level detail over semantics. We discuss additional related work in
App.~\ref{sec:related}.

To address both problems, we make the representation a latent variable of the
diffusion model. Intuitively, at every denoising step, the model first predicts
a representation from the current noisy state and then takes a denoising step
conditioned on it (Fig.~\ref{fig:graphical_model}). We derive a variational
lower bound on the data likelihood, in which the alignment loss appears as a KL
regularizer rather than a heuristic add-on and existing methods in
representation alignment are recovered as a special case (Sec.~\ref{sec:param}
and App.~\ref{app:prior_work}) and use it to motivate a practical training
objective. Our framework reveals a simple design choice that representation
alignment misses: \emph{feeding the learned representation back into the
denoiser} (Fig.~\ref{fig:overview}). We call the resulting framework
\emph{Representation-Augmented Diffusions} (RAD). To summarize, our core
contributions are:
\begin{itemize}
\item \textbf{A principled framework.} We treat semantic representations as a
    latent variable of the diffusion model and derive a variational lower bound
    in which representation alignment appears as a KL term, recovering existing
    alignment methods as special cases (Sec.~\ref{sec:param},
    App.~\ref{app:prior_work}). The bound exposes a simple change, conditioning
    the denoiser on the learned representations, with a large effect on sample
    quality (Table~\ref{tab:system_analysis}).
\item \textbf{Order-of-magnitude faster training.} On ImageNet $256 \times 256$,
    RAD matches SiT (trained for 7M steps) on sample quality in just 100K
    training steps ($70\times$ fewer, Fig.~\ref{fig:overview}b) and matches or
    outperforms baselines on sample quality with $4.8\times$ fewer parameters
    (Fig.~\ref{fig:efficiency}b).
\item \textbf{Fewer sampling steps.} RAD induces more expressive denoising
     policies, and needs ${\sim}10\times$ fewer sampling steps for equal FID
     (Sec.~\ref{sec:sampling_efficiency}) compared to baselines. With guidance,
     RAD reaches FID 1.96 comparable to state-of-the-art methods
     (Table~\ref{tab:gen_quality_cfg}).
\item \textbf{Steerable and semantic representations.} RAD allows smooth
    interpolation between samples via the learned representations
    (Fig.~\ref{fig:interpolation}) and preserves linearly decodable semantics
    deeper into the denoiser, with up to 4.4 points higher linear-probe accuracy
    in later blocks (Sec.~\ref{sec:linear_probe}).
\item \textbf{A drop-in addition.} RAD composes with other alignment methods,
    doubling the convergence speed of REG \citep{wu2025reg}, and transfers to
    text-to-image generation (Sec.~\ref{sec:generalization}).
\end{itemize}

\section{Representation-Augmented Diffusions}
\label{sec:method}
\paragraph{Intuition}
We introduce Representation-Augmented Diffusions (RAD), a class of diffusion
models in which a semantic representation guides the generative process itself.
The core idea is simple: at every denoising step, the model first predicts a
representation from an intermediate noisy state, and then conditions the
denoiser on it. This is in contrast to recent representation alignment methods
which predict representations but never condition the denoiser on them. During
training, the target for learning such semantic representations can be provided
by additional supervision. At inference, the model generates these
representations from its own features, and no external supervision is required.
Next, we discuss the core RAD modeling paradigm.

\subsection{Denoising meets representations}
\label{sec:graph}
Our key modeling choice is to augment the denoising process in standard
diffusion models with a sequence of latent representations $\rvu_{1:T}$, one for
every denoising step (see Fig.~\ref{fig:graphical_model}).

\textbf{Generative process.} Starting from noise $\rvx_T \sim
p\left(\rvx_T\right) = \gN(\vzero, \mI)$, each denoising step first samples a
latent representation $\rvu_t$ from a learned \emph{representation policy}
$p_\phi(\rvu_t \mid \rvx_t)$. Consequently, the cleaner state $\rvx_{t-1}$ is
then sampled from a \emph{denoising policy} $p_\theta(\rvx_{t-1} \mid \rvx_t,
\rvu_t)$ conditioned on the latent representation $\rvu_t$. Formally,
\begin{equation}
    \label{eq:gen_process}
    p_{\theta, \phi}\left(\rvx_{0: T}, \rvu_{1: T}\right)
        = p\left(\rvx_T\right) \prod_{t=1}^T p_\phi\left(\rvu_t \mid
            \rvx_t\right)
        p_\theta\left(\rvx_{t-1} \mid \rvx_t, \rvu_t\right).
\end{equation}
The representation $\rvu_t$ therefore acts as a hierarchical latent variable of
the transition \citep{ranganath2016hierarchical}: the model first samples a
semantic representation of the data and then takes a denoising step that is
consistent with it. Consequently, we denote this class of diffusion models as
\emph{Representation-Augmented Diffusions} (RAD). It is worth noting that
marginalizing over $\rvu_t$ yields the transition,
\begin{equation}
    p_{\theta, \phi}\left(\rvx_{t-1} \mid \rvx_t\right) = \int
        p_\theta\left(\rvx_{t-1} \mid \rvx_t,
        \rvu_t\right) p_\phi\left(\rvu_t \mid \rvx_t\right) \mathrm{d}\rvu_t
        = \E_{\rvu_t \sim p_\phi}\left[ p_\theta\left(\rvx_{t-1} \mid \rvx_t,
            \rvu_t\right) \right],
    \label{eq:marginalized_transition}
\end{equation}
which intuitively, is a continuous mixture that is more expressive than the
Gaussian modeling choice in standard diffusion models. Since denoising is
iterative and can be compute intensive, this property is specifically favorable
for few-step sampling in diffusion models where the Gaussianity assumption over
denoising breaks. We empirically demonstrate this property later where RAD
strikes much better balance between sample quality vs inference compute (see
Sec.~\ref{sec:sampling_efficiency}). Moreover, when the denoising policy
$p_\theta(. \mid \rvx_t, \rvu_t)$ ignores $\rvu_t$, the transition in Eq.
\ref{eq:marginalized_transition} reduces to a Gaussian distribution reducing the
setup to standard diffusion models.

\textbf{Noising process.} Next, we define the forward noising (a.k.a inference)
process as,
\begin{equation}
    \label{eq:inf_process}
    q_{\theta^{\prime}}\left(\rvx_{1: T}, \rvu_{1: T} \mid \rvx_0\right)
        = \prod_{t=1}^T q\left(\rvx_t \mid \rvx_{t-1}\right)
            q_{\theta^{\prime}}\left(\rvu_t \mid
        \rvx_{\leq t}\right).
\end{equation}
Intuitively, at each timestep t, we infer the next noisy state $\rvx_t$ from the
cleaner state $\rvx_{t-1}$ via the noising kernel $q(.\mid \rvx_t)$ , and then
infer a representation $\rvu_t$ from the entire trajectory, $\rvx_{\leq t}$ up
to time t using the \emph{reference policy} $q_{\theta^\prime}(. \mid \rvx_{\leq
t})$. The representation $\rvu_t$ is therefore inferred from information that is
unavailable to the generative model at step $t$, i.e. the cleaner trajectory and
thus can serve as an external supervision for the representation policy
$p_\phi(. \mid \rvx_t)$ (more on this in Sec.~\ref{sec:param}). In this sense,
the reference policy acts as a teacher with privileged access to the cleaner
trajectory, while the representation policy is a student that must predict
$\rvu_t$ from $\rvx_t$ alone and replaces the teacher at inference. With the
modeling choices in place, we formulate a tractable variational bound for RAD
next.

\subsection{Tractable optimization objective}
\label{sec:training}

We derive a variational lower bound on the log-likelihood of the data, $\log
 p(\rvx_0)$, which motivates our training objective. With
 Eq.~\ref{eq:inf_process} as the variational distribution, we can show that (see
 proof in App.~\ref{app:derivations}),
\begin{equation}
\label{eq:elbo}
    \log p(\rvx_0) \geq \sum_{t=1}^T \E_{\rvx_{\leq t}}\Big[
        \underbrace{\E_{\rvu_t}[\log
        p_\theta(\rvx_{t-1} \mid \rvx_t, \rvu_t)]}_{\text{Denoising}}
        - \underbrace{\KL(q_{\theta^{\prime}}(\rvu_t \mid \rvx_{\leq t}) \|
            p_\phi(\rvu_t \mid
        \rvx_t))}_{\text{Alignment}} \Big] + C,
\end{equation}
where the trajectory $\rvx_{\leq t}$ is sampled from the noising process, i.e.,
$\rvx_{1:t} \sim q(\rvx_{1:t} \mid \rvx_0)$, and $\rvu_t$ is sampled from the
reference policy $q_{\theta^{\prime}}(\rvu_t \mid \rvx_{\leq t})$. Here, $C$ is
a constant that depends only on the noising process and not on $(\theta, \phi)$
and can be ignored in practice (see App.~\ref{app:elbo}). Intuitively, the
objective in Eq.~\ref{eq:elbo} is equivalent to learning a conditional VAE at
any time $t$. The first term in Eq. \ref{eq:elbo} is analogous to the denoising
objective of a standard diffusion model with an additional control input. During
training, the denoiser learns to reconstruct $\rvx_{t-1}$ from the current noisy
state $\rvx_t$ and a representation $\rvu_t$ drawn from the reference policy
$q_{\theta^{\prime}}(. \mid \rvx_{\leq t})$. Similarly, the second term
penalizes the deviation of the learned generative policy $p_\phi(.\mid \rvx_t)$
from the reference policy $q_{\theta^{\prime}}(.\mid \rvx_{\leq t})$ therefore
creating an information bottleneck. Due to the nature of this objective, one
would expect the latents $\rvu_t$ to capture meaningful semantic information.
Indeed, the latent $\rvu_t$ provides an interface for steering generation, e.g.,
for interpolating between images (Fig.~\ref{fig:interpolation}) which is often
hard to achieve with standard diffusion models \citep{preechakul2022diffae}.
Next, we discuss parameterization of different distributions and draw
connections to existing work in representation alignment which leads to a simple
recipe for training RAD.

\subsection{Design Choices and Training recipe}
\label{sec:param}
We start by proposing different choices for the forward and the generative
distributions from Sec.~\ref{sec:graph} and then simplify the RAD training
objective bound in Eq. \ref{eq:elbo}.

\paragraph{Noising Process.}
We choose the forward noising process the same as standard diffusion/stochastic
interpolant formulations i.e. $q(\rvx_t \mid \rvx_0) = \gN(\alpha_t \rvx_0,
\sigma_t^2 \mI_d)$. Following recent work on stochastic interpolants
\citep{albergo2023stochastic,ma2024sit}, we set $\alpha_t=1-t$ and $\sigma_t=t$
for $t \in [0,1]$. Next, we parameterize the reference policy at any time $t$
as,
\begin{equation}
    q_{\theta^\prime}(\rvu_t \mid \rvx_{\leq t}) = \gN\left(\alpha_t
        f_{\mathrm{pre}}(\rvx_0),\;
        \beta^2 \sigma_t^2 \mI\right), \qquad \beta > 0
\end{equation}
Throughout this work we fix $f_\text{pre}$ to a frozen pretrained DINOv2
\citep{oquab2024dinov2} encoder which extracts clean representations from data
$\rvx_0$. We provide an ablation over different choices of $f_\text{pre}$ in
Table~\ref{tab:design_choices}. Moreover, we adopt the same noising schedule for
the reference policy as the forward diffusion process. Our intuition behind this
choice is that during denoising at higher noise levels, only very coarse
structures emerge and therefore the representations at these noise levels must
encode more uncertainty. However, we emphasize that different choices of the
noising schedule are possible and remains an interesting direction for further
work. Moreover, since the supervision comes from $\rvx_0$ alone, the reference
policy drops its dependence on the trajectory and carries no learnable
parameters.

\paragraph{Choice of Representation Policy.} We first consider a simple class of
unimodal Gaussian policies,
\begin{align}
    p_\phi\left(\rvu_t \mid \rvx_t\right) = \gN\left(\alpha_t \vh_\phi(\rvx_t,
        t),\; \beta^2 \sigma_t^2 \mI\right) \label{eq:ext_prior}
\end{align}
where $\beta$ is a scalar which controls the width of the policy. We will later
show that controlling the width of the policy can significantly improve sample
quality (Fig.~\ref{fig:beta_inf_fid}). We write $\vh_\phi(\rvx_t, t) =
h_\phi(\rvh_t)$, where $\rvh_t = \vv_\theta^{(\ell)}(\rvx_t, t)$ is the hidden
state of the denoiser $\vv_\theta$ after block $\ell$ and $h_\phi$ is an MLP
projector. The semantic representation can then be sampled via $\rvu_t =
\alpha_t \vh_\phi(\rvx_t, t) + \beta\sigma_t \bar{\rvepsilon}$ where
$\bar{\rvepsilon} \sim \gN(\vzero, \mI_d)$. From a VAE perspective
(Sec.~\ref{sec:training}), the initial $\ell$ blocks of the denoiser combined
with the MLP head act as a noisy encoder for extracting representations from
$\rvx_t$.

 While the choice of unimodal Gaussian for the representation policy can induce
 multimodal denoising transitions (Eq.~\ref{eq:marginalized_transition}), we
 also consider conditional normalizing flow based policies
 \citep{dinh2017realnvp} with the following form (see App.~\ref{sec:design} for
 more details),
\begin{equation}
    \rvu_t = \alpha_t \vh_\phi(\rvx_t, t) + \beta\sigma_t \vg_\zeta(\rvh_t,
        \bar{\rvepsilon}),\qquad \bar{\rvepsilon}
    \sim \gN(\vzero,\mI_d)
\end{equation}
We find that using more expressive policies improves sample quality
(Table~\ref{tab:system_analysis}).

\paragraph{Choice of Denoising Policy.} We parameterize the denoising policy as
$p_\theta\left(\rvx_{t-1} \mid \rvx_t, \rvu_t\right) =
\gN\left(\vmu_\theta(\rvx_t, \rvu_t, t),\; \tilde{\sigma}_{t-1 \mid t}^2
\mI\right)$ and model $\vmu_\theta$ through the velocity network $\vv_\theta$,
which uses the SiT architecture throughout this work. The sampled representation
$\rvu_t \in \R^{N \times D}$ enters the backbone $\vv_\theta$ at the same layer
$\ell$ where it is predicted, after three operations
(Fig.~\ref{fig:architecture}):
\begin{align}
    \rvu_t^{n} = \frac{\rvu_t - \gamma\, \bar{u}_t}{\bar{\sigma}_t + \delta}
        \;\;\implies\;\;
    \rvu_t^{c} = \text{MLP}\left(\rvu_t^{n}\right)
        \;\;\implies\;\;
    \rvu_t^{f} = \text{Scale-Shift}\left(\rvu_t^{c}, \vc\right),
        \label{eq:scale_shift}
\end{align}
where $\bar{u}_t, \bar{\sigma}_t \in \R^{D}$ are the per-channel mean and
standard deviation of $\rvu_t$ across its $N$ tokens, $\gamma \in [0, 1]$
controls how much of the mean is removed (we use $\gamma = 0.6$), and $\delta$
is a small constant. Intuitively, Eq.~\ref{eq:scale_shift} first removes global
information, shared across all tokens, from $\rvu_t$ (as it could be harmful for
sample quality \citep{singh2026irepa}), then compresses the information into a
single token $\rvu_t^{c}$ with an MLP, and finally applies a learnable
AdaLN-style scale-shift transform conditioned on the conditioning embedding
$\vc$. We then add $\rvu_t^{f}$ back into the hidden state $\rvh_t$ which is
then passed into downstream SiT blocks to predict the velocity field. We present
an ablation over these design choices in Sec.~\ref{sec:ablations} and defer full
details to App.~\ref{sec:design}.

\paragraph{Simplified Training Objective}
Under these parameterizations, reweighting the terms of the bound in Eq.
\ref{eq:elbo} reduces to the following loss (see App.~\ref{app:objective}):
\begin{equation}
    \label{eq:objective}
    \Ls(\theta, \phi) = \E_{t, \rvx_0, \rvepsilon,\, \rvu_t \sim q}\Big[ \big\|
        \vv_\theta(\rvx_t,
        \rvu_t, t) - \vv_\text{ref} \big\|_2^2 +
            \tfrac{\mathrm{SNR}(t)}{2\beta^2} \big\|
        \rvf_\text{pre}(\rvx_0) - \vh_\phi(\rvx_t, t) \big\|_2^2 \Big],
\end{equation}
where $\text{SNR}(t) = \alpha_t^2 / \sigma_t^2$ and $\vv_\text{ref}=\dot\alpha_t
\rvx_0 + \dot\sigma_t \rvepsilon$. The simplified objective in Eq.
\ref{eq:objective} is similar to the heuristic alignment objective proposed in
REPA with the exception of the latent representations $\rvu_t$ now also part of
the denoiser. Indeed, we show that existing external alignment methods like REPA
are special cases of our framework (see App.~\ref{app:prior_work}). During
training, we always fix $\beta=1.0$. However, we find that tuning $\beta$ during
inference improves sample quality (Fig.~\ref{fig:beta_inf_fid}).
\figAlgorithms

\paragraph{Sampling.} Following Eq.~\ref{eq:marginalized_transition}, existing
off-the-shelf samplers for flow matching models can be directly applied to RAD
by estimating the marginal vector field and plugging it into the ODE/SDE
sampler. For instance,
\begin{equation}
    \frac{\mathrm{d}\rvx_t}{\mathrm{d}t} = \vv_{\theta, \phi}(\rvx_t, t) =
        \E_{\rvu_t \sim p_\phi(\rvu_t \mid
        \rvx_t)}\left[\vv_\theta(\rvx_t, \rvu_t, t)\right] \approx \frac{1}{M}
            \sum_{i=1}^M \vv_\theta(\rvx_t, \rvu_t^{(i)}, t).
        \label{eq:rad_sampling}
\end{equation}
where $\rvu_t^{(i)} \sim p_\phi(\rvu_t \mid \rvx_t)$ are samples from the
representation policy. In early experiments, we find that a single sample
($M=1$) is sufficient for high-quality generation. We provide a pseudo-code in
Alg.~\ref{alg:sampling}.

However, estimating the expectation via Monte Carlo sampling in
Eq.~\ref{eq:rad_sampling} makes each velocity evaluation stochastic. Therefore,
when sampling $\rvu_t \sim p_\phi(\rvu_t \mid \rvx_t)$, we consider two choices
for the policy noise. With \emph{fresh} policy noise, a new $\bar{\rvepsilon}
\sim \gN(\vzero, \mI)$ is drawn at every denoising step, so each step uses an
independent sample from the representation policy $p_\phi(\rvu_t \mid \rvx_t)$,
while the update of $\rvx_t$ itself injects no noise (Alg.~\ref{alg:sampling}).
With \emph{fixed} policy noise, $\bar{\rvepsilon}$ is drawn once together with
$\rvx_1$ and kept fixed along the trajectory. This ensures that the sample is a
deterministic function of the initial noise. Both variants require one network
evaluation per step. In practice, we find that both variants yield similar
sample quality (see Table~\ref{tab:policy_noise}) and therefore, use the fresh
policy noise variant (unless specified otherwise) for all experiments in the
main paper.

Since sampling in RAD follows
the generative ODE in Eq.~\ref{eq:rad_sampling}, it can be steered for
controllable generation by replacing the velocity field $\vv_{\theta,
\phi}(\rvx_t, \rvu_t, t)$ with a conditional velocity field $\vv_{\theta, \phi}(\rvx_t,
t, \rvu_t, \vy)$, where $\vy$ is an observed conditioning signal. This conditional velocity
field can be obtained through various
test-time~\citep{chung2023diffusion,ICLR2024_77110269,ICLR2024_dfbb3d18,pandey2025variational} 
or finetuning~\citep{domingo-enrich2025adjoint, pandey2026hierarchical} approaches. 
We do not present this conditional generation formulation in this work, 
but it is an interesting direction for future research.

\section{Experiments}
\label{sec:experiments}

We evaluate RAD on class-conditional ImageNet $256\times256$ against
REPA \citep{yu2025repa}, our primary baseline, and study the effect of feeding
the aligned representation back into the denoiser. We make the following observations: 
\textbf{(i) Training and data efficiency.} RAD trains faster and
is more parameter- and data-efficient (Sec.~\ref{sec:gen_quality}). At XL/2
scale, RAD matches the sample quality of REPA in $27\times$ fewer training
iterations (Fig.~\ref{fig:overview}b). At B/2 scale, RAD outperforms REPA trained
on $10\times$ more data (Fig.~\ref{fig:efficiency}b,c). \textbf{(ii) Sampling
efficiency.} RAD improves the quality-vs-NFE tradeoff
(Sec.~\ref{sec:sampling_efficiency}) needing ${\sim}10\times$ 
fewer sampling steps to reach the same FID as
a REPA model of similar sample quality. \textbf{(iii)
Representation quality.} RAD preserves representation semantics deeper into
the denoiser and supports smooth interpolation in representation space
(Sec.~\ref{sec:rep_quality}). \textbf{(iv) Generalization.} RAD can be combined
with other alignment methods such as REG \citep{wu2025reg} and transfers to text-to-image
generation (Sec.~\ref{sec:generalization}). We summarize the experimental setup
below with full details in App.~\ref{app:exp_details}.

\paragraph{Tasks and Datasets.} We evaluate RAD under two settings. For
\textbf{class-conditional generation}, we exactly follow the experimental setup
of REPA and train on ImageNet $256\times256$ in the latent space of the Stable
Diffusion VAE~\citep{rombach2022ldm}. For \textbf{text-to-image synthesis}, we
train on the BLIP3o pre-training captions~\citep{chen2025blip3o} at
$256\times256$ resolution, conditioning on caption features from a frozen
Qwen3-0.6B text encoder~\citep{yang2025qwen3}, and, for qualitative samples
only, fine-tune on the curated BLIP3o-60k instruction set.

\paragraph{Models and Training.} We use SiT-\{B/2,L/2,XL/2\} backbones for
ImageNet and SiT-\{B/2,XL/2\} for text-to-image synthesis (/2 indicates 2x2
patch size). For all experiments, we use the DINOv2-B model as the default
external supervisor $f_{\mathrm{pre}}$. For notational convenience, we denote a
SiT backbone trained with RAD by its size, e.g., RAD-XL/2. Following REPA, all
models are trained with a global batch size of 256 using the AdamW
optimizer~\citep{loshchilov2018decoupled} with a learning rate of 1e-4, and an
Exponential Moving Average (EMA) of the weights, for up to 400K iterations
(primarily due to compute limitations) unless stated otherwise.

\paragraph{Baselines.} Besides REPA, we compare against the unaligned SiT
baseline \citep{ma2024sit} and existing representation alignment methods that
build directly on top of the REPA framework
(Table~\ref{tab:gen_quality_baselines}), though we hypothesize that such methods
would also benefit from our framework (Sec.~\ref{sec:generalization}). Note that
REPA and RAD share an identical backbone, encoder, data, and optimization
schedule. The difference is that REPA does not feed the aligned representation
back into the denoiser.

\paragraph{Evaluation Metrics.} To measure sample quality, we report
FID~\citep{heusel2017fid}, and sFID~\citep{nash2021sfid} on 50k samples. For
class conditional ImageNet, we additionally report Inception
Score~\citep{salimans2016is}, and precision/recall~\citep{kynkaanniemi2019prec}
to measure the fidelity--diversity tradeoff. We use the Euler--Maruyama SDE
sampler with 250 steps and no classifier-free guidance~\citep{ho2022cfg} unless
noted. For text-to-image synthesis, we report compositional accuracy on
GenEval~\citep{ghosh2023geneval}.

\subsection{Generation Quality}
\label{sec:gen_quality}

\tabGenQualityComparison

\paragraph{Faster training convergence at every model scale.} We compare the
training efficiency of RAD in Fig.~\ref{fig:efficiency} (full results in
Table~\ref{tab:gen_quality} in App.~\ref{app:gen_results}). Interestingly, after
only 100K training iterations, RAD matches or surpasses REPA at 400K iterations
in sample quality for all three backbones, a $4\times$ minimum reduction in
training steps. Moreover, the gains grow further against long training runs
(Fig.~\ref{fig:overview}b and Table~\ref{tab:convergence_fid}): after 100K
iterations, RAD-XL/2 achieves a lower FID than SiT-XL/2 trained for 7M
iterations ($70\times$ fewer), and after 150K iterations it outperforms
REPA-XL/2 trained for 4M iterations ($27\times$ fewer). Since RAD and REPA share
the backbone, encoder, and training schedule, we attribute these gains to
conditioning the denoiser on sampled representations (we provide more detailed
ablations in Sec.~\ref{sec:ablations}).

\paragraph{Better parameter and data efficiency.} We find that RAD exhibits
superior parameter efficiency compared to REPA. At the same budget of 400K
iterations, RAD-B/2 outperforms REPA-L/2 (FID 7.7 vs.\ 10.0) with $3.2\times$
fewer parameters (Fig.~\ref{fig:efficiency}b), and performs on par with
REPA-XL/2 (FID 7.9) with $4.8\times$ fewer parameters (Fig.~\ref{fig:efficiency}a). 
RAD-B/2 even surpasses SiT-XL/2 trained for 7M iterations (FID 8.3), despite using $4.8\times$ fewer
parameters and $17.5\times$ fewer training iterations.

RAD is also more data efficient. We train B/2 models for 400K
iterations on subsets of ImageNet and report the best FID of each run over its
checkpoints. Trained on 10\% of ImageNet, RAD outperforms REPA 
trained on the full dataset i.e., RAD needs $10\times$ fewer
distinct training images under the same iteration budget. Moreover, REPA
does not improve beyond 10\% of the data suggesting underfitting while RAD keeps
improving.

\paragraph{Comparison with other baselines.} Next, we compare RAD with
additional alignment baselines (Table~\ref{tab:gen_quality_baselines}) reporting
FID-50k after 80 epochs without classifier-free guidance. Across different model
scales, RAD registers strong sample quality achieving best FID among competitors
on B/2 and L/2 scales while being competitive to REG at the XL/2 scale. However, we
show that our framework can be readily combined with other baselines
(Sec.~\ref{sec:generalization}) and thus hypothesize that a lot of the
techniques used by existing representation alignment would be complementary to
RAD.

\figEfficiency

\figSamplingEfficiency

\figInterpolation

\paragraph{Token noise coupling makes classifier-free guidance effective.}
Classifier-free guidance (CFG)~\citep{ho2022cfg} is the standard way to improve
the sample quality of diffusion models. However, we find that vanilla CFG yields
less significant improvements in sample quality in RAD
(Table~\ref{tab:guided_ablation}). We hypothesize this to a lack of sample-level
diversity from the representation policy: the policy noise is independent across
the $N$ tokens of the representation $\rvu_t$, so it largely averages out once
$\rvu_t$ is compressed into a single token (Eq.~\ref{eq:scale_shift}). We
therefore correlate noise between different tokens in the representation
$\rvu_t$ by introducing \emph{noise coupling} (NC) which shares part of the
policy noise $\boldsymbol{\xi}$ across tokens (to prevent averaging),
\begin{equation}
    \bar{\rvepsilon}_n = \rho\,\boldsymbol{\xi} +
        \sqrt{1-\rho^2}\,\boldsymbol{\eta}_n,
    \qquad \boldsymbol{\xi},\, \boldsymbol{\eta}_n \sim \gN(\vzero, \mI_d),
    \quad n = 1, \dots, N,
    \label{eq:token_coupling}
\end{equation}
so that the policy of each individual token is unchanged. The representation is
then sampled using the policy noise $\bar{\rvepsilon}_n$. On its own, noise
coupling worsens sample quality, but combined with CFG it yields the largest
improvement (Table~\ref{tab:guided_ablation}). We find that similar observations
also hold for autoguidance (AG) \citep{karras2024autoguidance}. Our best model
combines such noise coupling with CFG and AG and achieves an FID-50k of 1.96
after only 80 epochs and with 50 sampling steps
(Table~\ref{tab:gen_quality_cfg}) which is comparable to SOTA methods (see
App.~\ref{app:guided_sampling} for more details).

\subsection{Sampling Efficiency}
\label{sec:sampling_efficiency}

\paragraph{Better quality-vs-compute tradeoff at inference.}
Marginalizing out the representations can lead to more expressive denoising
policies than standard diffusion models (Sec.~\ref{sec:method}). This could be
particularly useful in a few step sampling regime where a more expressive
transition should tolerate larger steps. To test this, we compare two models of
similar quality: RAD-XL/2 trained for 150K iterations and the official REPA-XL/2
checkpoint trained for 4M iterations, which reach FID 5.7 and 5.9, respectively,
with the standard 250-step Euler--Maruyama SDE sampler. We then reduce the
number of sampling steps. With the deterministic Euler ODE sampler
(Fig.~\ref{fig:sampling_nfe} and Table~\ref{tab:sampling_efficiency}), RAD
outperforms REPA at every step budget and needs roughly $10\times$ fewer
sampling steps to reach the same FID. With the stochastic Euler--Maruyama SDE
sampler (Fig.~\ref{fig:sampling_nfe_sde} and
Table~\ref{tab:sampling_efficiency_sde}), RAD needs around $5\times$ fewer
steps.

\paragraph{Compatibility with SOTA samplers.}
RAD is readily compatible with alternative state-of-the-art samplers for
diffusion models (Fig.~\ref{fig:samplers}). With UniPC~\citep{zhao2023unipc},
RAD-XL/2 (at 400k steps) reaches FID 6.5 at 5 steps. Moreover, sampler
performance is robust to the choice of fresh or fixed policy noise
(Tab.~\ref{tab:policy_noise}).

\subsection{Representation Quality}
\label{sec:rep_quality}

\paragraph{Interpolation in Representation Space}
\label{sec:interpolation}
As the denoiser conditions on $\rvu_t$ at every step, the learned representation
can be used to manipulate generations. We test this by interpolating between two
real ImageNet images, $\rvx_0^A$ and $\rvx_0^B$. From each image we extract a
representation $\bar{\rvu} = \vh_\phi(\rvx_{t_c}, t_c)$ at a small noise level
$t_c$. For a weight $m \in [0, 1]$, we linearly interpolate the representations,
$\bar{\rvu}_m = m\,\bar{\rvu}^A + (1-m)\,\bar{\rvu}^B$, and likewise the class
embeddings, and obtain the starting noise $\rvx_{1,m}$ by spherical linear
interpolation (slerp) between $\rvx_1^A$ and $\rvx_1^B$, the noises obtained by
inverting each image with the Euler ODE sampler. We then generate each
interpolated sample by running the sampler from $\rvx_{1,m}$, conditioned on
$\bar{\rvu}_m$ and the interpolated class embedding. Since REPA does not feed
its representations back into the denoiser, we interpolate only its starting
noise and class embedding in the same way. Fig.~\ref{fig:interpolation}
interpolates from a baboon to a macaque with RAD-XL/2 at 400K iterations and
REPA-XL/2 at 4M iterations. RAD changes the subject gradually while the pose,
viewpoint, and framing stay fixed, indicating a smooth representation space. In
contrast, REPA's intermediate samples switch between poses and scenes, even
though it is trained $10\times$ longer. Additional samples and implementation
details are given in App.~\ref{app:interp_results}.

\paragraph{Classification Accuracy}
\label{sec:linear_probe}
Following REPA, we train a linear classifier on the hidden state at block $k$ of
a noised input $\rvx_t$ and report ImageNet top-1 accuracy for RAD-B/2 and
REPA-B/2, both trained for 400K iterations and aligned at block 4 (see
App.~\ref{app:exp_details} for details). We make two key observations from
Fig.~\ref{fig:linear_probe}. First, RAD preserves discriminative features deeper
into the network than REPA (Fig.~\ref{fig:linear_probe}, top). At the aligned
block, where both models regress the reference DINOv2 features, the two have
similar linear probing accuracy. For the remaining blocks, RAD is more accurate
than REPA by up to 4.4 points at $t=0.1$. More interestingly, the linear probing
accuracy of RAD decreases more slowly than that of REPA. Intuitively, this is
likely due to feeding the representations back into the denoiser which keeps the
later blocks linearly separable. Second, the features improve with scale
(Fig.~\ref{fig:linear_probe}, bottom). RAD-XL/2 improves over RAD-B/2 by 8 to 9
points consistently for the most part of the diffusion trajectory $t \leq 0.5$
(71.7\% vs.\ 63.4\% at $t=0.1$).

\figRegTextToImageLinearProbe

\subsection{Generalization Capabilities}
\label{sec:generalization}

\paragraph{RAD benefits REG convergence.}

REG denoises the class token of the encoder jointly with the image tokens. We
extend REG within the RAD framework by treating the projection of the
intermediate noisy patch tokens in REG as the mean of the representation policy
(see App.~\ref{app:exp_details} for details). Table~\ref{tab:reg_convergence}
shows that combining REG with our approach improves the training convergence of
REG at both model scales. Notably, REG combined with RAD converges twice as
fast: on SiT-B/2 it matches the 200K-iteration FID of REG after 100K iterations,
and on SiT-L/2 it matches the 400K-iteration FID of REG after 200K iterations
(Table~\ref{tab:gen_quality_reg}).

\paragraph{RAD transfers to Text-to-Image Generation}
Table~\ref{tab:t2i} shows that RAD improves sample quality without sacrificing
prompt adherence, i.e., at both B/2 and XL/2 scales it has lower FID and equal
or higher GenEval than REPA, so it dominates REPA on both axes. The gains are
largest for the smaller backbone: RAD-B/2 reduces FID from 10.77 to 8.42 and
improves GenEval from 0.138 to 0.166, while RAD-XL/2 improves FID (3.94 vs.\
4.31) with similar GenEval (0.266 vs.\ 0.258).

\subsection{Ablation Experiments}
\label{sec:ablations}

\paragraph{Impact of $\beta$.} The policy scale $\beta$ sets the noise level of
the representation, both during training and, as $\beta_{\mathrm{inf}}$, at
inference. During training, removing the noise entirely ($\beta = 0$, i.e.,
noise-free teacher features) heavily degrades sample quality (See
Fig.~\ref{fig:beta_train}, Table~\ref{tab:beta_train}) which implies the
importance of using stochastic policies for training RAD. We find that
$\beta=1.0$ works reasonably well at all model scales and therefore, propose to
set it to 1.0 in practice. At inference, we find that reducing $\beta$ usually
helps with improving sample quality without affecting recall adversely
(Figs.~\ref{fig:beta_inf_fid} and~\ref{fig:beta_inf}, and
Table~\ref{tab:beta_inf}).

\paragraph{Reference vs.\ on-policy supervision.} During training, the controls
fed to the denoiser can be sampled either from the reference policy $q(\rvu_t
\mid \rvx_0)$, as in our objective (Eq.~\ref{eq:objective}), or from the learned
representation policy $p_\phi(\rvu_t \mid \rvx_t)$, which matches how the
controls are obtained at inference. We find that training with the reference
policy yields better sample quality: with on-policy supervision, RAD-B/2 reaches
an FID of 22.4, compared with 12.6 with reference supervision. Therefore, we
sample teacher representations for all experiments considered in this work.

\paragraph{Design Choices.} Table~\ref{tab:design_choices} highlights the impact
of the choice of external teacher, the representation injection depth and the
alignment loss weighting in RAD-B/2. RAD is relatively robust to the choice of
the encoder. DINOv2-B and DINOv2-L perform comparably, and, similar to REPA and
REG, we use DINOv2-B for all our analysis. Moreover, we find that an injection
depth of 2 and an alignment weighting coefficient of $p=1$ works the best at
RAD-B/2 scale.

\tabAblations

\paragraph{System Analysis.} Table~\ref{tab:system_analysis} isolates the
contribution of each component of RAD-B/2 towards sample quality. We start from
REPA (config \textsc{a}) and add components sequentially, ending at the
configuration used in Table~\ref{tab:gen_quality_baselines} (config \textsc{f}).
We emphasize that conditioning the denoiser on representations $\rvu_t$ drawn
from a Gaussian policy of Eq.~\ref{eq:ext_prior} (configs
\textsc{b}--\textsc{c}) already lowers the FID from 24.4 to 10.3. Using a more
expressive representation policy, the normalizing flow of
Eq.~\ref{eq:flow_policy} (config \textsc{d}), lowers FID further from 10.3 to
8.7, indicating that more expressive policies can be readily adopted with RAD.
Control normalization and modulation (config \textsc{e}) and earlier injection
with the weighted alignment loss (config \textsc{f}) add smaller gains resulting
in FID of 7.7.

\paragraph{Deterministic vs Stochastic Policies.} To isolate the role of
stochasticity in the representation policy $p_\phi(\rvu_t \mid \rvx_t)$, we
train a deterministic variant of RAD-B/2 that removes the noise from both the
reference and the representation policy while keeping everything else
(denoiser architecture, loss formulation) unchanged from Config \textsc{c} in
Table~\ref{tab:system_analysis}. Table~\ref{tab:det_vs_stoch} shows that
stochastic policies outperform the deterministic one by a large margin.

\paragraph{Additional compute costs.} Our control pathway is relatively
lightweight. At B/2 model scale, the whole pathway costs 12\% additional
inference FLOPs over REPA, most of it from the projector $h_\phi$, which REPA
also trains but discards at inference. However, the control pathway has the same
architecture for L/2 and XL/2: therefore the overhead falls to 3.5\% at L/2 and
2.5\% at XL/2.

\section{Conclusion}
\label{sec:conclusion}

We introduced Representation-Augmented Diffusions (RAD), a class of diffusion
models that embed semantic representations as latent variables within the
denoising process: at every step, the denoiser conditions on a representation
predicted from its own noisy state. The framework recovers representation
alignment methods as special cases and reveals a simple design choice they leave
unused: feeding the aligned representation back into the denoiser. RAD converges
in fewer iterations during training, is more data- and parameter-efficient,
needs fewer sampling steps, and preserves discriminative representations deeper
in the denoiser. However, some limitations remain. For unguided RAD models, 
we find that the gains in sample quality come at the cost of
diversity. RAD attains substantially higher precision than REPA but lower recall
across increased training and model scales. Improving this
fidelity--diversity tradeoff, for instance through the variance of the
representation policy or the sampler, is an important direction for future work.

\subsection*{Acknowledgments}
We would like to thank Jan Niklas Groeneveld for additional discussions and feedback on the paper. 
Stephan Mandt acknowledges support from the National Science Foundation (NSF) through CAREER Award IIS-2047418 and Grant IIS-2611549.

\subsection*{AI use statement}
In this work, we used generative AI tools for formulating mathematical claims,
assisting in the writing of proofs, and implementing methods. We did not use
generative AI tools to help develop theoretical models or conceptual frameworks,
to provide critical ingredients for proving mathematical claims, or to interpret
results.

Additionally, we used generative AI tools for creating or modifying scientific
figures or images, creating or editing software code, drafting parts of our
research paper, summarizing or analyzing existing literature, searching for
information, editing our research paper to improve readability, and identifying
relevant literature. We reviewed all AI-assisted work: two authors verified all
LLM-generated code, we verified information, and we checked research ideas for
potential plagiarism through a manual literature survey. We take responsibility
for the final content of this work, including text, claims, or artifacts
produced with the aid of generative AI.

\bibliographystyle{rad_arxiv}
\bibliography{rad_arxiv}

\clearpage
\appendix
\section{Appendix}

\etocsettocstyle{\subsection*{Contents}}{}
\etocsetnexttocdepth{subsubsection}
\localtableofcontents
\clearpage

\subsection{Background}
\label{sec:background}

\subsubsection{Stochastic Interpolants}
\label{sec:bg-diffusion}

Flow matching or Stochastic Interpolants convert data to noise by defining a
noising process $\rvx_t = \alpha_t \rvx_0 + \sigma_t \rvepsilon$, $t \in [0,
1]$. A commonly adopted choice is the straight-line interpolant
\citep{liu2023flow} where $\alpha_t=1-t$ and $\sigma_t=t$. The marginals of
$\rvx_t$ are transported by the velocity field $\vv(\rvx_t, t) = \E[\dot\alpha_t
\rvx_0 + \dot\sigma_t \rvepsilon \mid \rvx_t]$, which a network $\vv_\theta$
learns by regressing onto the conditional velocity target,
\begin{equation}
    \Ls_{\mathrm{FM}}(\theta) = \E_{t, \rvx_0, \rvepsilon}\big[ \|
        \vv_\theta(\rvx_t, t)
    - (\dot\alpha_t \rvx_0 + \dot\sigma_t \rvepsilon) \|_2^2 \big].
    \label{eq:fm}
\end{equation}
During inference, the \emph{probability-flow} ODE $\mathrm{d}\rvx_t =
\vv_\theta(\rvx_t, t)\,\mathrm{d}t$, started from $\rvx_1 \sim \gN(\vzero,
\mI)$, converts noise back to data. For all our experiments in this work, we use
the widely adopted Scalable Interpolant Transformer (SiT) as the denoiser
backbone in combination with the straight-line interpolants.

\subsubsection{Representation Alignment}
\label{sec:bg-repa}

Representation alignment adds an auxiliary loss to the velocity matching
objective in Eq. \ref{eq:fm} that pulls the hidden states of the velocity field
toward a semantic target. Let $\rvh_t = \vv_\theta^{(\ell)}(\rvx_t, t)$ denote
the hidden state of the denoiser $\vv_\theta$ after $\ell$ SiT blocks. The main
idea behind representation alignment is to align the outputs $\rvy =
f_{\mathrm{pre}}(\rvx_0) \in \R^{N \times D}$ with the intermediate layer
representations $\rvh_t$, where $f_\text{pre}$ is a frozen pretrained encoder
such as DINOv2, applied to the clean image. A trainable MLP projector $h_\phi$
maps the hidden state to the encoder's dimension, and the two are aligned token
by token with a cosine loss,
\begin{equation}
    \Ls_{\mathrm{REPA}}(\theta, \phi) = -\E_{t, \rvx_0, \rvepsilon}\Big[
        \frac{1}{N} \sum_{n=1}^{N}
    \cos\big( \rvy^{[n]}, h_\phi(\rvh_t)^{[n]} \big) \Big],
    \qquad
    \Ls = \Ls_{\mathrm{FM}} + \lambda \, \Ls_{\mathrm{REPA}},
    \label{eq:repa}
\end{equation}
where $\lambda > 0$ is a scalar hyperparameter. In most representation alignment
methods, the projector is discarded after training i.e. at sampling time the
denoiser receives $\rvx_t$ alone. In contrast, we propose a class of diffusion
models where representations are natively embedded within the denoising process
and help guide generation (Sec.~\ref{sec:method}). We show that this can
largely benefit a number of aspects during training and inference
(Sec.~\ref{sec:experiments}).

\subsection{Derivations}
\label{app:derivations}

\subsubsection{Variational Bound}
\label{app:elbo}

We derive the variational bound in Eq.~\ref{eq:elbo}, using the noising process
in Eq.~\ref{eq:inf_process} as the variational distribution. Throughout, $t \in
\{1, \dots, T\}$ indexes a uniform grid on $[0, 1]$, with $t = T$ corresponding
to the terminal time $1$ of the interpolant. Recall the generative process,
\begin{equation}
    p_{\theta, \phi}(\rvx_{0:T}, \rvu_{1:T})
        = p(\rvx_T) \prod_{t=1}^T p_\phi(\rvu_t \mid \rvx_t)\,
            p_\theta(\rvx_{t-1} \mid \rvx_t,
        \rvu_t),
    \label{eq:app_gen}
\end{equation}
and the noising process,
\begin{equation}
    q_{\theta^{\prime}}(\rvx_{1:T}, \rvu_{1:T} \mid \rvx_0)
        = \prod_{t=1}^T q(\rvx_t \mid \rvx_{t-1})\, q_{\theta^{\prime}}(\rvu_t
            \mid \rvx_{\leq t}),
    \label{eq:app_inf}
\end{equation}
where $\rvx_{\leq t} = (\rvx_0, \dots, \rvx_t)$. The reference policy may depend
on any subset $\rvx_S$ of these states, $S \subseteq \{0, \dots, t\}$, i.e.,
$q_{\theta^{\prime}}(\rvu_t \mid \rvx_{\leq t}) = q_{\theta^{\prime}}(\rvu_t
\mid \rvx_S)$; for example, $S = \{0\}$ for the pretrained encoder in
Sec.~\ref{sec:param}. Conditioning on $\rvx_{\leq t}$ covers all such choices,
and the derivation below holds for any of them, including a reference policy
whose parameters $\theta^{\prime}$ are not optimized, such as a frozen
pretrained encoder. We also use that the noising process is Markov, i.e.,
$q(\rvx_t \mid \rvx_{t-1}) = q(\rvx_t \mid \rvx_{t-1}, \rvx_0)$.

With the noising process as the variational distribution, Jensen's inequality
gives the evidence lower bound,
\begin{align}
    \log p(\rvx_0)
    &= \log \E_{q_{\theta^{\prime}}(\rvx_{1:T}, \rvu_{1:T} \mid \rvx_0)}\left[
        \frac{p_{\theta, \phi}(\rvx_{0:T},
            \rvu_{1:T})}{q_{\theta^{\prime}}(\rvx_{1:T}, \rvu_{1:T}
        \mid \rvx_0)} \right] \notag \\
    &\geq \E_{q_{\theta^{\prime}}(\rvx_{1:T}, \rvu_{1:T} \mid \rvx_0)}\left[
        \log
        \frac{p_{\theta, \phi}(\rvx_{0:T},
            \rvu_{1:T})}{q_{\theta^{\prime}}(\rvx_{1:T}, \rvu_{1:T}
        \mid \rvx_0)} \right].
    \label{eq:app_jensen}
\end{align}
Substituting the factorizations in Eqs.~\ref{eq:app_gen} and~\ref{eq:app_inf},
the log-ratio in Eq.~\ref{eq:app_jensen} becomes
\begin{equation}
    \log \frac{p_{\theta, \phi}(\rvx_{0:T},
        \rvu_{1:T})}{q_{\theta^{\prime}}(\rvx_{1:T},
        \rvu_{1:T} \mid \rvx_0)}
    = \log \frac{p(\rvx_T) \prod_{t=1}^T p_\phi(\rvu_t \mid \rvx_t)\,
        p_\theta(\rvx_{t-1} \mid
        \rvx_t, \rvu_t)}{\prod_{t=1}^T q(\rvx_t \mid \rvx_{t-1})\,
            q_{\theta^{\prime}}(\rvu_t \mid
        \rvx_{\leq t})}.
    \label{eq:app_ratio}
\end{equation}
For $t \geq 2$, the Markov property and Bayes' rule give $q(\rvx_t \mid
\rvx_{t-1}) = q(\rvx_{t-1} \mid \rvx_t, \rvx_0)\, q(\rvx_t \mid \rvx_0) /
q(\rvx_{t-1} \mid \rvx_0)$. In the product over $t$, the marginal ratios
cancel and, with the factor $q(\rvx_1 \mid \rvx_0)$ at $t = 1$, leave only
$q(\rvx_T \mid \rvx_0)$:
\begin{equation}
    \prod_{t=1}^T q(\rvx_t \mid \rvx_{t-1})
        = q(\rvx_T \mid \rvx_0) \prod_{t=2}^T q(\rvx_{t-1} \mid \rvx_t, \rvx_0).
    \label{eq:app_flip}
\end{equation}
Substituting Eq.~\ref{eq:app_flip} into Eq.~\ref{eq:app_ratio} and grouping the
factors of each step,
\begin{align}
    \log \frac{p_{\theta, \phi}(\rvx_{0:T},
        \rvu_{1:T})}{q_{\theta^{\prime}}(\rvx_{1:T},
        \rvu_{1:T} \mid \rvx_0)}
    = {} & \sum_{t=1}^T \Big[ \log p_\theta(\rvx_{t-1} \mid \rvx_t, \rvu_t)
        + \log \frac{p_\phi(\rvu_t \mid \rvx_t)}{q_{\theta^{\prime}}(\rvu_t \mid
            \rvx_{\leq t})}
        \Big] \notag \\
    & - \sum_{t=2}^T \log q(\rvx_{t-1} \mid \rvx_t, \rvx_0)
        + \log \frac{p(\rvx_T)}{q(\rvx_T \mid \rvx_0)}.
    \label{eq:app_decomp}
\end{align}

\paragraph{Reducing the expectations.} The summand with index $t$ in
Eq.~\ref{eq:app_decomp} depends only on $(\rvx_{\leq t}, \rvu_t)$, and no state
of the noising process depends on a representation, so all other variables
integrate out of the expectation in Eq.~\ref{eq:app_jensen}. We write
$\E_{\rvx_{\leq t}, \rvu_t} := \E_{q(\rvx_{1:t} \mid \rvx_0)}\,
\E_{q_{\theta^{\prime}}(\rvu_t \mid \rvx_{\leq t})}$ for the resulting
expectation and $\E_{\rvx_{\leq t}}$ for its outer expectation.

\paragraph{Simplifying the terms.} The first term of Eq.~\ref{eq:app_decomp}
gives the denoising term $\E_{\rvx_{\leq t}, \rvu_t}[\log p_\theta(\rvx_{t-1}
\mid \rvx_t, \rvu_t)]$. For the remaining terms,
\begin{align}
    \E_{\rvx_{\leq t}, \rvu_t}\left[ \log \frac{p_\phi(\rvu_t \mid
        \rvx_t)}{q_{\theta^{\prime}}(\rvu_t
        \mid \rvx_{\leq t})} \right]
    &= -\E_{\rvx_{\leq t}}\big[ \KL\big( q_{\theta^{\prime}}(\rvu_t \mid
        \rvx_{\leq t}) \,\|\,
        p_\phi(\rvu_t \mid \rvx_t) \big) \big], \label{eq:app_align} \\
    -\E_{\rvx_{\leq t}, \rvu_t}\big[ \log q(\rvx_{t-1} \mid \rvx_t, \rvx_0)
        \big]
    &= \E_{q(\rvx_t \mid \rvx_0)}\big[ \gH\big( q(\rvx_{t-1} \mid \rvx_t,
        \rvx_0) \big) \big]
        =: \gH_t, \label{eq:app_entropy} \\
    \E_{q(\rvx_T \mid \rvx_0)}\left[ \log \frac{p(\rvx_T)}{q(\rvx_T \mid
        \rvx_0)} \right]
    &= -\KL\big( q(\rvx_T \mid \rvx_0) \,\|\, p(\rvx_T) \big).
        \label{eq:app_prior}
\end{align}
Eq.~\ref{eq:app_align} is the alignment term. Eq.~\ref{eq:app_entropy} holds
because the integrand depends only on $(\rvx_{t-1}, \rvx_t)$, whose marginal is
$q(\rvx_t \mid \rvx_0)\, q(\rvx_{t-1} \mid \rvx_t, \rvx_0)$ by the chain rule.

\paragraph{Combining.} Substituting these terms into Eq.~\ref{eq:app_jensen}, we
arrive at
\begin{align}
    \log p(\rvx_0) \geq \sum_{t=1}^T \Big[
        & \underbrace{\E_{\rvx_{\leq t}, \rvu_t}\big[ \log p_\theta(\rvx_{t-1}
            \mid \rvx_t, \rvu_t)
        \big]}_{\text{Denoising}} \notag \\
        & - \underbrace{\E_{\rvx_{\leq t}}\big[ \KL\big(
            q_{\theta^{\prime}}(\rvu_t \mid
        \rvx_{\leq t}) \,\|\, p_\phi(\rvu_t \mid \rvx_t) \big)
            \big]}_{\text{Alignment}} \Big] + C,
    \label{eq:app_elbo}
\end{align}
where $C = \sum_{t=2}^T \gH_t - \KL(q(\rvx_T \mid \rvx_0) \,\|\, p(\rvx_T))$ is
determined by the noising process alone and does not depend on $(\theta, \phi)$.
The KL term in $C$ vanishes when $q(\rvx_T \mid \rvx_0) = p(\rvx_T) =
\gN(\vzero, \mI)$, as for the interpolant in Sec.~\ref{sec:param}. This reduces
Eq.~\ref{eq:app_elbo} to Eq.~\ref{eq:elbo}, which concludes the proof.

\paragraph{Remark.} In standard diffusion models, the denoising term and the
entropy $\gH_t$ combine into $-\E\big[\KL(q(\rvx_{t-1} \mid \rvx_t, \rvx_0)
\,\|\, p_\theta(\rvx_{t-1} \mid \rvx_t))\big]$. In RAD, the same holds for $t
\geq 2$ only if $\rvu_t$ is conditionally independent of $\rvx_{t-1}$ given
$(\rvx_t, \rvx_0)$, e.g., for $S = \{0\}$. In general,
$q_{\theta^{\prime}}(\rvu_t \mid \rvx_{\leq t})$ depends on $\rvx_{t-1}$, and
the denoising term keeps the log-likelihood form of Eq.~\ref{eq:elbo}.

\subsubsection{Parameterization}
\label{app:param}

We now specify the distributions in Eq.~\ref{eq:elbo}. We begin with the two
distributions of the alignment term, the reference policy
$q_{\theta^{\prime}}\left(\rvu_t \mid \rvx_{\leq t}\right)$ and the
representation policy $p_\phi\left(\rvu_t \mid \rvx_t\right)$. We then describe
the representation-augmented denoiser $p_\theta\left(\rvx_{t-1} \mid \rvx_t,
\rvu_t\right)$ and the forward process, and close with the resulting objective.

\paragraph{Reference and representation policies.} A frozen pretrained encoder
$f_{\mathrm{pre}}$ extracts the intermediate representations $\rvu_t$ from the
clean image $\rvx_0$. Following REPA, $f_{\mathrm{pre}}$ can be DINOv2 or any
other pretrained model that produces semantically meaningful features. Since the
supervision comes from $\rvx_0$ alone, the reference policy drops its dependence
on the trajectory and carries no learnable parameters,
$q_{\theta^{\prime}}\left(\rvu_t \mid \rvx_{\leq t}\right) = q\left(\rvu_t \mid
\rvx_0\right)$. We parameterize it, together with the representation policy, as
Gaussians:
\begin{align}
    q\left(\rvu_t \mid \rvx_0\right) &= \gN\left(\alpha_t
        f_{\mathrm{pre}}(\rvx_0),\; \beta^2
        \sigma_t^2 \mI\right), \label{eq:ext_posterior} \\
    p_\phi\left(\rvu_t \mid \rvx_t\right) &= \gN\left(\alpha_t h_\phi(\rvh_t),\;
        \beta^2 \sigma_t^2
        \mI\right), \qquad \rvh_t = \vv_\theta^{(\ell)}(\rvx_t, t).
            \label{eq:ext_prior_app}
\end{align}
Here $(\alpha_t, \sigma_t)$ is the interpolant schedule of
App.~\ref{sec:bg-diffusion}, shared between $\rvx_t$ and $\rvu_t$ so that the
representation shares the noise schedule of the state: $\rvu_t$ is pure noise at
the terminal time and approaches the clean features $f_{\mathrm{pre}}(\rvx_0)$
as $t \to 0$. On the generative side, $\rvh_t$ is the hidden state of the
denoiser $\vv_\theta$ after block $\ell$, $h_\phi$ is an MLP projector that maps
it to the feature dimension of the target, and we write $\vh_\phi(\rvx_t, t) =
h_\phi(\rvh_t)$ for the projection. Finally, $\beta > 0$ scales the standard
deviation of both policies, so that the two share their covariance. The more
expressive flow policy used in our experiments is described in
App.~\ref{sec:design}.

\paragraph{Generative transition and forward process.}
The generative transition is a Gaussian with fixed variance,
\begin{equation}
    \label{eq:transition}
    p_\theta\left(\rvx_{t-1} \mid \rvx_t, \rvu_t\right) =
        \gN\left(\vmu_\theta(\rvx_t, \rvu_t,
        t),\; \tilde{\sigma}_{t-1 \mid t}^2 \mI\right),
\end{equation}
where $\tilde{\sigma}_{t-1 \mid t}^2$ is the variance of the forward posterior
$q\left(\rvx_{t-1} \mid \rvx_t, \rvx_0\right)$. It differs from the standard
diffusion transition only in that the mean depends on $\rvu_t$. We parameterize
$\vmu_\theta$ through a velocity network $\vv_\theta(\rvx_t, \rvu_t, t)$ by
substituting the estimate of $\rvx_0$ implied by $\vv_\theta$ into the mean of
the forward posterior.

The forward process is the fixed Gaussian process of
App.~\ref{sec:bg-diffusion}, with marginals $q\left(\rvx_t \mid \rvx_0\right) =
\gN\left(\alpha_t \rvx_0, \sigma_t^2 \mI\right)$.

\paragraph{Training objective.} Since the reference and representation policies
in Eqs.~\ref{eq:ext_posterior} and~\ref{eq:ext_prior_app} are isotropic
Gaussians with the same covariance, the KL term of Eq.~\ref{eq:elbo} is
available in closed form,
\begin{equation}
    \label{eq:kl_closed}
    \KL\left(q\left(\rvu_t \mid \rvx_0\right) \| p_\phi\left(\rvu_t \mid
        \rvx_t\right)\right)
        = \frac{\mathrm{SNR}(t)}{2\beta^2} \left\| f_{\mathrm{pre}}(\rvx_0) -
            h_\phi(\rvh_t) \right\|_2^2,
        \qquad \mathrm{SNR}(t) = \frac{\alpha_t^2}{\sigma_t^2},
\end{equation}
Substituting it into Eq.~\ref{eq:elbo} gives, after reweighting the time steps,
the training objective in Eq.~\ref{eq:objective} (App.~\ref{app:objective} gives
the derivation). No gradient is propagated through $f_{\mathrm{pre}}(\rvx_0)$,
and $f_{\mathrm{pre}}$ is not required at sampling time: each step draws
$\rvu_t$ from the model's own policy $p_\phi\left(\rvu_t \mid \rvx_t\right)$.

\subsubsection{Simplified Training Objective}
\label{app:objective}

We derive Eq.~\ref{eq:objective} from the bound in Eq.~\ref{eq:app_elbo} under
the parameterization of App.~\ref{app:param}. Here, $\alpha_t$, $\sigma_t$,
$\dot\alpha_t$, $\dot\sigma_t$ denote the interpolant schedule and its time
derivative at the corresponding grid time. Since in our experiments the
reference policy depends on $\rvx_0$ alone, $q_{\theta^{\prime}}(\rvu_t \mid
\rvx_{\leq t}) = q(\rvu_t \mid \rvx_0)$, the alignment term depends on $(\rvx_0,
\rvx_t)$ only, and $\rvu_t$ is conditionally independent of $\rvx_{t-1}$ given
$(\rvx_t, \rvx_0)$.

\paragraph{Alignment term.} Both $q(\rvu_t \mid \rvx_0) = \gN(\alpha_t
f_{\mathrm{pre}}(\rvx_0), \beta^2 \sigma_t^2 \mI)$ and $p_\phi(\rvu_t \mid
\rvx_t) = \gN(\alpha_t \vh_\phi(\rvx_t, t), \beta^2 \sigma_t^2 \mI)$ are
isotropic Gaussians with the same covariance, so their KL divergence reduces to
the squared distance between the means scaled by the precision $1 / (\beta^2
\sigma_t^2)$,
\begin{equation}
    \KL\big( q(\rvu_t \mid \rvx_0) \,\|\, p_\phi(\rvu_t \mid \rvx_t) \big)
        = \frac{\mathrm{SNR}(t)}{2 \beta^2} \big\| f_{\mathrm{pre}}(\rvx_0) -
            \vh_\phi(\rvx_t, t)
        \big\|_2^2, \qquad \mathrm{SNR}(t) = \frac{\alpha_t^2}{\sigma_t^2}.
    \label{eq:app_kl_gauss}
\end{equation}
If the reference variance is chosen differently from $\beta^2 \sigma_t^2$, the
KL changes only by terms that do not depend on $\phi$, since the mean term
involves the variance of $p_\phi$ alone.

\paragraph{Denoising term.} The forward posterior of the Gaussian noising
process is
\citep{kingma2021vdm}
\begin{equation}
    q(\rvx_{t-1} \mid \rvx_t, \rvx_0) = \gN\big( \tilde{\vmu}_t(\rvx_t,
        \rvx_0),\;
        \tilde{\sigma}_{t-1 \mid t}^2 \mI \big), \qquad
    \tilde{\vmu}_t(\rvx_t, \rvx_0) = a_t \rvx_t + b_t \rvx_0,
    \label{eq:app_posterior}
\end{equation}
with $a_t = \alpha_{t \mid t-1} \sigma_{t-1}^2 / \sigma_t^2$, $b_t =
\alpha_{t-1} \sigma_{t \mid t-1}^2 / \sigma_t^2$ and $\tilde{\sigma}_{t-1 \mid
t}^2 = \sigma_{t \mid t-1}^2 \sigma_{t-1}^2 / \sigma_t^2$, where $\alpha_{t \mid
t-1} = \alpha_t / \alpha_{t-1}$ and $\sigma_{t \mid t-1}^2 = \sigma_t^2 -
\alpha_{t \mid t-1}^2 \sigma_{t-1}^2$ are the coefficients of the transition
$q(\rvx_t \mid \rvx_{t-1})$. For $t = 1$, $\rvx_0$ is observed and $(a_1, b_1) =
(0, 1)$; we then use a fixed decoder variance in place of $\tilde{\sigma}_{0
\mid 1}^2$. Since $p_\theta(\rvx_{t-1} \mid \rvx_t, \rvu_t)$ is a Gaussian with
the same variance, the expectation over $\rvx_{t-1}$ in the denoising term is
\begin{equation}
    \E_{q(\rvx_{t-1} \mid \rvx_t, \rvx_0)}\big[ \log p_\theta(\rvx_{t-1} \mid
        \rvx_t, \rvu_t) \big]
        = -\frac{1}{2 \tilde{\sigma}_{t-1 \mid t}^2} \big\|
            \tilde{\vmu}_t(\rvx_t, \rvx_0) -
        \vmu_\theta(\rvx_t, \rvu_t, t) \big\|_2^2 + \text{const},
    \label{eq:app_denoise_gauss}
\end{equation}
where the constant collects the entropy of the posterior and the normalizer of
$p_\theta$, neither of which depends on $\theta$. We parameterize $\vmu_\theta$
through the velocity network by substituting the estimate of $\rvx_0$ implied by
$\vv_\theta$ into the posterior mean. Solving $\rvx_t = \alpha_t \rvx_0 +
\sigma_t \rvepsilon$ and $\vv_{\mathrm{ref}} = \dot\alpha_t \rvx_0 +
\dot\sigma_t \rvepsilon$ for $\rvx_0$ gives
\begin{equation}
    \rvx_0 = \frac{\sigma_t \vv_{\mathrm{ref}} - \dot\sigma_t \rvx_t}{\sigma_t
        \dot\alpha_t -
        \dot\sigma_t \alpha_t}, \qquad
    \hat{\rvx}_\theta(\rvx_t, \rvu_t, t) := \frac{\sigma_t \vv_\theta(\rvx_t,
        \rvu_t, t) -
        \dot\sigma_t \rvx_t}{\sigma_t \dot\alpha_t - \dot\sigma_t \alpha_t},
    \label{eq:app_x0_param}
\end{equation}
and we set $\vmu_\theta(\rvx_t, \rvu_t, t) := \tilde{\vmu}_t\big(\rvx_t,
\hat{\rvx}_\theta(\rvx_t, \rvu_t, t)\big)$. For the straight-line interpolant,
$\sigma_t \dot\alpha_t - \dot\sigma_t \alpha_t = -1$. Since $\tilde{\vmu}_t$ is
affine in its second argument, Eq.~\ref{eq:app_x0_param} yields
$\tilde{\vmu}_t(\rvx_t, \rvx_0) - \vmu_\theta(\rvx_t, \rvu_t, t) = b_t (\rvx_0 -
\hat{\rvx}_\theta) = b_t \sigma_t (\vv_{\mathrm{ref}} - \vv_\theta) / (\sigma_t
\dot\alpha_t - \dot\sigma_t \alpha_t)$, and Eq.~\ref{eq:app_denoise_gauss}
becomes a weighted velocity regression,
\begin{align}
    \E_{q(\rvx_{t-1} \mid \rvx_t, \rvx_0)}\big[ \log p_\theta(\rvx_{t-1} \mid
        \rvx_t, \rvu_t) \big]
        &= -w_t \big\| \vv_\theta(\rvx_t, \rvu_t, t) - \vv_{\mathrm{ref}}
            \big\|_2^2 + \text{const},
    \label{eq:app_velocity} \\
    w_t &= \frac{b_t^2 \sigma_t^2}{2 \tilde{\sigma}_{t-1 \mid t}^2 (\sigma_t
        \dot\alpha_t -
        \dot\sigma_t \alpha_t)^2}. \notag
\end{align}

\paragraph{Combining.} Substituting Eqs.~\ref{eq:app_kl_gauss}
and~\ref{eq:app_velocity} into Eq.~\ref{eq:app_elbo} and writing the noising
process as $\rvx_t = \alpha_t \rvx_0 + \sigma_t \rvepsilon$ with $\rvepsilon
\sim \gN(\vzero, \mI)$, the negative log-likelihood is bounded by
\begin{equation}
    -\log p(\rvx_0) \leq T\, \E_{t, \rvepsilon,\, \rvu_t \sim
        q}\Big[
        w_t \big\| \vv_\theta(\rvx_t, \rvu_t, t) - \vv_{\mathrm{ref}} \big\|_2^2
        + \frac{\lambda_t}{2} \big\| f_{\mathrm{pre}}(\rvx_0) -
            \vh_\phi(\rvx_t, t)
        \big\|_2^2 \Big] + \text{const},
    \label{eq:app_weighted}
\end{equation}
where $\lambda_t = \mathrm{SNR}(t) / \beta^2$, $t$ is uniform on $\{1, \dots,
T\}$, and $\rvu_t \sim q(\rvu_t \mid \rvx_0)$. Following common practice
\citep{kingma2023understanding}, we replace the weight $w_t$ of the denoising
term by one, which changes the relative weighting of the time steps in the
denoising term. Since $T$ only scales the bound, we drop it, and we take the
continuous-time limit of the uniform grid by sampling $t \sim \gU(0, 1)$.
Dropping the constant and taking the expectation over $\rvx_0$ gives
Eq.~\ref{eq:objective}, which completes the derivation.

The alignment loss penalizes deviation of the representation policy from the
reference policy much more at higher SNR levels. Intuitively, the alignment loss
ensures that the representation policy recovers more semantic information when
the current state is cleaner (i.e., higher SNR values), thus encoding the
uncertainty in the diffusion process. Moreover, as $t \rightarrow 0$, the SNR
and therefore the alignment weight $\lambda_t$ grows unbounded. Therefore, we
instead use a bounded version (between 0 and 1) of the alignment weight. In
practice, we find that different choices of $\lambda_t$ like $\alpha_t^2 /
(\beta^2 (1 + \sigma_t^2))$ and $\mathrm{SNR}(t) / (\beta^2 + \mathrm{SNR}(t))$
work equally well without changing the semantics of the alignment loss
weighting. We choose the latter for the rest of our discussion.

\subsubsection{Derivation of the Generative ODE/SDE}
\label{app:ode_sde}

Here we derive the generative ODE and SDE for RAD. We first derive a forward SDE
corresponding to the noising process $q(\rvx_t \mid \rvx_0) = \gN(\alpha_t
\rvx_0, \sigma_t^2 \mI)$, then derive the reverse SDE based on the standard
result of \citet{anderson1982reverse} and \citet{song2021score}. We then
leverage the relationship between the score and the velocity field to express
the reverse SDE/ODE in terms of the velocity field. Finally, we

\paragraph{Forward SDE.} With the perturbation kernel $q(\rvx_t \mid \rvx_0) =
\gN(\alpha_t \rvx_0, \sigma_t^2 \mI)$, where $\alpha_0 = 1$ and $\sigma_0 = 0$,
the forward process is given by the linear SDE
\begin{equation}
    \mathrm{d}\rvx_t = f_t \rvx_t \,\mathrm{d}t + g_t \,\mathrm{d}\rvw_t,
    \qquad f_t = \frac{\dot\alpha_t}{\alpha_t}, \qquad
    g_t^2 = \frac{\mathrm{d}\sigma_t^2}{\mathrm{d}t} - 2 f_t \sigma_t^2
        = -\sigma_t^2 \frac{\mathrm{d}}{\mathrm{d}t} \log \mathrm{SNR}(t),
    \label{eq:fwd_sde}
\end{equation}
where $\rvw_t$ is a standard Wiener process and $g_t^2 \geq 0$ since
$\mathrm{SNR}(t)$ is decreasing in $t$. These coefficients are standard for
Gaussian perturbation kernels \citep{kingma2021vdm}.

\textit{Proof sketch.} Since Eq.~\ref{eq:fwd_sde} is linear with a deterministic
initial condition $\rvx_0$, its solution is Gaussian. It therefore suffices to
show that its mean $\vm_t = \E[\rvx_t]$ and covariance $\mSigma_t = \E[(\rvx_t -
\vm_t)(\rvx_t - \vm_t)^\top]$ equal $\alpha_t \rvx_0$ and $\sigma_t^2 \mI$. For
a linear SDE $\mathrm{d}\rvx_t = \mF_t \rvx_t \,\mathrm{d}t + \mL_t
\,\mathrm{d}\rvw_t$, these moments satisfy the ODEs \citep{sarkka2019applied}
\begin{equation}
    \frac{\mathrm{d}\vm_t}{\mathrm{d}t} = \mF_t \vm_t, \qquad
    \frac{\mathrm{d}\mSigma_t}{\mathrm{d}t} = \mF_t \mSigma_t + \mSigma_t
        \mF_t^\top + \mL_t \mL_t^\top.
    \label{eq:moment_odes}
\end{equation}
For completeness, we derive them for Eq.~\ref{eq:fwd_sde}, where $\mF_t = f_t
\mI$ and $\mL_t = g_t \mI$. For the mean,
\begin{equation*}
    \frac{\mathrm{d}}{\mathrm{d}t} \big( \alpha_t \rvx_0 \big) = \dot\alpha_t
        \rvx_0 = \frac{\dot\alpha_t}{\alpha_t}\, \alpha_t \rvx_0 \qquad \alpha_0
            \rvx_0 = \rvx_0.
\end{equation*}
For the covariance, by the definition of $g_t^2$ in Eq.~\ref{eq:fwd_sde},
\begin{equation*}
    \frac{\mathrm{d}}{\mathrm{d}t} \big( \sigma_t^2 \mI \big) =
        \frac{\mathrm{d}\sigma_t^2}{\mathrm{d}t} \mI = \big( 2 f_t \sigma_t^2 +
        g_t^2 \big) \mI = 2 f_t\, (\sigma_t^2 \mI) + g_t^2 \mI, \qquad
        \sigma_0^2 \mI = \vzero.
\end{equation*}
Both ODEs are linear with the same initial conditions as $\vm_t$ and
$\mSigma_t$, so their solutions are unique, and $\vm_t = \alpha_t \rvx_0$ and
$\mSigma_t = \sigma_t^2 \mI$. Hence $q(\rvx_t \mid \rvx_0) = \gN(\alpha_t
\rvx_0, \sigma_t^2 \mI)$.

\paragraph{Reverse SDE and probability flow ODE.} Following
\citet{anderson1982reverse} and \citet{song2021score}, the forward process in
Eq.~\ref{eq:fwd_sde} has the reverse-time SDE
\begin{equation}
    \mathrm{d}\rvx_t = \big[ f_t \rvx_t - g_t^2 \nabla_{\rvx_t} \log q_t(\rvx_t)
        \big] \mathrm{d}t + g_t \,\mathrm{d}\bar{\rvw}_t,
    \label{eq:rev_sde}
\end{equation}
and the probability flow ODE
\begin{equation}
    \frac{\mathrm{d}\rvx_t}{\mathrm{d}t} = f_t \rvx_t - \tfrac{1}{2} g_t^2
        \nabla_{\rvx_t} \log q_t(\rvx_t),
    \label{eq:pf_ode}
\end{equation}
where $\bar{\rvw}_t$ is a standard Wiener process in reverse time and
$\mathrm{d}t < 0$. More generally, for any diffusion coefficient $\omega_t \geq
0$, the reverse-time SDE
\begin{equation}
    \mathrm{d}\rvx_t = \big[ f_t \rvx_t - \tfrac{1}{2} (g_t^2 + \omega_t)
        \nabla_{\rvx_t} \log q_t(\rvx_t) \big] \mathrm{d}t + \sqrt{\omega_t}
        \,\mathrm{d}\bar{\rvw}_t
    \label{eq:rev_sde_family}
\end{equation}
has the same marginals \citep{albergo2023stochastic,ma2024sit}, since the
additional drift $-\tfrac{1}{2} \omega_t \nabla \log q_t$ and the additional
diffusion $\sqrt{\omega_t}$ cancel in the Fokker--Planck equation. Setting
$\omega_t = g_t^2$ recovers Eq.~\ref{eq:rev_sde}, and $\omega_t = 0$ recovers
Eq.~\ref{eq:pf_ode}.

\paragraph{Expressing the score in terms of the velocity.} Following
\citet{albergo2023stochastic}, the score can be expressed through the velocity
field. Since $\rvx_t = \alpha_t \rvx_0 + \sigma_t \rvepsilon$, Tweedie's formula
gives
\begin{equation*}
    \nabla_{\rvx_t} \log q_t(\rvx_t) = -\frac{\E[\rvepsilon \mid
        \rvx_t]}{\sigma_t}.
\end{equation*}
The velocity field is $\vv(\rvx_t, t) = \E[\dot\alpha_t \rvx_0 + \dot\sigma_t
\rvepsilon \mid \rvx_t]$. Substituting $\rvx_0 = (\rvx_t - \sigma_t \rvepsilon)
/ \alpha_t$ gives
\begin{equation*}
    \vv(\rvx_t, t) = f_t \rvx_t + (\dot\sigma_t - f_t \sigma_t)\, \E[\rvepsilon
        \mid \rvx_t],
\end{equation*}
so $\E[\rvepsilon \mid \rvx_t] = (\vv(\rvx_t, t) - f_t \rvx_t) / (\dot\sigma_t -
f_t \sigma_t)$. Combining the two and using $g_t^2 = 2 \sigma_t (\dot\sigma_t -
f_t \sigma_t)$ from Eq.~\ref{eq:fwd_sde},
\begin{equation}
    \nabla_{\rvx_t} \log q_t(\rvx_t) = -\frac{\vv(\rvx_t, t) - f_t
        \rvx_t}{\sigma_t
        (\dot\sigma_t - f_t \sigma_t)} = \frac{2}{g_t^2} \big( f_t \rvx_t -
        \vv(\rvx_t, t) \big).
    \label{eq:score_velocity}
\end{equation}
Substituting Eq.~\ref{eq:score_velocity} into Eq.~\ref{eq:rev_sde_family}, the
drift becomes $f_t \rvx_t - \tfrac{1}{2} (g_t^2 + \omega_t) \cdot
\frac{2}{g_t^2} (f_t \rvx_t - \vv) = (1 + \omega_t / g_t^2)\, \vv - (\omega_t /
g_t^2) f_t \rvx_t$, which gives the generative SDE in terms of the marginal
velocity,
\begin{equation}
    \mathrm{d}\rvx_t = \Big[ \Big( 1 + \frac{\omega_t}{g_t^2} \Big)\,
        \vv(\rvx_t, t) - \frac{\omega_t}{g_t^2} f_t \rvx_t \Big] \mathrm{d}t +
        \sqrt{\omega_t} \,\mathrm{d}\bar{\rvw}_t,
    \label{eq:sde_velocity}
\end{equation}
integrated from $t = 1$ to $t = 0$ with $\mathrm{d}t < 0$. For $\omega_t =
g_t^2$ this is $\mathrm{d}\rvx_t = [2 \vv(\rvx_t, t) - f_t \rvx_t] \mathrm{d}t +
g_t \,\mathrm{d}\bar{\rvw}_t$, and for $\omega_t = 0$ it is the probability flow
ODE
\begin{equation}
    \frac{\mathrm{d}\rvx_t}{\mathrm{d}t} = \vv(\rvx_t, t).
    \label{eq:ode_velocity}
\end{equation}

\paragraph{Marginal velocity from the conditional velocity of RAD.} RAD does not
learn the marginal velocity $\vv(\rvx_t, t)$ directly. Instead, it learns a
velocity $\vv_\theta(\rvx_t, \rvu_t, t)$ conditioned on the representation, and
a policy $p_\phi(\rvu_t \mid \rvx_t)$. We now relate the two. The velocity
regression in Eq.~\ref{eq:objective}, where $\rvu_t \sim q(\rvu_t \mid \rvx_0)$,
is minimized by the conditional velocity
\begin{equation*}
    \vv^*(\rvx_t, \rvu_t, t) = \E[\vv_{\mathrm{ref}} \mid \rvx_t, \rvu_t].
\end{equation*}
By the tower property,
\begin{equation}
    \vv(\rvx_t, t) = \E[\vv_{\mathrm{ref}} \mid \rvx_t] = \int \E[
        \vv_{\mathrm{ref}} \mid \rvx_t, \rvu_t]\, q_t(\rvu_t \mid
        \rvx_t)\, \mathrm{d}\rvu_t = \E_{q_t(\rvu_t \mid \rvx_t)}\big[
            \vv^*(\rvx_t,
        \rvu_t, t) \big],
    \label{eq:marginal_conditional_velocity}
\end{equation}
where $q_t(\rvu_t \mid \rvx_t) = \int q(\rvu_t \mid \rvx_0)\, q(\rvx_0 \mid
\rvx_t)\, \mathrm{d}\rvx_0$, since $\rvu_t$ is independent of $\rvx_t$ given
$\rvx_0$. This posterior is also the target of the policy. Using the same
conditional independence, the alignment term decomposes as
\begin{equation*}
    \E_{q(\rvx_0 \mid \rvx_t)}\big[ \KL(q(\rvu_t \mid \rvx_0) \,\|\,
        p_\phi(\rvu_t \mid \rvx_t)) \big] = -\E_{q_t(\rvu_t \mid
        \rvx_t)}\big[ \log p_\phi(\rvu_t \mid \rvx_t) \big] + \text{const},
\end{equation*}
where the constant does not depend on $\phi$. By Gibbs' inequality, it is
minimized over all conditional densities by $p_\phi(\rvu_t \mid \rvx_t) =
q_t(\rvu_t \mid \rvx_t)$. Hence, when $\vv_\theta$ and $p_\phi$ approximate
these optima, Eq.~\ref{eq:marginal_conditional_velocity} gives
\begin{equation}
    \vv(\rvx_t, t) \approx \vv_{\theta, \phi}(\rvx_t, t) = \E_{\rvu_t \sim
        p_\phi(\rvu_t \mid \rvx_t)}\big[ \vv_\theta(\rvx_t, \rvu_t, t) \big].
    \label{eq:velocity_rad}
\end{equation}
Substituting Eq.~\ref{eq:velocity_rad} into Eq.~\ref{eq:ode_velocity} recovers
the RAD sampler of Eq.~\ref{eq:rad_sampling}, and into
Eq.~\ref{eq:sde_velocity} gives the generative SDE.

\subsubsection{Practical Design Choices}
\label{sec:design}

\textbf{Choice of policy.} We refer to the representation policy
$p_\phi\left(\rvu_t \mid \rvx_t\right)$ simply as the policy. Under the Gaussian
parameterization of Eq.~\ref{eq:ext_prior}, a sample takes the form $\rvu_t =
\alpha_t h_\phi(\rvh_t) + \beta \sigma_t \bar{\rvepsilon}$ with
$\bar{\rvepsilon} \sim \gN(\vzero, \mI)$. However, a Gaussian policy is unimodal
which limits expressivity. To obtain a more expressive policy, we replace the
noise with the output of a conditional normalizing flow \citep{dinh2017realnvp},
\begin{equation}
    \label{eq:flow_policy}
    \rvu_t = \alpha_t h_\phi(\rvh_t) + \beta \sigma_t\, \vg_\zeta(\rvh_t,
        \bar{\rvepsilon}),
        \qquad \bar{\rvepsilon} \sim \gN(\vzero, \mI),
\end{equation}
where $\vg_\zeta(\rvh_t, \bar{\rvepsilon})$ is a stack of affine coupling layers
whose conditioning is a linear map of $h_\phi(\rvh_t)$, and $\zeta$ is trained
jointly with $\phi$. Consequently,
\begin{equation}
    \bar{\rvepsilon} = \vg_\zeta^{-1}\Big(\rvh_t, \frac{\rvu_t - \alpha_t
        h_\phi(\rvh_t)}{\beta \sigma_t}\Big).
    \label{eq:flow_policy_inv}
\end{equation}
Intuitively, the second argument in Eq.~\ref{eq:flow_policy_inv} represents the
whitened version of $\rvu_t$ which we denote as $\rvz_t$, for brevity. Next, the
KL term of Eq.~\ref{eq:elbo} can be simplified as,
\begin{equation}
\KL\left(q\left(\rvu_t \mid \rvx_0\right) \| p_\phi\left(\rvu_t
    \mid\rvx_t\right)\right)
= -\E_{\rvu_t \sim q}\big[ \log p_\phi(\rvu_t \mid \rvx_t) \big] + \text{const}.
\end{equation}
The above result follows directly from the definition of the KL divergence. The
additional constant term is the negative entropy of the reference policy, which
does not depend on the parameters of the representation policy and can be
ignored during optimization. The KL can be further simplified as,
\begin{align}
-\E_{\rvu_t \sim q}\Big[ \log p_\phi(\rvu_t \mid \rvx_t) \Big]
&= -\E_{\rvu_t \sim q}\Big[ \log p_{\bar{\rvepsilon}}(\bar{\rvepsilon} =
    \vg_\zeta^{-1}(\rvh_t, \rvz_t)) + \log \Big| \det
            \tfrac{\partial
        \vg_\zeta^{-1}}{\partial \rvz_t} \Big|\Big] + \text{const},\\
&= \E_{\rvu_t \sim q}\Big[ \tfrac{1}{2}
        \big\|
        \vg_\zeta^{-1}(\rvh_t, \rvz_t) \big\|_2^2 - \log \Big| \det
            \tfrac{\partial
        \vg_\zeta^{-1}}{\partial \rvz_t} \Big| \Big] + \text{const},
\end{align}
where the constants collect $d \log(\beta \sigma_t)$ from the change of
variables from $\rvu_t$ to $\rvz_t$ and the normalizer of
$p_{\bar{\rvepsilon}}$, neither of which depends on $(\phi, \zeta)$. Therefore,
\begin{equation}
    \label{eq:flow_nll}
    \ell_{\mathrm{flow}}(\phi, \zeta) = \E_{\rvu_t \sim q}\Big[ \tfrac{1}{2}
        \big\|
        \vg_\zeta^{-1}(\rvh_t, \rvz_t) \big\|_2^2 - \log \Big| \det
            \tfrac{\partial
        \vg_\zeta^{-1}}{\partial \rvz_t} \Big| \Big], \qquad
    \rvz_t = \frac{\rvu_t - \alpha_t h_\phi(\rvh_t)}{\beta \sigma_t},
\end{equation}
At the start of training, the flow is initialized to the identity, i.e.,
$\vg_\zeta^{-1}(\rvh_t, \rvz_t) = \rvz_t$. The log-determinant then vanishes
and, in expectation over $\rvu_t$, the first term equals the alignment term of
Eq.~\ref{eq:kl_closed} up to an additive constant. In practice, we find that
keeping the explicit alignment term helps training stability. Therefore, we
combine the flow and Gaussian policies.

\paragraph{Combining the Gaussian and flow policies.} Let $p^{\gN}_\phi$ denote
the Gaussian policy of Eq.~\ref{eq:ext_prior} and $p^{\mathrm{flow}}_{\phi,
\zeta}$ the flow policy of Eq.~\ref{eq:flow_policy}. We keep the explicit
alignment term of the Gaussian policy when training the flow policy. Since
$\KL(q \,\|\, p^{\gN}_\phi) \geq 0$, with $q = q(\rvu_t \mid \rvx_0)$, it
follows,
\begin{equation}
    \KL(q \,\|\, p^{\mathrm{flow}}_{\phi, \zeta}) \leq \KL(q \,\|\,
        p^{\gN}_\phi) + \KL(q \,\|\, p^{\mathrm{flow}}_{\phi, \zeta}),
\end{equation}
so the result is still a lower bound on the log-likelihood of the model with
the flow policy, which is the model we sample from. By
Eqs.~\ref{eq:kl_closed} and~\ref{eq:flow_nll},
\begin{equation}
    \label{eq:comb_kl}
    \KL\big(q \,\|\, p^{\gN}_\phi\big) + \KL\big(q \,\|\,
        p^{\mathrm{flow}}_{\phi, \zeta}\big)
    = \frac{\mathrm{SNR}(t)}{2 \beta^2} \big\| f_{\mathrm{pre}}(\rvx_0) -
        h_\phi(\rvh_t) \big\|_2^2 + \ell_{\mathrm{flow}}(\phi, \zeta) +
        \text{const}.
\end{equation}
As for the Gaussian policy (App.~\ref{app:objective}), we set the weight of the
denoising term to one, and weight the KL term by $1 - \lambda_t$ with
$\lambda_t = \mathrm{SNR}(t) / (\mathrm{SNR}(t) + \beta^2)$, the policy's share
of the variance of $\rvu_t$. This removes the factor $\mathrm{SNR}(t)$, which
is unbounded as $t \to 0$, since $(1 - \lambda_t)\, \mathrm{SNR}(t) / (2
\beta^2) = \lambda_t / 2$, and gives the objective
\begin{align}
    \label{eq:flow_objective}
    \Ls(\theta, \phi, \zeta) = \E\Big[ \big\| \vv_\theta(\rvx_t, \rvu_t, t) -
        (\dot\alpha_t
        \rvx_0 + \dot\sigma_t \rvepsilon) \big\|_2^2
        &+ \frac{\lambda_t}{2}\, \big\| f_{\mathrm{pre}}(\rvx_0) -
            h_\phi(\rvh_t) \big\|_2^2
        + (1 - \lambda_t)\, \ell_{\mathrm{flow}}(\phi, \zeta) \Big],
\end{align}
where $\rvu_t \sim q(\rvu_t \mid \rvx_0)$ is drawn from the reference policy as
before, the outer expectation is over $t$, $\rvx_0$ and the noise, and
$h_\phi(\rvh_t)$ is not detached in $\ell_{\mathrm{flow}}$, so both terms train
the projector. The weight $1 - \lambda_t$ vanishes at $t = 0$, where the policy
is deterministic and the likelihood degenerates, and approaches one as $t \to
1$. At sampling, we draw $\rvu_t$ from the flow policy. During training, we fix $\beta=1.0$
and vary it during inference to control the temperature of the policy which
improves sample quality without affecting diversity (Fig.~\ref{fig:beta_inf}).

\figArchitecture

\textbf{Architecture.} We build on SiT. We first compute the hidden state
$\rvh_t$, from which we draw $\rvu_t \in \R^{N \times D}$ from the
representation policy $p_\phi(\rvu_t\mid\rvx_t)$. We then process the controls
as follows before feeding it to the denoiser.
\begin{itemize}
    \item \textbf{Control Normalization.}
        \begin{equation*}
            \rvu_t^{n} = \frac{\rvu_t - \gamma \bar{u}_t}{\bar{\sigma}_t +
                \delta},
        \end{equation*}
        where $\bar{u}_t$ and $\bar{\sigma}_t$ are the mean and standard
        deviation of $\rvu_t$ across its $N$ tokens, $\gamma \in [0, 1]$
        controls how much of the mean is removed, and $\delta$ is a small
        constant. This helps remove global representation information from the
        injected representation which helps with downstream sample quality
        \citep{singh2026irepa}. Throughout this work, we set $\gamma = 0.6$ and
        $\delta = 10^{-6}$.
    \item \textbf{Compression.} We then aggregate the normalized representation
        along the token axis using an MLP of $\rvu_t^{n}$ to $M$ tokens,
        \begin{equation*}
            \rvu_t^{c} = \mathrm{MLP}(\rvu_t^{n}) \in \R^{M \times D},
        \end{equation*}
        where we use $M = 1$. A linear control projection then maps $\rvu_t^{c}$
        from the encoder dimension $D$ to the embedding dimension of the
        backbone, so that it can be added to the hidden state.
    \item \textbf{Modulation.}
        \begin{equation*}
            \rvu_t^{f} = \big(1 + \va(\vc)\big) \odot \rvu_t^{c} + \vb(\vc),
        \end{equation*}
        where $\va$ and $\vb$ are linear functions of the conditioning embedding
        $\vc$, as in adaLN, and are zero-initialized so that training starts
        from the unmodulated representation.
\end{itemize}
The result is added to every token of the hidden state, $\bar{\rvh}_t^{[n]} =
\rvh_t^{[n]} + \rvu_t^{f}$. We inject early, with $\ell$ within the first third
of the backbone, which we find to be the most influential architectural choice
(Table~\ref{tab:design_choices}).

\subsection{Related Work}
\label{sec:related}

\subsubsection{Contextualizing RAD within Generative Modeling}

\paragraph{External Representation Alignment} methods align the intermediate
representations of the diffusion denoiser with the features of a frozen
pretrained vision encoder like DINOv2 \citep{oquab2024dinov2}. REPA
\citep{yu2025repa} pulls the hidden states of a diffusion transformer toward the
DINOv2 features of the clean image, which leads to an order-of-magnitude gain in
training efficiency. Consequently, different aspects of external alignment have
received a lot of recent interest. iREPA \citep{singh2026irepa} shows that the
spatial structure of the target features, rather than their global semantic
content, drives these gains, and speeds up convergence by replacing the
projector $h_\phi$ with a convolutional layer and normalizing the target tokens
across space to suppress their shared global component. REPA-E
\citep{leng2025repae} trains the VAE and the denoiser end to end using the REPA
loss. In all these methods, the representation acts as a training-time
regularizer, and the learned representations are never injected back into the
denoiser during sampling. In contrast, RAD additionally conditions every
denoising step on a latent $\rvu_t$ drawn from the representation policy
$p_\phi(. \mid \rvx_t)$, and the alignment loss arises as a KL term in a
likelihood bound instead of as a heuristic addition to the denoising loss.
Another line of work aligns the autoencoder's latent space with the encoder
features \citep{yao2025vavae} or replaces the autoencoder with a frozen encoder
and a trained decoder \citep{zheng2025rae}. Exploring design choices related to
the VAE and other architectural modifications as proposed in these works in the
context of RAD could be an interesting direction for future work.

Closest to our work, REED~\citep{wang2025reed} also proposes a general
variational framework for representation alignment. The main idea is to
introduce a \emph{global representation} $\rvz$, sampled once along the reverse
process, on which subsequent denoising steps are conditioned. However, in
practice, REED drops this explicit conditioning (along with a normalization
term), reducing its objective to the standard REPA objective with multimodal
alignment targets and a training curriculum. This is supported by an ablation
(their Table~10) in which the denoiser is conditioned on the ground-truth
encoder representation during training but sampled from the unconditional
branch; this variant performs on par with REPA. In contrast, RAD samples a
latent $\rvu_t$ for each transition from a learned policy whose noise level
matches that of the state, and conditions the denoiser on it explicitly. Because
$\rvu_t$ is stochastic, each transition is a continuous mixture of Gaussians,
whereas REED's transitions at sampling time, conditioned implicitly on a point
estimate, remain single Gaussians. Our bound has no normalization term and
recovers REPA as an exact special case when the denoiser ignores $\rvu_t$, using
the same von Mises--Fisher argument as in \citet{wang2025reed} (see also
App.~\ref{app:prior_work} for a more formal discussion).

Finally, another line of work makes the encoder representation part of the
diffusion state and denoises it jointly with the image latents:
REG~\citep{wu2025reg} denoises the encoder's class token alongside the image
tokens, and ReDi~\citep{kouzelis2026boosting} jointly diffuses VAE latents and
dimension-reduced DINOv2 features. Here the representation follows its own
noising trajectory and is generated with the image. In RAD, the
representation is not part of the diffusion state: a fresh latent $\rvu_t$ is
predicted from the current state at every step and conditions only that
transition. The two are complementary, and combining RAD with REG further
improves convergence (Sec.~\ref{sec:generalization}).

\paragraph{Internal Representation Alignment} methods remove the external
encoder and let the model supervise itself. MaskDiT \citep{zheng2024maskdit}
drops a subset of tokens and asks the network to reconstruct them while
denoising. SRA \citep{jiang2025sra} aligns the features of an earlier block at a
higher noise level with those of a deeper block of an EMA teacher at a lower
noise level. Another line of work \citep{zhu2024sddit, chefer2026selfflow}
noises the student's tokens and pulls its features toward an EMA teacher that
sees the cleaner view, which yields a self-supervised signal. SRA-like methods
share the design of external alignment: an auxiliary loss on the hidden states
that leaves the generative model unchanged. While the RAD framework allows for
incorporating internal representation alignment via parameterization of the
reference policy $q_{\theta^{\prime}}$, in our initial experiments, we find that
adapting SRA with RAD yields only modest gains (2--3 FID points on the B/2 model
scale for class-conditional ImageNet generation) and therefore restrict our
discussion of RAD to external alignment methods. However, adapting RAD for
internal alignment would be an interesting direction for further work. Lastly,
approaches based on dispersive loss \citep{wang2025dispersive} instead spread
the hidden states apart with a contrastive-style regularizer that needs no
positive pairs, which could be interesting to explore in the context of RAD.

\paragraph{Combining VAEs with Diffusion Models.}
LSGM \citep{vahdat2021lsgm} and D2C \citep{sinha2021d2c} train a VAE with a
diffusion prior over its latents. Diffusion autoencoders
\citep{preechakul2022diffae}, PDAE \citep{zhang2022pdae}, and InfoDiffusion
\citep{wang2023infodiffusion} learn a global latent with an auxiliary encoder
and condition the entire denoising trajectory on it, whereas DiffuseVAE
\citep{pandey2022diffusevae} pairs a VAE with a diffusion decoder that refines
its output. DisCo-Diff \citep{xu2024discodiff} similarly augments the denoiser
with a few discrete global latents, inferred by an encoder that is trained end
to end with the denoiser, to simplify the noise-to-data mapping, and fits an
autoregressive prior over these latents in a second stage. More recently, RCG
\citep{li2024rcg} generates a self-supervised representation with a diffusion
prior and conditions image generation on it. RAD shares the idea of a latent
that steers generation but differs in two respects. First, the latent $\rvu_t$
is per-step with its own prior $p_\phi(. \mid \rvx_t)$ conditioned on the
current state, so the model is a conditional VAE \citep{sohn2015cvae} at every
transition rather than a single global code. Second, the per-step prior $p_\phi$
is trained jointly with the denoiser under a single bound, whose KL term
recovers the representation-alignment loss. Methods like RCG, DisCo-Diff, and
DiffuseVAE instead train the latent prior or the image generator in a separate
stage, and DisCo-Diff learns its latents from scratch rather than aligning them
with a pretrained representation. Another work DDGAN \citep{xiao2022ddgan}
models each denoising transition with a conditional GAN so that the transition
can be multimodal and large denoising steps remain accurate. RAD shares this
motivation (Eq.~\ref{eq:marginalized_transition}), but its per-step latent has a
learned, state-conditioned prior, is trained under a variational bound rather
than adversarially, and is aligned with a semantic representation rather than
being unstructured noise.

\paragraph{Connections to other diffusion model families.} Similar to RAD, 
diffusion models such as CLD
 \citep{dockhorn2022scorebased} and PSLD \citep{pandey2023complete} extend the
 diffusion state with auxiliary velocity or momentum variables that are coupled
 to the data through a fixed forward SDE, which smooths the diffusion paths,
 simplifies score learning, and improves sample quality at a given sampling
 budget. As in REG and ReDi, these auxiliary variables are part of the diffusion
 state and are carried across steps. In contrast, RAD's latent $\rvu_t$ is drawn
 at every step from a learned, state-conditioned policy and aligned with a
 semantic representation.

 Interestingly, scale mixtures of Gaussians exhibit heavy-tailed behavior
 \citep{andrews1974scale}. Therefore, extending RAD so that the policy also
 modulates the transition variance, which yields a scale mixture, could model
 heavy-tailed behavior in scientific applications \citep{pandey2025heavytailed}
 and could be an interesting direction for future work.

\subsubsection{Prior Work as Special Cases under RAD}
\label{app:prior_work}
Prior work on representation alignment can be broadly categorized by the choice
of the reference policy $q_{\theta^{\prime}}(\rvu_t \mid \rvx_{\leq t})$: an
external pretrained model such as DINOv2 (referred to as external alignment) or
the denoiser itself (internal alignment). Our framework accommodates both; they
differ only in the reference policy.

\paragraph{External alignment.}
In external alignment, a frozen pretrained encoder $f_{\mathrm{pre}}$ such as
DINOv2 extracts the target representation from the clean image $\rvx_0$, so the
reference policy depends on $\rvx_0$ alone, $q_{\theta^{\prime}}\left(\rvu_t
\mid \rvx_{\leq t}\right) = q\left(\rvu_t \mid \rvx_0\right)$. With this
assumption, REPA is recovered by two choices. First, we parameterize
the denoiser to ignore $\rvu_t$, $p_\theta\left(\rvx_{t-1} \mid \rvx_t,
\rvu_t\right) = p_\theta\left(\rvx_{t-1} \mid \rvx_t\right)$, so that the
denoising term of Eq.~\ref{eq:elbo} reduces to the velocity-matching objective
of Eq.~\ref{eq:fm}. Second, we parameterize both the reference and the
representation policies as von Mises--Fisher (vMF) distributions on
the unit sphere, token by token, with a shared concentration $\kappa > 0$,
\begin{equation}
    \label{eq:vmf}
    q\left(\rvu_t^{[n]} \mid \rvx_0\right) = \mathrm{vMF}\left(\bar{\rvy}^{[n]},
        \kappa\right),
    \qquad
    p_\phi\left(\rvu_t^{[n]} \mid \rvx_t\right) =
        \mathrm{vMF}\left(\bar{h}_\phi(\rvh_t)^{[n]},
        \kappa\right),
\end{equation}
where $\rvy = f_{\mathrm{pre}}(\rvx_0)$, $\rvh_t$ is the hidden state of the
denoiser after block $\ell$, $h_\phi$ is the projector of
Eq.~\ref{eq:ext_prior_app}, and the bar denotes $\ell_2$ normalization. Since
$\mathrm{vMF}(\vmu, \kappa) \propto \exp(\kappa\, \vmu^\top \rvu)$, the KL
divergence between two vMF distributions with the same concentration can be
stated as,
\begin{equation}
    \label{eq:vmf_kl}
    \KL\left(\mathrm{vMF}(\vmu_1, \kappa) \,\|\, \mathrm{vMF}(\vmu_2,
        \kappa)\right)
        \propto \left(1 - \vmu_1^\top \vmu_2\right).
\end{equation}
Summing over tokens, the alignment term of Eq.~\ref{eq:elbo} becomes $\sum_n
\big(1 - \cos(\rvy^{[n]}, h_\phi(\rvh_t)^{[n]})\big)$, which is REPA's cosine
alignment loss of Eq.~\ref{eq:repa} up to an additive constant and a positive
scale, which plays the role of $\lambda$. The bound therefore reduces to the
REPA objective and REPA becomes a special case in our framework.

\paragraph{Internal alignment.}
\label{sec:internal}
In this setting, the representations are produced by the denoiser itself, as in
SRA. The target is the hidden state
$\vv_{\theta^{\prime}}^{(\ell^{\prime})}(\rvx_{t-k}, t-k)$ after block
$\ell^{\prime}$ of an EMA copy
$\vv_{\theta^{\prime}}$ of the denoiser, evaluated at a less noisy state
$\rvx_{t-k}$, $k \in \{1, \dots, t\}$, of the same trajectory. Since the target
depends on $\rvx_{t-k}$ alone, the reference policy reduces to
$q_{\theta^{\prime}}\left(\rvu_t \mid \rvx_{\leq t}\right) =
q_{\theta^{\prime}}\left(\rvu_t \mid \rvx_{t-k}\right)$, which we parameterize
as
\begin{equation}
    \label{eq:int_posterior}
    q_{\theta^{\prime}}\left(\rvu_t \mid \rvx_{t-k}\right)
        = \gN\left(\alpha_t \vv_{\theta^{\prime}}^{(\ell^{\prime})}(\rvx_{t-k},
            t-k),\; \beta^2
            \sigma_t^2 \mI\right).
\end{equation}
The parameters $\theta^{\prime}$ track $\theta$ through the EMA update and
receive no gradient. The representation policy $p_\phi\left(\rvu_t \mid
\rvx_t\right)$ is unchanged from Eq.~\ref{eq:ext_prior_app}: the model at the
noisier state $\rvx_t$ predicts the features that its EMA copy extracts from the
cleaner state $\rvx_{t-k}$. For $k = t$ the target is computed from the clean
image, as in the external setting, with
$\vv_{\theta^{\prime}}^{(\ell^{\prime})}(\rvx_0, 0)$ in place of
$f_{\mathrm{pre}}(\rvx_0)$.

\subsection{Experimental Details}
\label{app:exp_details}

\paragraph{Data and latents.} For class-conditional generation we use the
ImageNet training set~\citep{deng2009imagenet} of 1.28M images at
$256\times256$, center-cropped as in ADM~\citep{dhariwal2021adm}, without
further augmentation. Following REPA, images are encoded once with the Stable
Diffusion VAE~\citep{rombach2022ldm} into $32\times32\times4$ latents. At every
training step, a latent is drawn from the stored posterior and scaled by
$0.18215$. The external supervisor $f_{\mathrm{pre}}$ is a frozen
DINOv2-B/14~\citep{oquab2024dinov2}: the $256\times256$ image is normalized and
resized to $224\times224$, and the 256 patch tokens of dimension 768 (the class
token is dropped) serve both as alignment targets and as the clean
representation $f_{\mathrm{pre}}(\rvx_0)$ from which the training-time controls
$\rvu_t$ are noised. For the encoder ablation (Table~\ref{tab:design_choices}),
DINOv2-L and SigLIP-L~\citep{zhai2023siglip} use the standard preprocessing of
the respective models.

\paragraph{Backbones and injection depth.} We use the unmodified SiT-B/2 (12
blocks, width 768), SiT-L/2 (24 blocks, width 1024), and SiT-XL/2 (28 blocks,
width 1152) architectures with a $2\times2$ patch size, so that the latent is
processed as 256 tokens. The hidden state after block $\ell$ is aligned with the
encoder and, for RAD, $\rvu_t$ is injected after the same block. RAD-B/2 uses
$\ell = 2$ for sample quality (Table~\ref{tab:design_choices}) and $\ell = 4$
for linear probing. RAD-L/2 and RAD-XL/2 use $\ell = 8$, the depth at which REPA
aligns these backbones. In case the control dimension differs from the backbone
width (for L/2 or XL/2 backbones), we use a linear layer to map the controls to
the backbone width.

\paragraph{Optimization.} Following REPA, all RAD models
are trained with AdamW~\citep{loshchilov2018decoupled} at a constant learning
rate of $10^{-4}$ without warmup, $(\beta_1, \beta_2) = (0.9, 0.999)$, no weight
decay, gradient clipping at norm 1, and a global batch size of 256, in fp16
mixed precision, for 400K iterations unless stated otherwise. We keep an
exponential moving average (EMA) of the weights with decay $0.9999$, updated
every step, and use them for evaluation. The noising process is
the linear interpolant $\alpha_t = 1 - t$, $\sigma_t = t$ with $t \sim \gU(0,
1)$, the denoiser predicts the velocity, and the class label is dropped with
probability $0.1$. We use random seed 0 throughout.

\paragraph{RAD hyperparameters.} We highlight different RAD hyperparameters in
Table~\ref{tab:hyperparams}. The projector $h_\phi$ is the REPA MLP (width 2048,
two hidden layers, SiLU) applied to every token. Models with the flow policy are
trained with Eq.~\ref{eq:flow_objective}: the alignment term is weighted by
$0.5\,\lambda_t$ and the flow likelihood by $1 - \lambda_t$, with $\beta
\sigma_t$ clamped below by $\sigma_{\min} = 10^{-3}$ in the whitened residual.
The adaptive weighting of the alignment loss multiplies the per-sample error $e$
by $1 / (\mathrm{sg}(e) + 10^{-3})^p$, where $\mathrm{sg}$ stops the gradient.
Setting $p = 0$ recovers the unweighted loss, and $p = 1$ is used for the final
models. The policy scale is $\beta = 1$ during training. The control pathway of
App.~\ref{sec:design} uses $\gamma = 0.6$ and $\delta = 10^{-6}$ for the
normalization, $M = 1$ control token, and zero-initialized modulation. The flow
policy has four affine coupling layers with a hidden width of 512 and a
256-dimensional context computed from $\vh_\phi(\rvx_t, t)$ by one linear layer.
Its last layers are zero-initialized, so training starts from the Gaussian
policy. At inference we draw fresh policy noise at every step, keep the controls
active over the whole trajectory, and sweep $\beta_{\mathrm{inf}} \in \{0.5,
0.6, \dots, 0.9\}$ for each checkpoint (Fig.~\ref{fig:beta_inf_fid}). The best
value, $0.7$ for B/2 and $0.8$ for L/2 and XL/2, is the same at every training
budget.

\tabHyperparams

\paragraph{Samplers.} The Euler--Maruyama SDE sampler follows REPA and uses 250
steps with diffusion coefficient $w_t = 2t$. The Euler ODE sampler uses $N$
uniform steps from $t = 1$ to $t = 0$, so that $N$ steps cost $N$ network
evaluations. UniPC~\citep{zhao2023unipc} is run with the same step budgets at
orders 2 and 3 (Table~\ref{tab:policy_noise}). No classifier-free guidance is
used unless stated; the guided sampling protocol is described in
App.~\ref{app:guided_sampling}.

\paragraph{Evaluation.} We report FID, sFID, Inception Score, precision, and
recall computed with the ADM evaluation suite~\citep{dhariwal2021adm} against
its ImageNet $256\times256$ reference batch, on 50k samples with uniformly drawn
class labels, except for the guided results in Table~\ref{tab:gen_quality_cfg},
which use class-balanced sampling (App.~\ref{app:guided_sampling}).

\paragraph{Data efficiency.} For the data-efficiency experiment
(Fig.~\ref{fig:efficiency}c, Table~\ref{tab:data_efficiency}), we train both
REPA-B/2 and RAD-B/2 ourselves with the recipe above on random subsets of 3\%,
10\%, and 30\% of ImageNet and on the full dataset, for the same 400K iterations
at batch size 256. We evaluate every checkpoint with the Euler ODE sampler with
50 steps and report the best FID of each run.

\paragraph{Linear Probing.} Following REPA, we probe pre-trained EMA models for
RAD and REPA with the null class label. At B/2, both REPA and RAD are aligned at
block 4. For a diffusion time $t$, every image is noised once, and the hidden
state after block $k$ is mean-pooled over its tokens. A parameter-free batch
normalization followed by a linear layer is then trained on the ImageNet
training set for 90 epochs with Adam (learning rate $10^{-3}$, cosine decay,
batch size 16384) and scored by top-1 accuracy on the 50k validation images. No
augmentation is used. For RAD, the controls are sampled from the policy as at
inference, with $\beta_{\mathrm{inf}} = 0.7$ for B/2
(Fig.~\ref{fig:linear_probe}a) and $0.8$ for XL/2. Fig.~\ref{fig:linear_probe}b
reports the best block for each $t$.

\paragraph{REG within RAD.} REG denoises the class token of the encoder jointly
with the image tokens and weights its class-token loss by $0.03$. We keep this
recipe unchanged and add the control pathway of RAD with the hyperparameters
above. However, following REG we fix the injection depth at $\ell = 4$ for B/2
and $\ell = 8$ for L/2. The projection of the noisy patch tokens serves as the
mean of the representation policy, and REG's class token is not part of the
control.

\paragraph{Text-to-image.} We train on the pre-training captions of
BLIP3o~\citep{chen2025blip3o} (the JourneyDB and short-caption subsets, about 9M
image--caption pairs) at $256\times256$ with the same VAE latents. We encode the
captions using a frozen Qwen3-0.6B~\citep{yang2025qwen3} model. Its final hidden
states (dimension 1024) pass through an RMSNorm and a linear layer and are
prepended in context to the 256 patch tokens with a key mask. A pooled caption
embedding is added to the adaLN conditioning through a small MLP. We cap the
number of caption tokens to 64, which truncates 3.5\% of the total captions. The
choice of this cap is primarily driven by computational constraints. Similar to
the class labels on ImageNet, we drop the entire caption with probability $0.1$
by replacing it with a null caption whose tokens are all masked out of
attention.

The alignment loss, the flow policy, and the controls act on the patch tokens
only. RAD-B/2 injects after block 2 and RAD-XL/2 after block 8, both with $p =
1$. At XL/2, both REPA and RAD additionally use QK-normalization in attention,
without which we find training to be unstable. Pre-training follows the
class-conditional ImageNet recipe (400K iterations with a batch size of 256).

For evaluation, we use the Euler ODE sampler with 50 steps. FID is computed
against 50k images sampled from the BLIP3o pre-training set such that their
captions are diverse. For GenEval~\citep{ghosh2023geneval} benchmarking, we
generate four images for each of its 553 prompts and report the mean of its six
task accuracies.

For the qualitative results in
Figs.~\ref{fig:samples_t2i_fox}--\ref{fig:samples_t2i_cactus}, we
additionally finetune RAD-XL/2 on the BLIP3o-60k dataset. We initialize from
the pretrained EMA weights and train for 11,250 steps (50 epochs) with the
learning rate $5 \times 10^{-4}$. We use a CFG scale of 3.0, $\beta=1.0$ and the
SDE sampler with 250 steps for all qualitative samples.

\paragraph{Compute.} All experiments run on a single node with 4 NVIDIA RTX PRO
6000 Blackwell GPUs. At throughputs of roughly 10, 7, and 5
iterations per second, 400K iterations take about 11, 16, and 22 hours for B/2,
L/2, and XL/2, respectively. The text-to-image XL/2 model takes about 37 hours,
and fine-tuning it on BLIP3o-60k about 1.5 hours. Linear probing takes
about one hour for B/2 and a few hours for XL/2.

\subsection{Additional Generation Results}
\label{app:gen_results}
Tables~\ref{tab:gen_quality} and~\ref{tab:gen_quality_reg} report the full
ImageNet $256\times256$ results of RAD at each training budget, alongside the
numbers reported in the official papers of REPA and REG at the same budgets. All
results use the Euler--Maruyama SDE sampler with 250 steps and no
classifier-free guidance.

\tabGenQuality

\tabGenQualityReg

\subsection{Guided Sampling}
\label{app:guided_sampling}

\paragraph{Sampling protocol.} Guided results use RAD-XL/2 trained for 400K
and 1M iterations (80 and 200 epochs), with the hyperparameters in
Table~\ref{tab:guided_hyperparams}. The Euler--Maruyama SDE sampler uses a
uniform time grid from $t = 1$ to $t = 0.04$, followed by a final noise-free
step to $t = 0$. For noise coupling (Eq.~\ref{eq:token_coupling}),
$\boldsymbol{\xi}$ and $\boldsymbol{\eta}_n$ are drawn once per sample together
with $\rvx_1$. For CFG, the conditional and unconditional branches share the
same representation $\rvu_t$, sampled from the class-conditional policy.
Autoguidance~\citep{karras2024autoguidance} uses an earlier EMA checkpoint of
the same run as the weak model $\vv_{\theta'}$, with the same policy
noise, and is applied on top of CFG:
\begin{align}
    \tilde{\vv} &= \vv_{\mathrm{cfg}} + (w - 1)\big(\vv_\theta(\rvx_t, \rvu_t,
        t, y)
        - \vv_{\theta'}(\rvx_t, \rvu_t', t, y)\big), \\
    \vv_{\mathrm{cfg}} &= \vv_\theta(\rvx_t, \rvu_t, t, \emptyset)
        + s\big(\vv_\theta(\rvx_t, \rvu_t, t, y) - \vv_\theta(\rvx_t, \rvu_t, t,
            \emptyset)\big),
\end{align}
where $\rvu_t'$ is drawn from the weak model's own policy. Both terms apply
only within the guidance interval~\citep{kynkaanniemi2024interval}.

\tabGuidedHyperparams

\tabGuidedAblation

\paragraph{Guidance with Noise coupling.} Table~\ref{tab:guided_ablation}
shows that combining classical guidance methods like CFG and Autoguidance 
with noise coupling improves sample quality significantly while also improving 
the recall in RAD by trading off some precision. Moreover, autoguidance 
alone outperforms vanilla CFG and also
benefits from noise coupling. However, we find that it can also be used as a complement to CFG, where
it improves FID by around 0.1. For the weak model, we compared EMA checkpoints
of the same run at 50K, 100K, 120K, and 150K iterations and found 100K to work
best.

\subsection{Additional Quantitative Results}
\label{app:figure_tables}
Tables~\ref{tab:convergence_fid}--\ref{tab:linear_probe_scale} list the values
plotted in the figures of Sec.~\ref{sec:experiments} that are not already
covered by Tables~\ref{tab:gen_quality_baselines}, \ref{tab:gen_quality},
and~\ref{tab:gen_quality_reg}.

\tabConvergenceFid

\tabDataEfficiency

\tabSamplingEfficiency

\figSamplingNfeSde

\tabSamplingEfficiencySde

\tabPolicyNoise

\figBetaInfTrain

\tabBetaInf

\tabBetaTrain

\tabLinearProbeBlocks

\tabLinearProbeScale

\subsection{Additional Interpolation Results}
\label{app:interp_results}

\paragraph{Protocol.} Given two real ImageNet images $\rvx_0^A$ and $\rvx_0^B$
with labels $y_A$ and $y_B$, we take the mean of the learned policy on each
noised image as its representation, $\bar{\rvu} = \vh_\phi(\rvx_{t_c}, t_c)$
with $t_c = 0.1$, and invert each image to noise $\rvx_1$ with the Euler ODE
sampler (50 steps). For a weight $m \in [0, 1]$, we then interpolate
\begin{equation}
\label{eq:interp}
    \bar{\rvu}_m = m\,\bar{\rvu}^A + (1-m)\,\bar{\rvu}^B, \qquad
    \rvx_{1,m} = \mathrm{slerp}(\rvx_1^A, \rvx_1^B; m), \qquad
    \ve_m = m\,\ve_{y_A} + (1-m)\,\ve_{y_B},
\end{equation}
where $\ve_y$ is the class embedding. Each sample is generated from $\rvx_{1,m}$
and $\ve_m$ with the same sampler, where at every step $\rvu_t$ is sampled from
the flow policy in Eq.~\ref{eq:flow_policy} with $\beta_{\mathrm{inf}} = 0.8$
and $\vh_\phi(\rvx_t, t)$ replaced by $\bar{\rvu}_m$. Hence, $m=1$ and $m=0$
approximately reconstruct A and B, respectively. REPA does not pass the
representations back into the denoiser, so its generations cannot be manipulated
through them directly. Hence, we interpolate $\rvx_1$ and the class embedding in
the same way.

\paragraph{Additional samples.} We show additional interpolations between real
ImageNet images across classes (Fig.~\ref{fig:interp_cross}) and within the same
class (Fig.~\ref{fig:interp_same}), using the protocol above and the checkpoints
of Sec.~\ref{sec:interpolation}. For same-class pairs, the class embedding is
fixed, so REPA interpolates $\rvx_1$ alone. As in Fig.~\ref{fig:interpolation},
RAD preserves the pose and framing of the endpoints while the appearance changes
gradually, e.g., the iguana's pale scales darken into the textured skin of the
Komodo dragon, the eye of the green chameleon grows into the turreted eye of the
African chameleon, and the white-lit castle turns into the orange-lit palace
while the night scene and its reflection on the water are kept. Within the same
class, RAD keeps the side profile of the wood rabbit as the background changes
from dry brush to green foliage. REPA again does not remain consistent: the head
of the iguana dissolves into a full-body lizard among leaves, the close-up of
the chameleon is replaced by full-body chameleons on branches, and the wood
rabbit is replaced by a rounder, crouched rabbit facing the opposite direction
for $0.35 \leq m \leq 0.5$.

\figInterpCross

\figInterpSame

\clearpage
\makeatletter
\setlength{\@fptop}{0pt}
\makeatother
\subsection{Additional Samples (Class Conditional ImageNet)}
\label{app:samples}

\figSamplesMacaw

\figSamplesLorikeet

\figSamplesLesserPanda
\clearpage

\figSamplesGiantPanda

\figSamplesRedFox

\figSamplesAnemoneFish
\clearpage

\figSamplesAgaric

\figSamplesCoralFungus

\figSamplesStrawberry
\clearpage

\figSamplesTeapot

\figSamplesSchooner

\figSamplesChurch
\clearpage

\figSamplesArcticFox

\figSamplesValley

\figSamplesCheeseburger

\clearpage
\subsection{Additional Text-to-Image Samples}
\label{app:samples_t2i}

\figSamplesTextToImageFox

\figSamplesTextToImageCat

\figSamplesTextToImageTeacup

\clearpage

\figSamplesTextToImageRetriever

\figSamplesTextToImageTerrarium

\figSamplesTextToImagePhoenix

\clearpage

\figSamplesTextToImageCastle

\figSamplesTextToImageBedroom

\figSamplesTextToImageCactus

\end{document}